\documentclass{article}
\usepackage{spconf,amsmath,graphicx}
\usepackage[ruled,vlined]{algorithm2e}
\usepackage{cite}
\usepackage{amsmath,amssymb,amsfonts}
\usepackage{algorithmic}
\usepackage{graphicx}
\usepackage{textcomp}
\usepackage{xcolor}
\usepackage{mathtools, nccmath}
\usepackage{times}
\usepackage{epsfig}
\usepackage{graphicx}

\usepackage{multirow}
\usepackage{hyperref}
\usepackage{subcaption}
\usepackage{float}
\title{LIGHTS: LIGHT Specularity Dataset for specular detection in Multi-view}
\name{Mohamed Dahy Elkhouly$^{\star,\ddagger}$,Theodore Tsesmelis$^{\star}$,
         Alessio {Del Bue}$^{\star,\dagger}$, Stuart James$^{\star}$
    \thanks{This project has received funding from the European Union's Horizon 2020 research and innovation programme under grant agreement No 870723.}}
\address{$^{\star}$Visual Geometry and Modelling (VGM) Lab, Istituto Italiano di Tecnologia (IIT), Italy \\
$^{\dagger}$Pattern Analysis and Computer Vision (PAVIS), Istituto Italiano di Tecnologia (IIT), Italy \\
$^{\ddagger}$Universit\`{a} degli studi di Genova, Italy}
\begin{document}
%\ninept
%
\maketitle
\begin{abstract}
Specular highlights are commonplace in images, however, methods for detecting them and in turn removing the phenomenon are particularly challenging. A reason for this, is due to the difficulty of creating a dataset for training or evaluation, as in the real-world we lack the necessary control over the environment. Therefore, we propose a novel physically-based rendered LIGHT Specularity (LIGHTS) Dataset for the evaluation of the specular highlight detection task. Our dataset consists of $18$ high quality architectural scenes, where each scene is rendered with multiple views. In total we have $2,603$ views with an average of $145$ views per scene. Additionally we propose a simple aggregation based method for specular highlight detection that outperforms prior work by $3.6\%$ in two orders of magnitude less time on our dataset. 
\end{abstract}
\section{Introduction}
Specular highlights occur when the light source and the direction of the viewer are halfway between the normal to the surface. In most cases specular highlights are nuisances creating undesired artifacts in the image, however, they can also be useful for determining the direction of the light source within the environment in such cases as photometric stereo or surface light field reconstruction~\cite{Park_2020_CVPR}.
% Lack of validation datasets

\begin{figure*}[t]
    \centering
    \begin{subfigure}[b]{\linewidth}
        \includegraphics[width=0.195\linewidth]{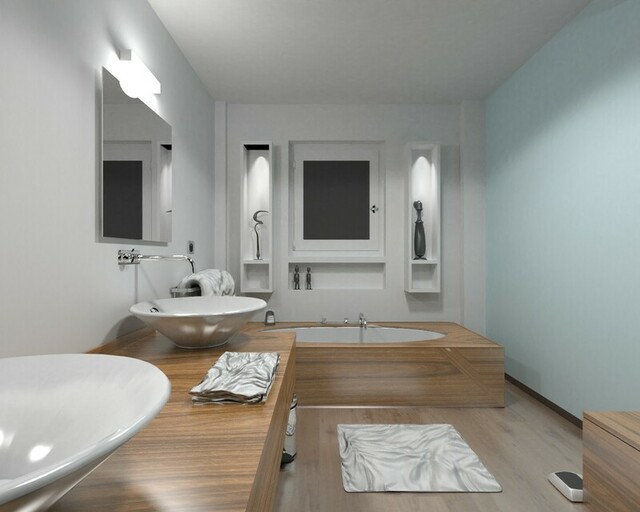}
        \includegraphics[width=0.195\linewidth]{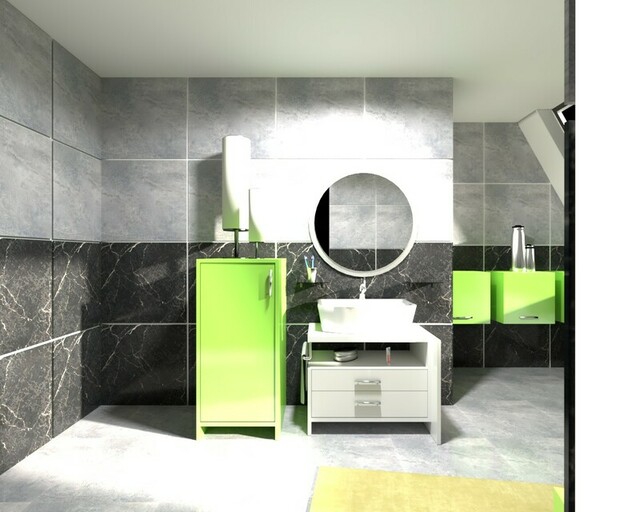}
        \includegraphics[width=0.195\linewidth]{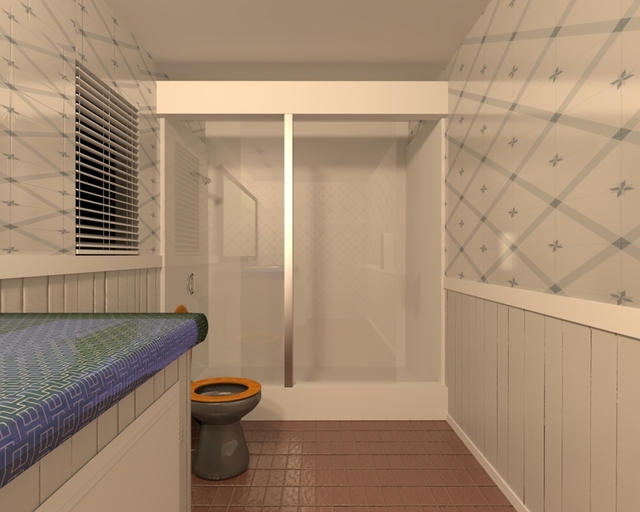}
        \includegraphics[width=0.195\linewidth]{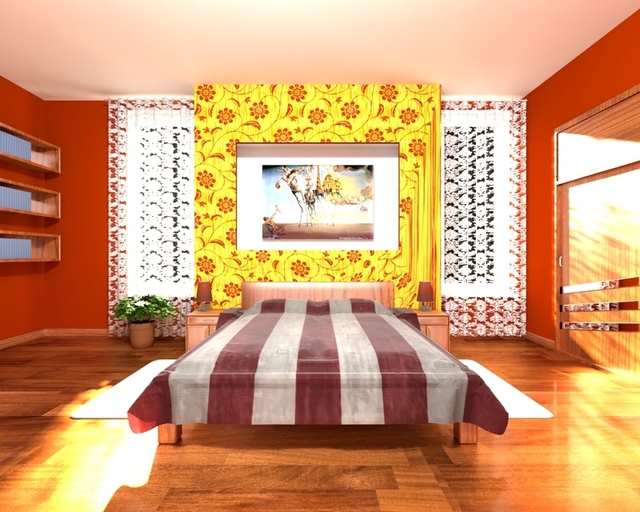}
        \includegraphics[width=0.195\linewidth]{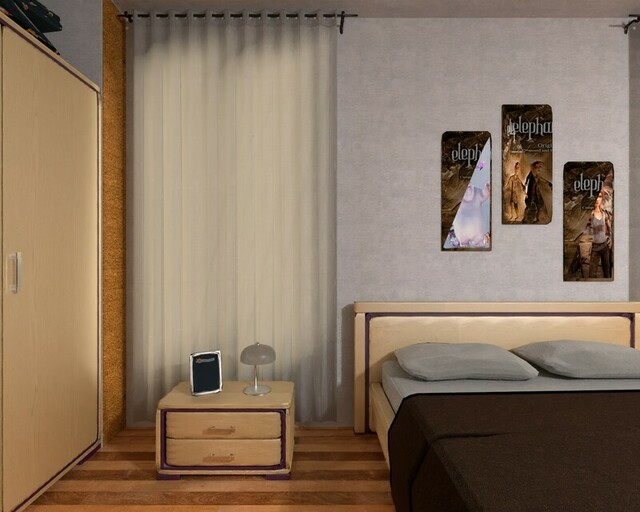} 
        \includegraphics[width=0.195\linewidth]{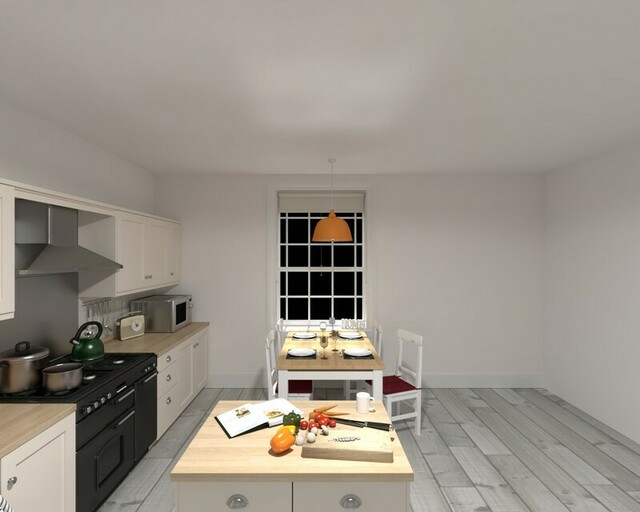}
        \includegraphics[width=0.195\linewidth]{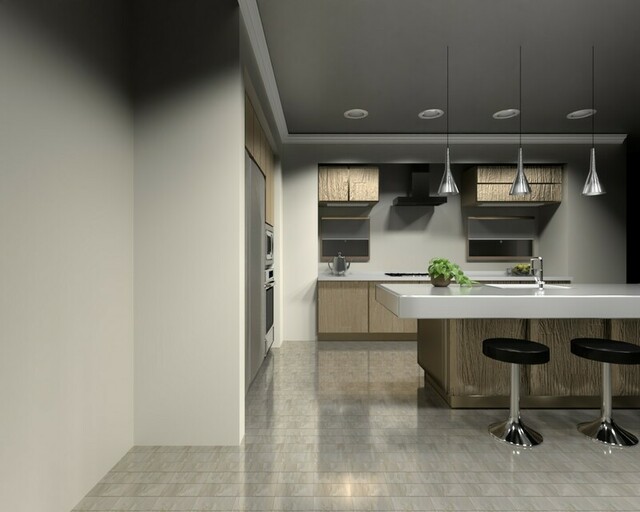}
        \includegraphics[width=0.195\linewidth]{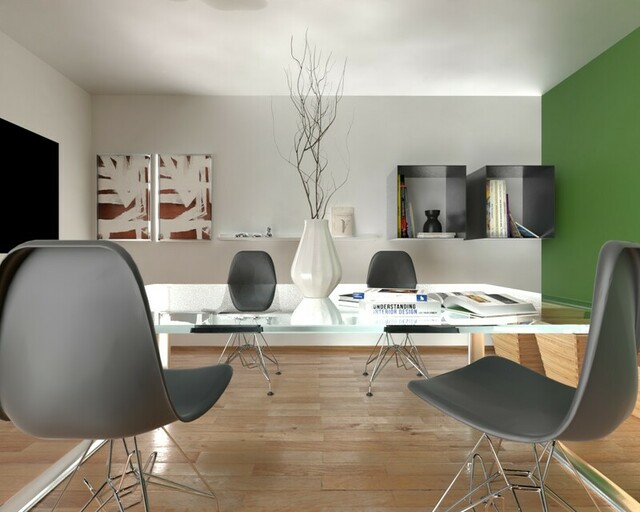}
        \includegraphics[width=0.195\linewidth]{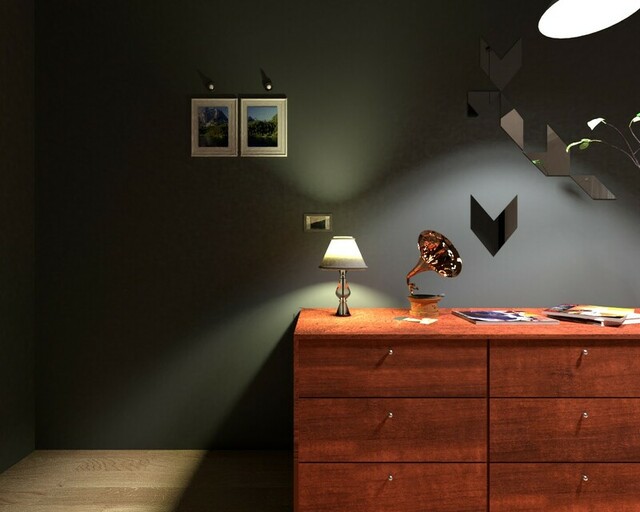}
        \includegraphics[width=0.195\linewidth]{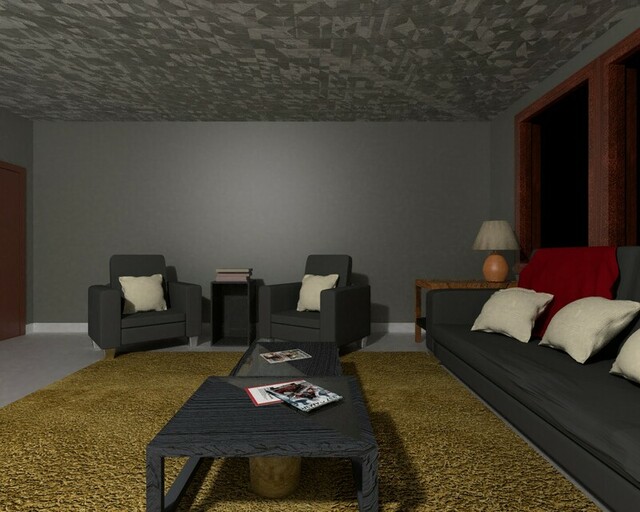}
        \includegraphics[width=0.195\linewidth]{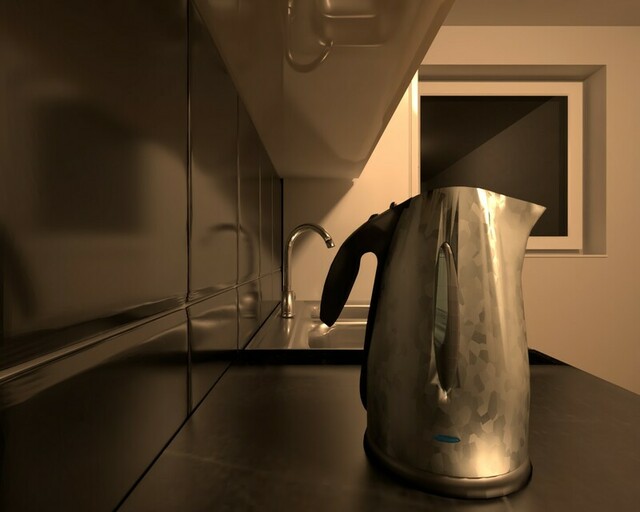}
        \includegraphics[width=0.195\linewidth]{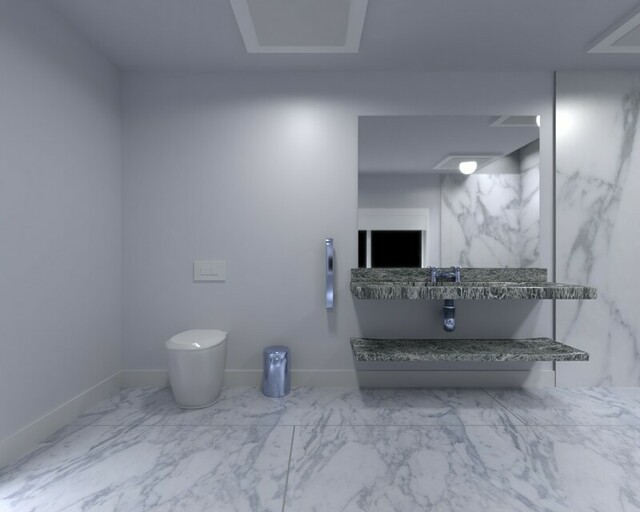}
        \includegraphics[width=0.195\linewidth]{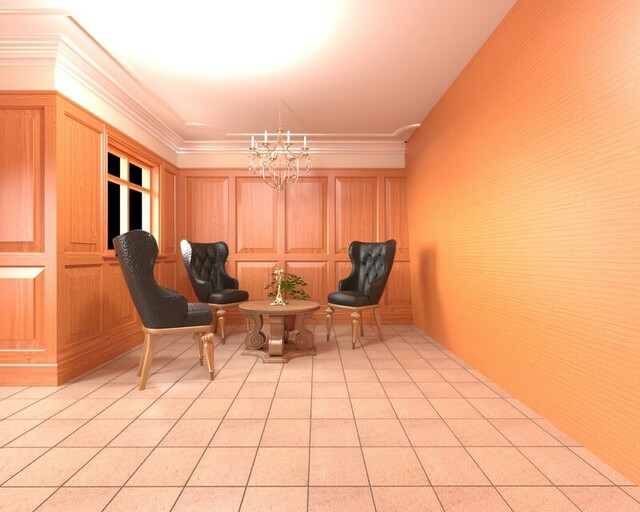}
        \includegraphics[width=0.195\linewidth]{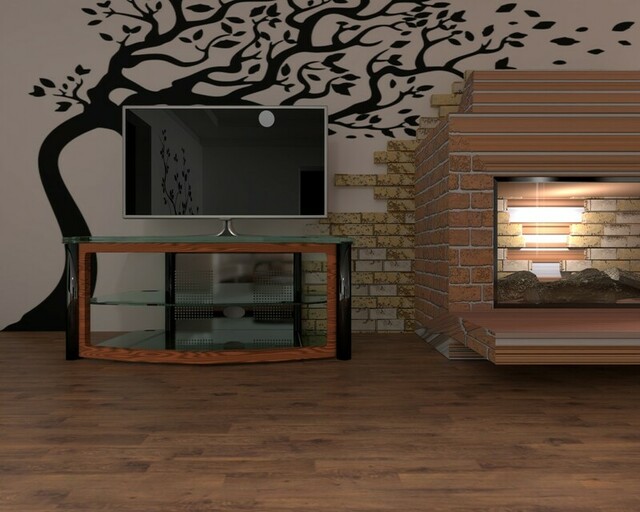}
        \includegraphics[width=0.195\linewidth]{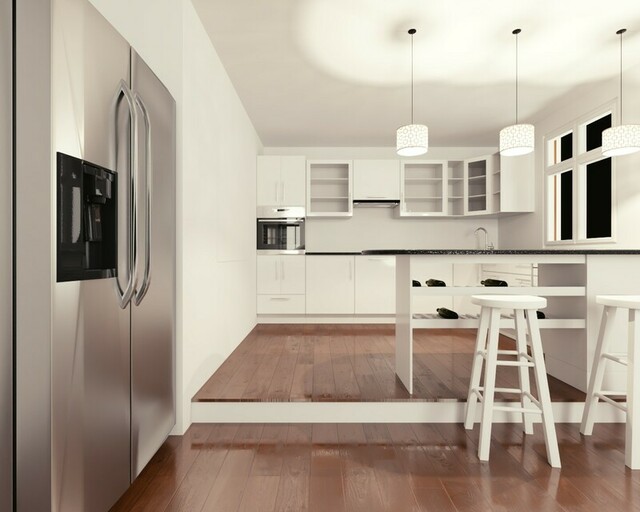}
    \end{subfigure}
    \vspace{20pt}
     \begin{subfigure}[b]{0.072\linewidth}
         \centering
         \includegraphics[width=\linewidth]{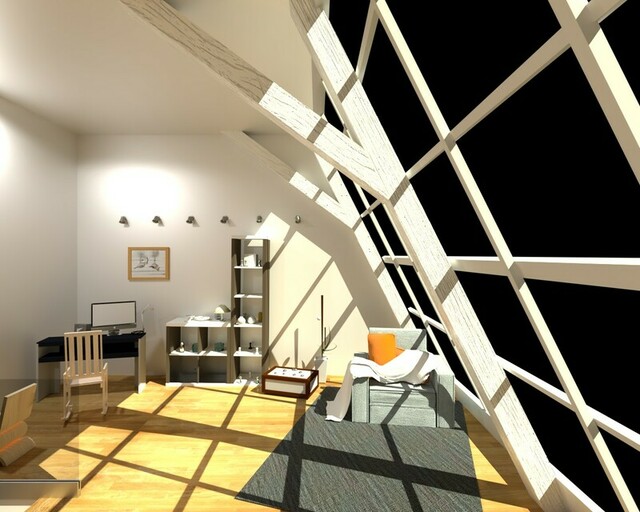}
         \includegraphics[width=\linewidth]{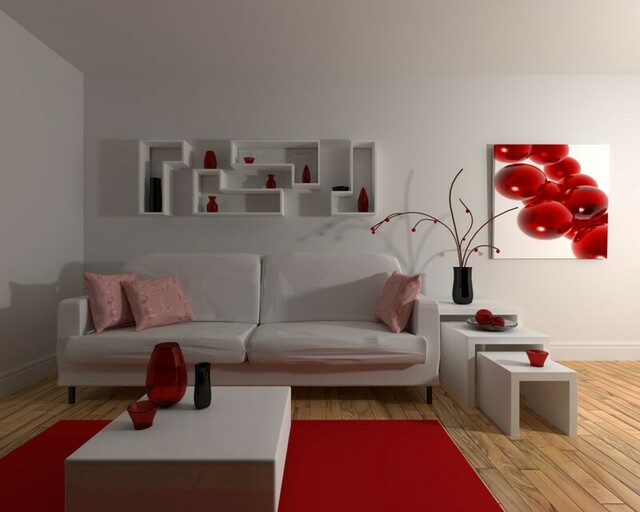}
         \caption{ }
         \label{fig:dataset_depth}
     \end{subfigure}
     \begin{subfigure}[b]{0.072\linewidth}
         \centering
         \includegraphics[width=\linewidth]{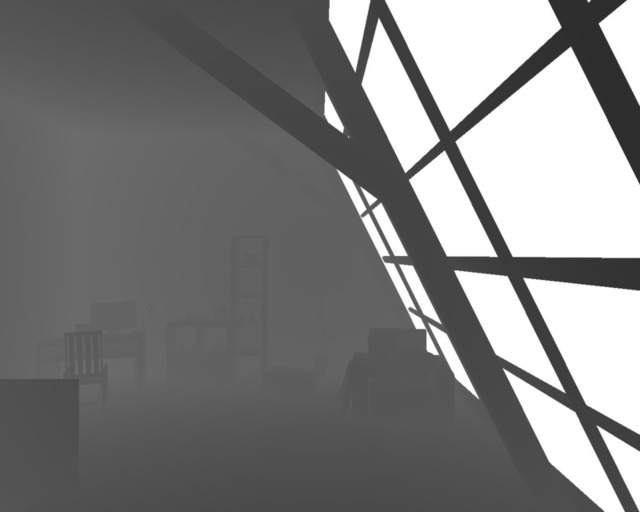}
         \includegraphics[width=\linewidth]{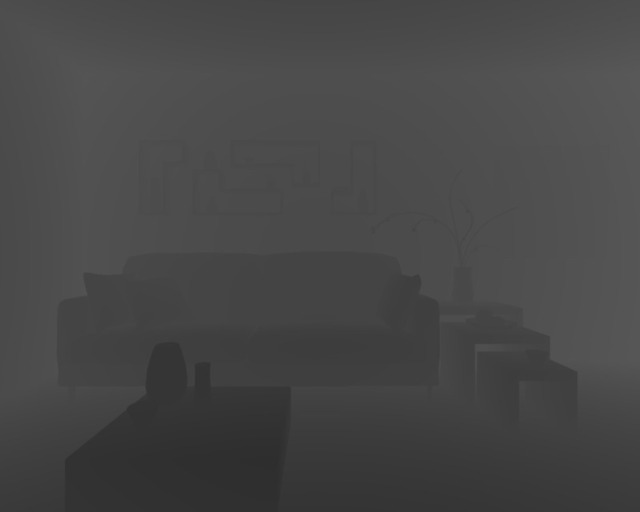}
         \caption{ }
         \label{fig:dataset_depth}
     \end{subfigure}
     \begin{subfigure}[b]{0.072\linewidth}
         \centering
         \includegraphics[width=\linewidth]{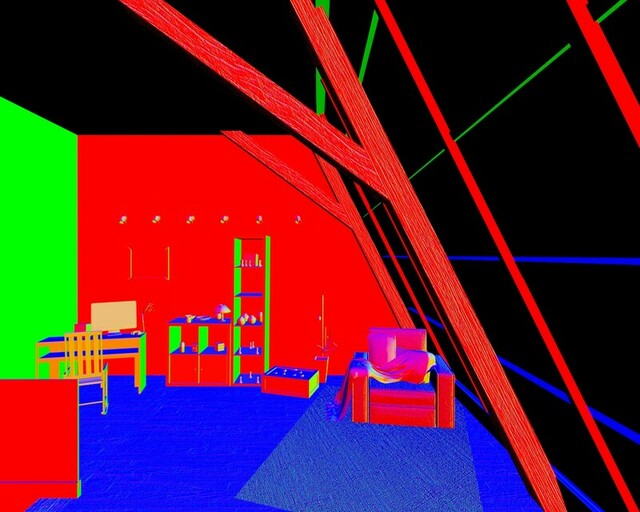}
         \includegraphics[width=\linewidth]{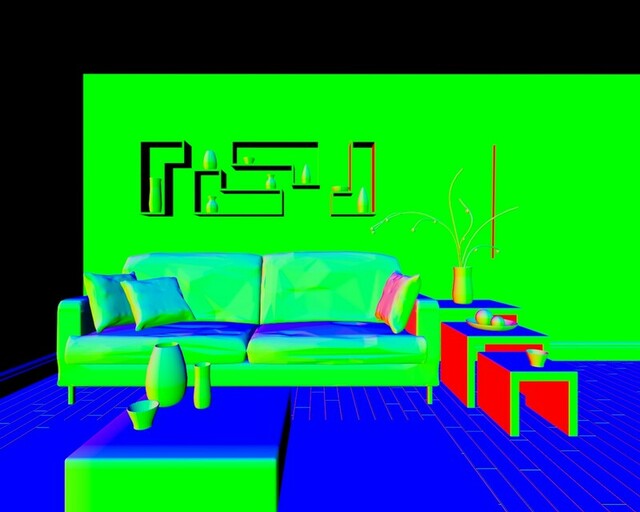}
         \caption{ }
         \label{fig:dataset_normal}
     \end{subfigure}
     \begin{subfigure}[b]{0.072\linewidth}
         \centering
         \includegraphics[width=\linewidth]{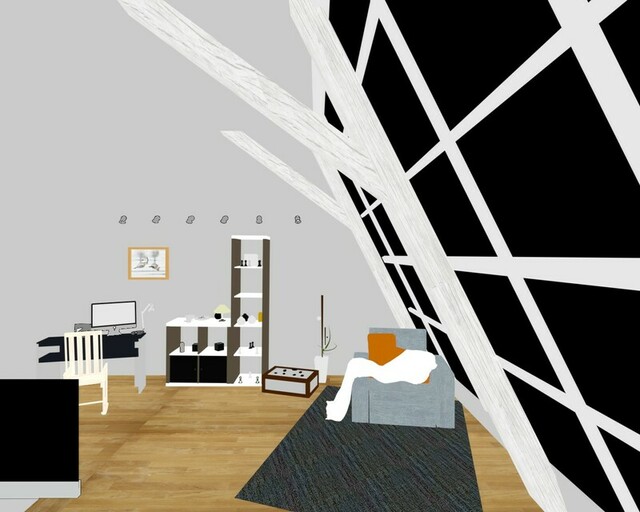}
         \includegraphics[width=\linewidth]{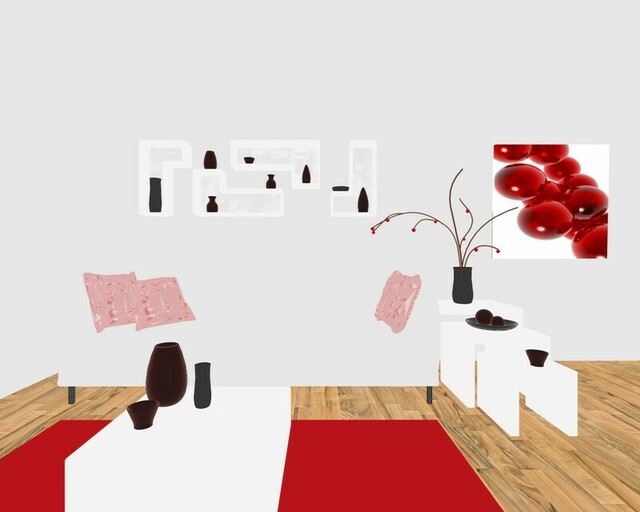}
         \caption{ }
         \label{fig:five over x}
     \end{subfigure}
     \begin{subfigure}[b]{0.072\linewidth}
         \centering
         \includegraphics[width=\linewidth]{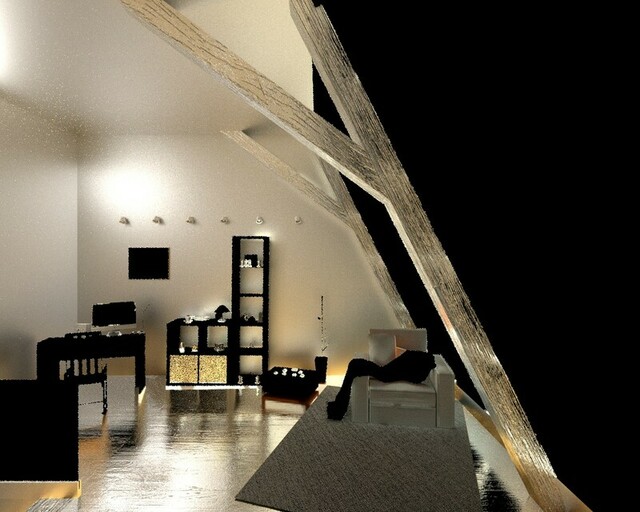}
         \includegraphics[width=\linewidth]{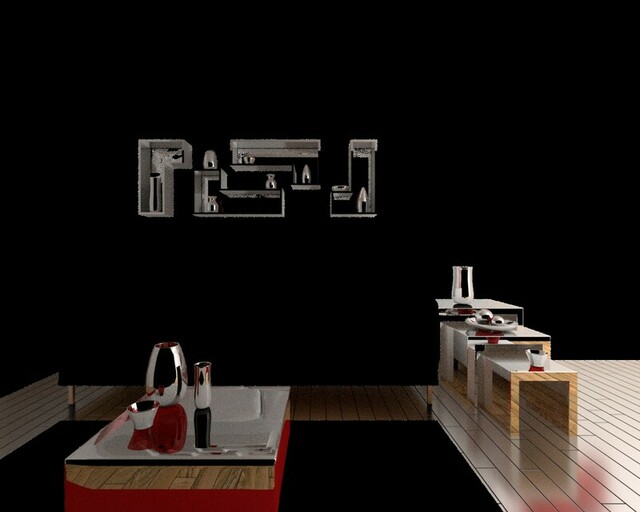}
         \caption{ }
         \label{fig:five over x}
     \end{subfigure}
     \hfill
     \begin{subfigure}[b]{0.072\linewidth}
         \centering
         \includegraphics[width=\linewidth]{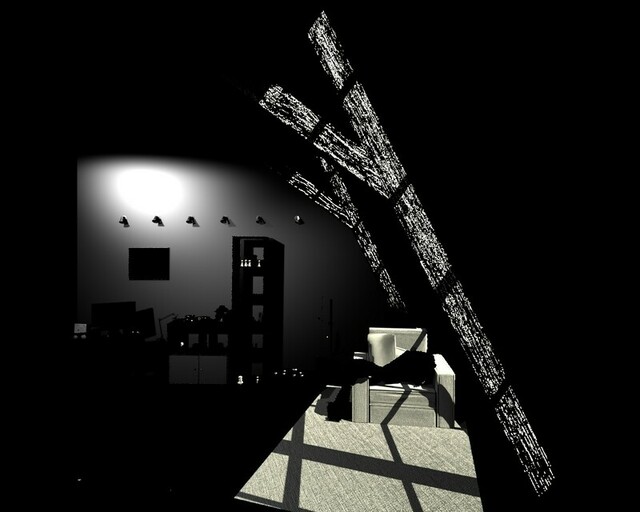}
         \includegraphics[width=\linewidth]{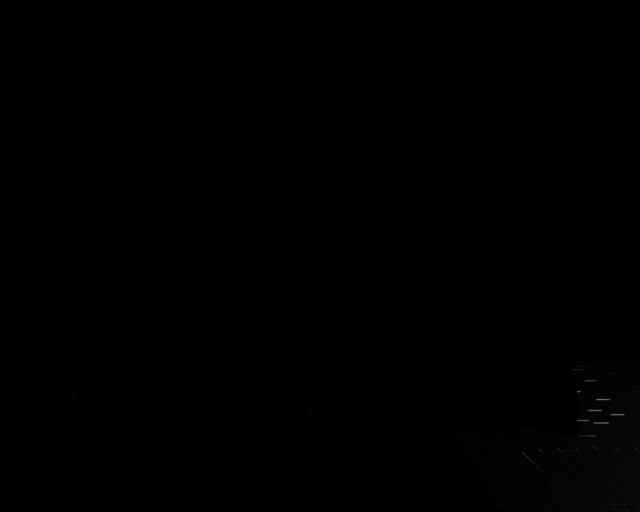}
         \caption{ }
         \label{fig:five over x}
     \end{subfigure}
     \begin{subfigure}[b]{0.072\linewidth}
         \centering
         \includegraphics[width=\linewidth]{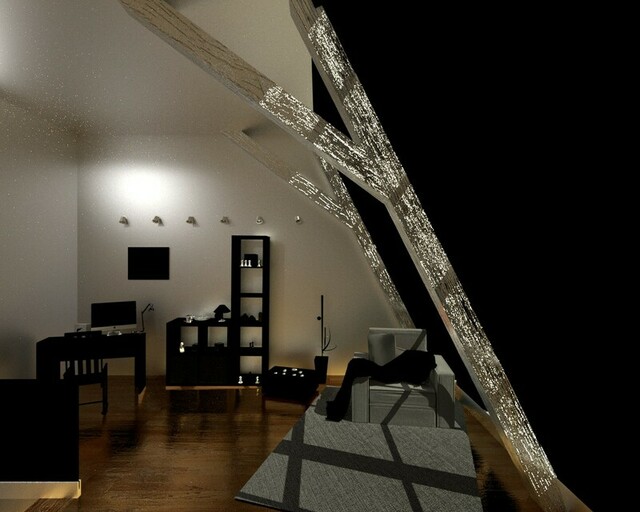}
         \includegraphics[width=\linewidth]{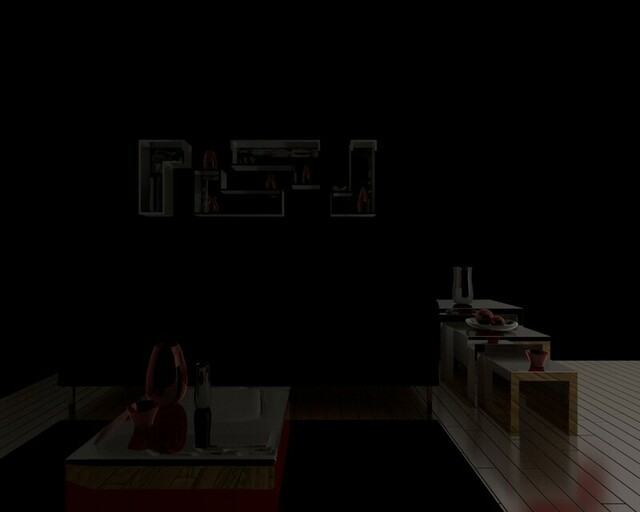}
         \caption{ }
         \label{fig:five over x}
     \end{subfigure}
     \begin{subfigure}[b]{0.072\linewidth}
         \centering
         \includegraphics[width=\linewidth]{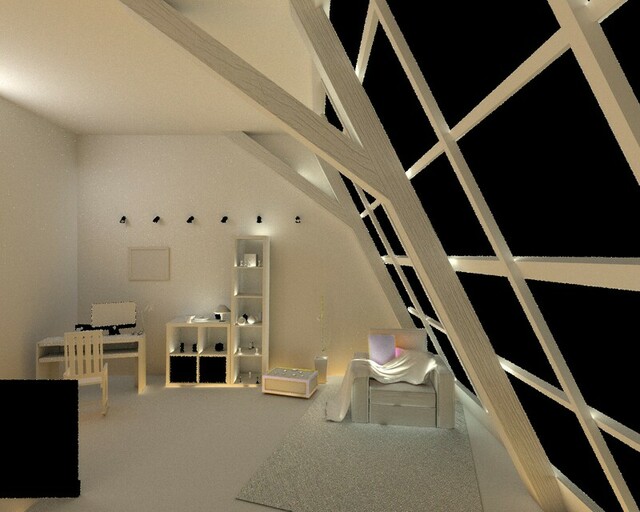}
         \includegraphics[width=\linewidth]{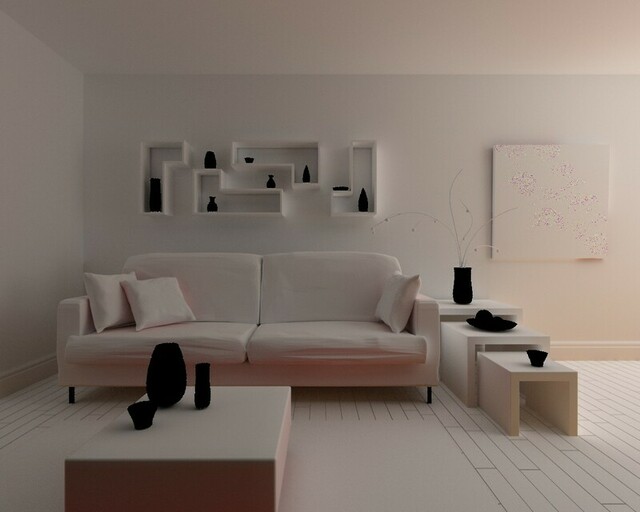}

         \caption{ }
         \label{fig:five over x}
     \end{subfigure}
     \begin{subfigure}[b]{0.072\linewidth}
         \centering
         \includegraphics[width=\linewidth]{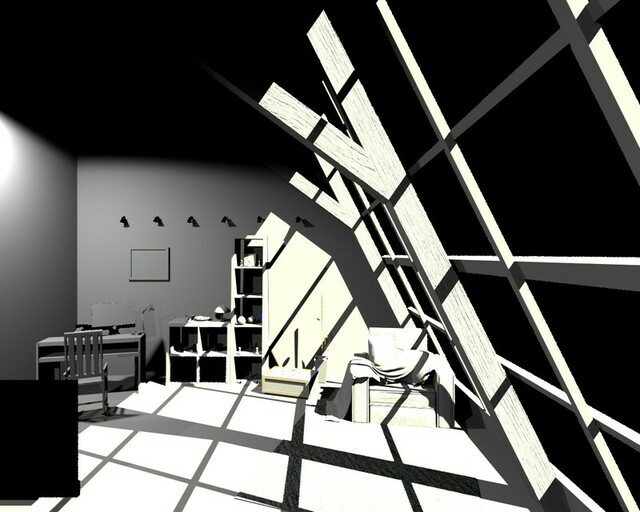}
         \includegraphics[width=\linewidth]{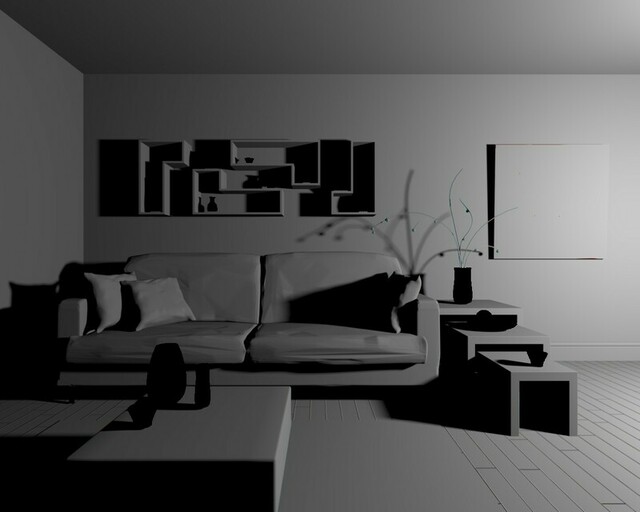}
         \caption{ }
         \label{fig:five over x}
     \end{subfigure}
     \begin{subfigure}[b]{0.072\linewidth}
         \centering
         \includegraphics[width=\linewidth]{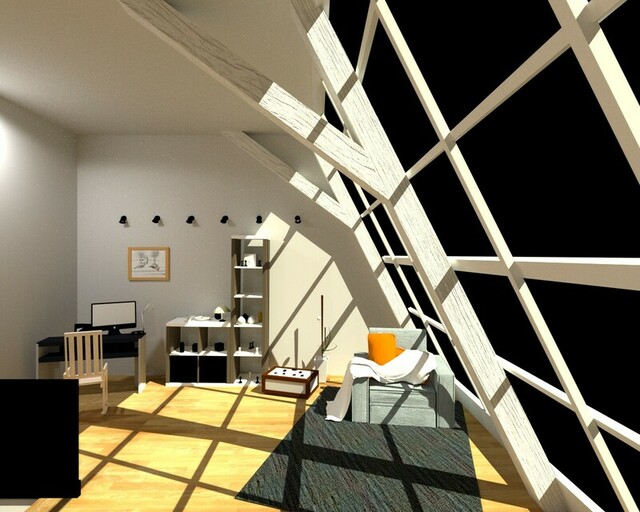}
         \includegraphics[width=\linewidth]{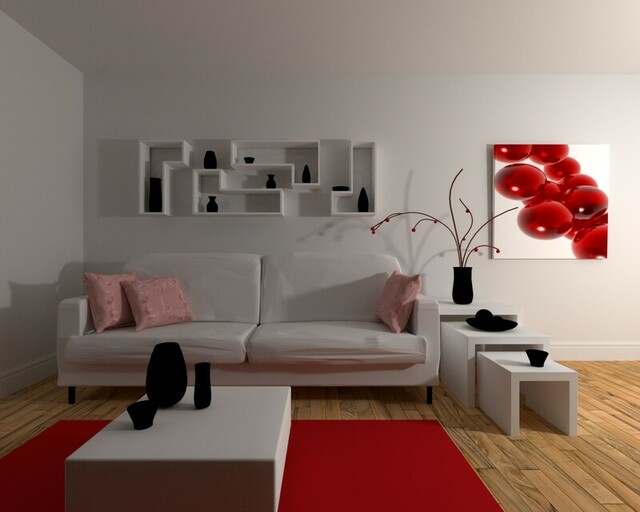}
         \caption{ }
         \label{fig:five over x}
     \end{subfigure}
     \begin{subfigure}[b]{0.072\linewidth}
         \centering
         \includegraphics[width=\linewidth]{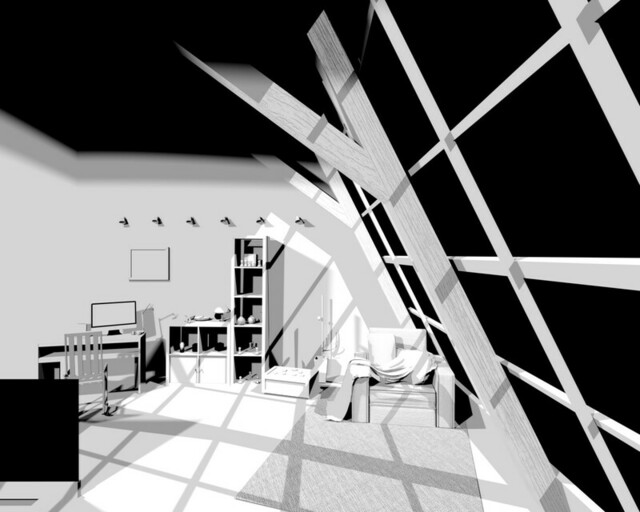}
         \includegraphics[width=\linewidth]{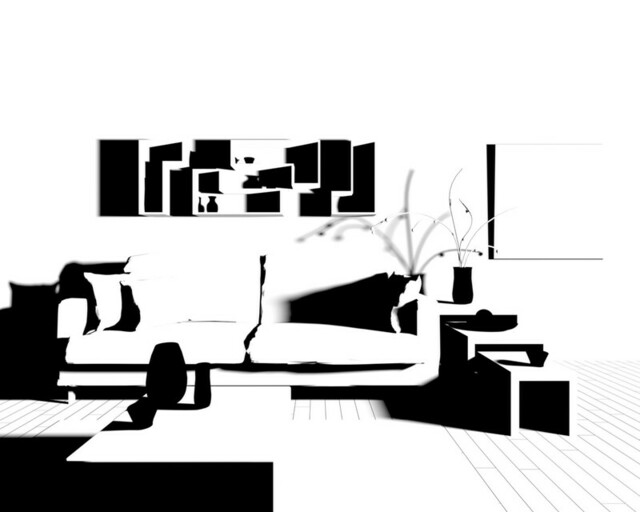}
         \caption{ }
         \label{fig:five over x}
     \end{subfigure}
     \begin{subfigure}[b]{0.072\linewidth}
         \centering
         \includegraphics[width=\linewidth]{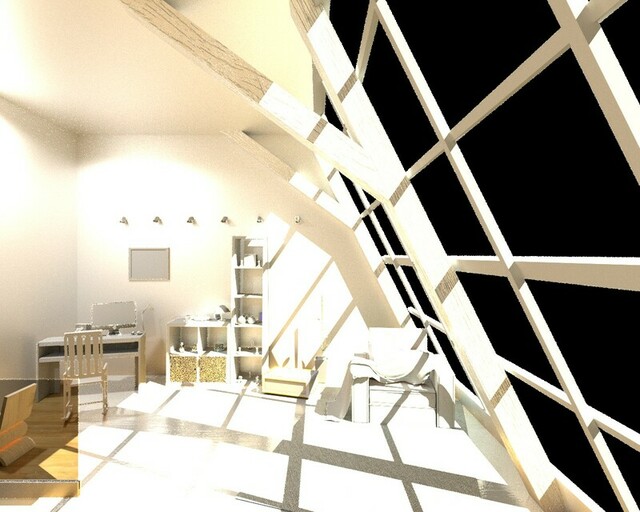}
         \includegraphics[width=\linewidth]{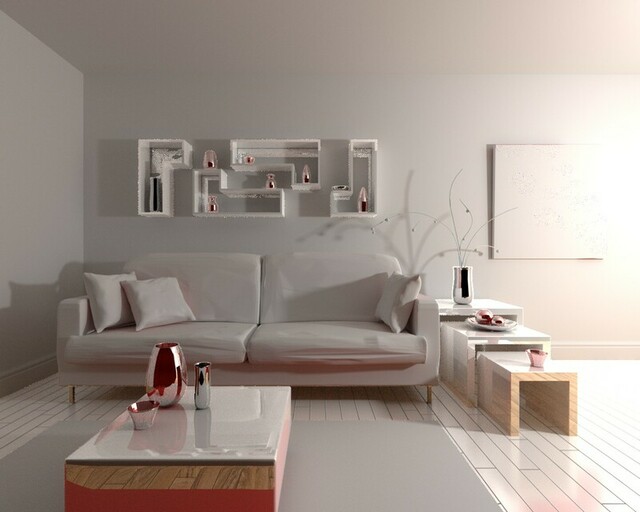}
         \caption{ }
         \label{fig:five over x}
     \end{subfigure}
     \begin{subfigure}[b]{0.072\linewidth}
         \centering
         \includegraphics[width=\linewidth]{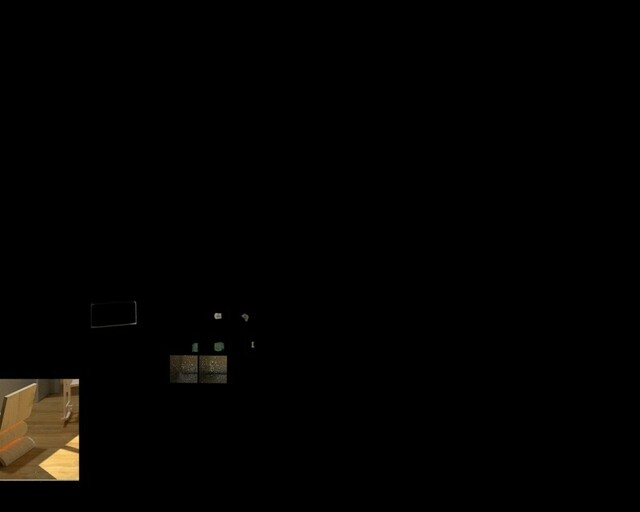}
         \includegraphics[width=\linewidth]{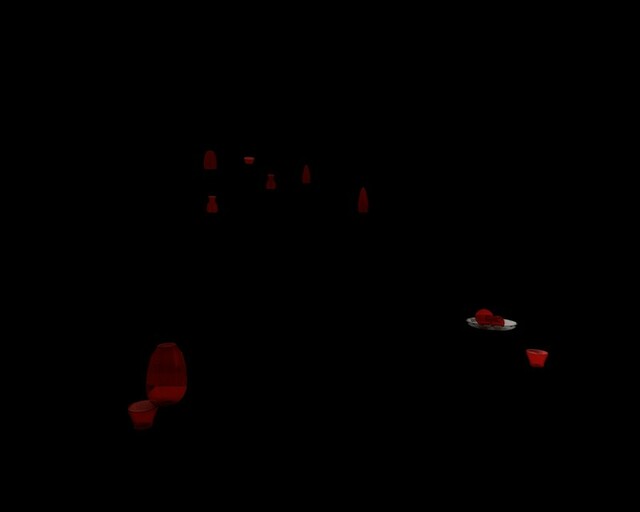}
         \caption{ }
         \label{fig:five over x}
     \end{subfigure}
  \vspace{-20px}
  \caption{Top: Example renderings of LIGHTS scenes (15 of 18)  with variation across types of room (kitchens, bedrooms, bathrooms and living-rooms). Bottom: Two examples of the rendered components of the dataset (a) Image (b) Depth (c) Normal (d) Albedo  (e) Indirect Specular (f) Direct Specular (g) Specular (h) Indirect Diffuse (i) Direct Diffuse (j) Diffuse (k) Shadow map (l)  Light map (m) Transmission.}
  \label{fig:ourdataset}
\end{figure*}

A challenge to specular highlight detection is the difficulty in creating a real-world dataset with accurate groundtruth for studying the phenomenon. In real-world images, the multitude of light sources within a scene makes accurate human annotation challenging. Existing 3D datasets such as Matterport3D~\cite{Chang3DV17Matterport3D} %or ScanNet~\cite{daiCVPR17scannet} 
would be considered too uncontrolled in terms of lighting and capture setup to create an accurate ground-truth. This leads to the point where specular detection techniques are focusing on datasets captured in controlled environments, e.g. labs, for approaches to  single images as well as for multi-view images~\cite{nurutdinova2017specularity,shah2017removal} making them unsuitable for general purpose applications. Alternatively, other synthetic datasets such as SunCG \cite{song2017semantic} lack sufficient model quality to recover specular highlights. Therefore,
% in this work 
we propose the
% LIGHT Specularity (
LIGHTS
% ) 
Dataset\footnote{Dataset: \url{https://pavis.iit.it/datasets/lights}}, constructed from high-quality architectural 3D models with variation in lighting design and rendering parameters to create near photo-realistic scenes (see{ Fig.~\ref{fig:ourdataset}}).
%  \begin{figure}[!t]
%   \centering
%   \includegraphics[width=\linewidth]{IMG/teaser_1.png}
%   \caption{Pipeline of our system. From the LIGHTS datast we first acquire multi-view images where specular highlights are apparent which we then process through our face based multi-view detector.}
%   \end{figure}
% Challenes of specular detection

The detection of specular highlights requires scene understanding, for example knowing the surface geometry, material properties, and the light source(s) position. Data-driven approaches, especially from single-view, fail to encompass this information resulting in a largely color-based detection that treats anomalous white regions as highlights. While in many cases this assumption holds true, it fails to understand the cause of the phenomenon and unduly discriminate against white objects. In contrast, we exploit the 3D geometry and estimate the surface properties based on the consensus of multiple views that reduces the influence of highlights. Our proposed method exploits the generalization of single view based methods while incorporating the surface geometry in aggregating across views to detect specularities.
Therefore, the contribution of this paper is two-fold:\\ (1) A photo-realistic dataset for specular highlight analysis; (2) A framework for multi-view extension from single-view approaches for specular highlight detection.

%%%%%%%%% BODY TEXT
\section{Related Work}\label{sec:related_work}
We review the literature for single and multi-view approaches to specular highlight detection as well as related datasets.\\
%However, in general specular highlight detection approaches encompass different sub-topics related to detection, removal, reflection separation and intrinsic image decomposition.

\noindent\textbf{Single image:}\label{subsec:single_related_work}  In the literature techniques can be grouped mainly into two categories~\cite{Alessandro2011CGF} the ones related to a) color space analysis techniques and b) neighborhood analysis techniques. 
In the first category, early works are based on color analysis and chromaticity distribution to extract image diffuse and specular components ~\cite{AKASHI2016CVIU}. Klinker et al \cite{klinker1988IJCV} used an average object color to separate the highlight, essentially identifying an anomaly, while \cite{schluns1995CIC} normalized values of the YUV color space. However, both methods are therefore struggle with white (or near white) objects. On the other hand Bajcsy et al\cite{bajcsy1996IJCV} proposed a specific new color space, S-space, which was discriminating the color variation of objects in terms of brightness. Alternatively, the second category focuses on the spatial consistency of pixel neighborhoods which exhibit similar properties ~\cite{Tan2005PAMI} or through clustering~\cite{Souza2018SIBGRAPI} while~\cite{TakahisaYamamoto2019ITE} proposed a post-processing technique to improve the separation of reflection components. 
Recently Lin et al~\cite{lin2019deep} used a GAN to correct for specular errors but was susceptible to darkening the image as opposed to correcting it, and trained on synthetic images. \noindent\textbf{Multi-view} techniques on the other hand, especially in the early works~\cite{SangECCV92} used Color Histogram Differencing (CHD) which suffered from occlusion as it did not consider geometric information. This limitation was addressed by ~\cite{Lin2002ECCV} which proposed a Multi-CHD by using consensus from views. 
Other lab-based setups exploit lighting~\cite{Agrawal2005ACM} %,Feris2004proc} 
or polarization at different orientations~\cite{WANG2017CVIU}%,lamond2007fast}
, however, such approaches requires a bespoke setup and does not generalize well. Enforcing sequential information for multi-view specular detection~\cite{Haghighat2020} used a low-rank approximation for the separation of specular and diffuse components. In contrast, our approach focuses on exploiting the geometry and not requiring sequential images in unconstrained non-lab settings.
\noindent\textbf{Datasets:} existing datasets for specular analysis are small in size from $2-7$ images~\cite{Tan2005PAMI} and/or with limited: ground truth, specular-free images, or variation of objects. In the lab constrained dataset of ~\cite{park2003truncated}, they used $100$ objects (generally single color) captured only in three illumination conditions. Alternatively,~\cite{lin2019deep} generated a synthetic dataset of $8$ images, for which each image contained one object, however it was not made public. In the contrary, our dataset we include numerous objects, a high level of detail, variation of backgrounds, as well as different lighting setups per scene from different graphic designers, moreover, it is a near photo-realistic. 
% . Each object in the scene is part of a larger environment with surroundings which means 
% and great background variations.
\section{Light Specularity Dataset (LIGHTS)} \label{subsec:dataset}
The LIGHTS dataset provides indoor photo-realistic images based on Physically Based Rendering scenes. We propose the dataset for the evaluation and comparison of techniques related to the interaction between objects and light and more specifically in our case with specular highlights. However, the dataset includes a wide range of images and complimentary information that makes it amenable for other applications beyond intrinsic image based problems such as shadow detection, normal estimation or depth estimation. 

The dataset is composed of 18 different scenes including bedrooms, bathrooms and living-rooms and are based on professional architectural designed CAD models. The spatial structure is designed to improve light scattering, while we further adjusted the light intensities to different levels (e.g. high, normal, and low) for further variation and more intense light phenomena, i.e. specularities, as a result of direct and/or indirect lighting.  Direct specularity is resulting from light bouncing on surface directly from the light source, while indirect specularity is a cause of light reflected from another surface for example a mirror. We provide $2603$ rendered views based on Blender %~\cite{blender18}  
with the cycles rendering engine and path tracing for the light transportation model. Moreover, we carefully adjusted the parameters for each light in the collected scenes. The renders were created in $1280\times1024$ resolution. A subset of the dataset can be seen in Fig.~\ref{fig:ourdataset} with examples of the complementary information. In contrast to prior datasets our dataset has rich material properties in contrast to SunCG~\cite{song2017semantic}, a high amount of camera variation in contrast to MatterPort3D~\cite{Chang3DV17Matterport3D}, and three orders of magnitiude more images than prior directly related datasets~\cite{Tan2005PAMI}.%,park2003truncated}.

The LIGHTS dataset is organized using similarly principles to Matterport3D for easy inclusion within existing pipelines. Each scene contains the following information:

\noindent\textbf{Camera information (Cam\_Info):} includes the intrinsic ($K \in {\rm I\!R}^{4\times 4}$) and rotation translation ($C \in {\rm I\!R}^{3\times 3}$) matrices following the Matterport3D prototype. 

\noindent\textbf{Light sources information (Light\_info):}, provides the type (Point, Sun, Spot, etc.), position, orientation, color, and strength encoded as a key-value structure in JSON files. 

\noindent\textbf{Mesh information (Mesh\_info):} The 3D mesh geometry ($x,y,z$) without color for the scene in a .ply file. 

\noindent\textbf{Rendering information (Rendering\_info):} which includes all the different maps describing the scene properties, i.e. albedo, diffuse, specular, transmission, depth, normals, lightmap, object segmentation mask as well as the final rendering images. Visualized in full in the Supp. Mat.

\noindent\textbf{Blender (Blender\_Data)} include the ``.blend'' files for each scene for direct use and the license files defining the terms of use.
% respectively. 

\noindent\textbf{License information (License\_info):}For the scenes selected they use either a CC0 Public Domain or CC3.0 Attribute (Share, Adapt for non-commercial use). We include a full acknowledgment of the creators in the supplementary material.

\subsection{Light modeling}
We summarize the rendered components of the dataset as Albedo, Depth and Normal are common for 3D datasets so we only mention unique components of the dataset, detailed here.
\noindent\textbf{Specular:} we render both the direct and indirect specular components. In contrast to prior datasets that ignore indirect specular effects, e.g. mirrors or highly reflective surfaces, we retain the indirect specular component. The combination of both are also included for simplicity.
\noindent\textbf{Diffuse:} in contrast to specular indirect and direct components, diffuse are common in 3D datasets and represent the ambient light falling on the surface within the Lambertian assumption.
\noindent\textbf{Shadow:} shadow map is normalized to range $[0-1]$ for shadows created by the light occlusion on objects. 
\noindent\textbf{Light Map:} defines the amount of light falling on the surface and can be seen as the combination of intensity of specular and diffuse light.
\noindent\textbf{Transmission:} is the amount of light passing through an object. Although in our scenes there are not high number of transparent or translucent objects this could be useful for their detection.

% \begin{algorithm}[t]
% \SetAlgoLined
% \KwResult{Specular faces at each view}
%  Calculate potential specularity mask $p$ using a single image detector\;
%  \For {each frame $i$}{
%   \For {each face in $i$}{
%   use the mask $p$ to mark faces as specular $s$ or diffuse $d$ using  Eq. \ref{eq:getpotentials}.
%   }
%   \For {each frame $j$ different from $i$}{
%   \eIf{exist common $d$ faces with frame $i$}{
%   calculate threshold $t$ using Eq. \ref{eq:specular_threshold}\;
%   exclude faces from $s$ using Eq. \ref{eq:excludedecision}\;
%   }{
%   continue\;
%   }
%   }
%   specular faces in frame $i$=$s$\;
%  }
%  \caption{Multi-view specular detection}
%  \label{alg:algorithm}
% \end{algorithm}
\section{Multi-view specular detection}\label{sec:mvdetection}
Our Multi-view method makes use of the single image method~\cite{Souza2018SIBGRAPI} as a preprocessing step to the scene geometry extracted from the multi-view images. Therefore, our method takes as input the luminance channel of a set of images in CIELAB color space as $L=\{L_{0}, ..., L_{n}\}$ where $L_i \in {\rm I\!R}^{w\times h}$, a set of camera matrices $C=\{C_{0}, ..., C_{n}\}$, and faces of a mesh $f=\{f_{0}, ..., f_{z}\}$ where $n, z$ is the number of images and faces respectively. For each $L_{i}$, we compute the corresponding specularity mask $M_{i}$, using~\cite{Souza2018SIBGRAPI}. 
For a given image luminance $L_i$, each face $f_{k}$ is backprojected using $C_i$ as $f_{ik}$ after occlusion and view culling, $f_{ik}$ has a set of corresponding pixels $h_ik$. The set of faces are then split into specular and diffuse components by thresholding with Eq.~\ref{eq:getpotentials}:
\begin{equation} \label{eq:getpotentials}
(S,D) = 
\begin{cases}
    s,& \text{if }  \frac{\sum_{p=1}^l {M_i(h_{ikp})}}{n} > 0.5\\
    d, & \text{otherwise}
\end{cases}
\end{equation}
where l is the number of pixels in the backprojected face. 
Thereafter, for every single view $i$, we iterate through all the other views $j$ to decide whether the current specular potential faces in view $i$ is specular or not, based on the thresholded difference of the intensity. As as in in Eq.~\ref{eq:specular_threshold}:
\begin{equation} \label{eq:specular_threshold}
  t(s,d) =  \mu (d) +\phi \mu(s) 
\end{equation}
where $\mu(.)$ is the trimmed mean of the luminance value of common diffuse- and specular-potential faces between the two frames respectively. $\phi$ is a constant that controls the sharpness of the detected specularities (we set $\phi=0.5$ identified through empirical study).
As we iterate through corresponding frames, faces that do not meet the threshold criteria are excluded as potential specular faces $S$ as in Eq.~\ref{eq:excludedecision}: 
 \begin{equation} \label{eq:excludedecision}
  w(x)= 
\begin{cases}
    keep\hspace{0.1cm} in\hspace{0.1cm} S, & \text{if } f_{iz}-f_{jz} \geq t\\
    exclude\hspace{0.1cm} from\hspace{0.1cm} S, & \text{otherwise}
\end{cases}\end{equation}

Finally, our detected specular mask is the remaining faces that are non excluded from potential specular faces $S$. 

\begin{figure*}[t]
\centering
  \includegraphics[width=1\linewidth]{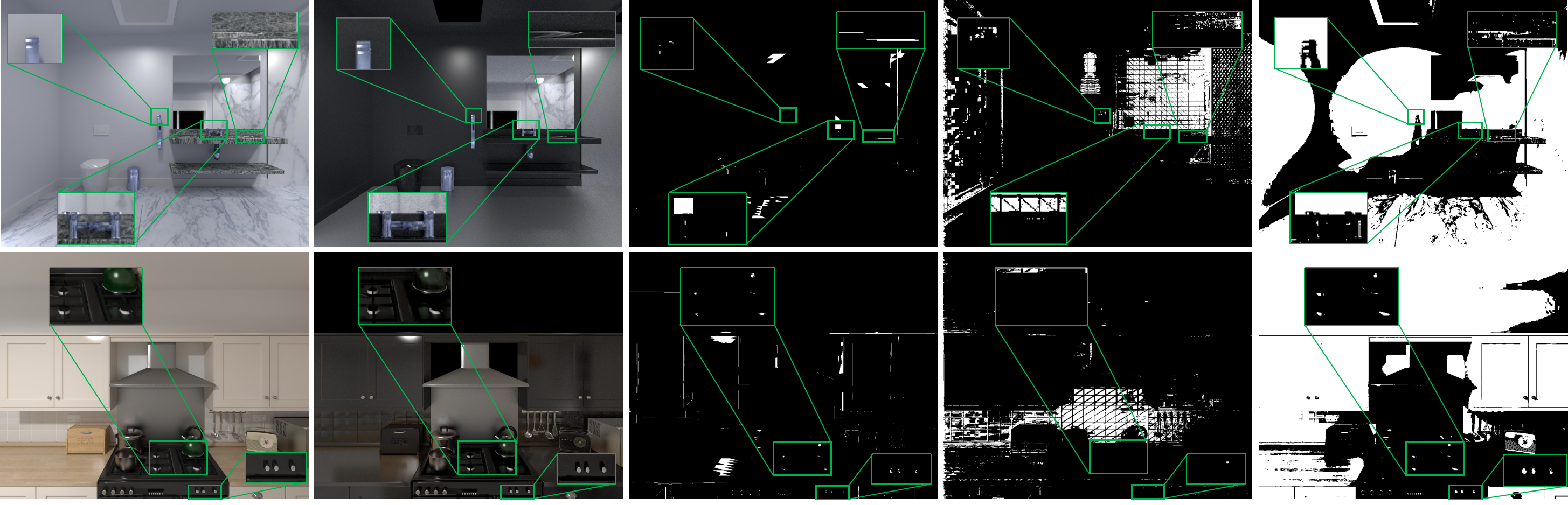}
%   \parbox[t]{\columnwidth}{\relax}
  
  \caption{
     Comparison of specular highlight detection. From left to right: rendered image, groundtruth specular map, our method, Lin et al.~\cite{Lin2002ECCV}, Souza et al.~\cite{Souza2018SIBGRAPI}. The zoomed bounding boxes example exerts from the image to highlight specular areas.
          }
  \label{fig:clear_comparison}
  \end{figure*}
\newpage  
\section{Evaluation} \label{sec:eval}
We evaluate a single-image method~\cite{Souza2018SIBGRAPI}, our multi-view method and~\cite{Lin2002ECCV} on the proposed LIGHTS dataset. 
%Instead of reconstructing the geometry and cameras for ~\cite{Souza2018SIBGRAPI} we use the 
For~\cite{Lin2002ECCV} we use the groundtruth geometry and cameras instead of reconstructing. To evaluate we propose an evaluation metric for specular methods (sec.~\ref{sec:metric}), and the the results in sec.~\ref{sec:results}.
%To be fair, we use the ground-truth geometry and camera information in~\cite{Souza2018SIBGRAPI} instead of reconstructing them. To overcome the difficulty in evaluation of specular methods we propose an evaluation metric in sec.~\ref{sec:metric}, and the results and analysis in sec.~\ref{sec:results}.
\subsection{Evaluation metric} \label{sec:metric}
We threshold the ground truth map using a range of thresholds incrementally, $S_T$ is the range of thresholds, for each threshold the groundtruth specular image $G_m$, is threshold-ed to produce $G_T = G_M > T, \;\;\;\; \forall T \subset  S_T$. 
while for each threshold, we calculate the accuracy between $G_T$ and our estimated mask $M$ using Intersection of union (IoU) normalized by the total number of used threshold ranges, yielding accuracy values as many as threshold values. The final accuracy for this view is calculated based on all calculated IOU accuracy from all threshold values.

%\begin{equation} \label{eq:single_error} A = \frac{\sum{\frac{M \cap G_T}{M \cup G_T}}}{N} , \;\;\;\; \forall T \subset  S_T
%\end{equation} 
At the end, the overall scene accuracy for number of views $A_O$ can be calculated as the average of all computed accuracies across all views.% using eq.~\ref{eq:overall_error}.
While the initial ground-truth specular map is in RGB format, we convert it to CIELAB color space and use the luminance channel $L$ as $G_m$. In our evaluation We set $S_T=[156,\ldots,255]$, as specularities appear in the higher range.
%\begin{equation} \label{eq:overall_error} A_O = \frac{\sum_{m=1}^{L}{A_m}}{L} 
%\end{equation} 

\begin{table}[!hb]
\caption{Accuracy using our metric (see sec.~\ref{sec:metric}) for \cite{Souza2018SIBGRAPI,Lin2002ECCV} and our method over the full (3rd column) and subset (2nd column) of the LIGHTS dataset.}
\label{table:accuracy}
\vspace{-5pt}
\begin{center}
% \resizebox{7cm}{!}{
\begin{tabular}{c|c|c}
\hline
Method & \textbf{$A_O$/\#Views} & \textbf{$A_O$/\#Views} \\ \hline
Single-view~{\cite{Souza2018SIBGRAPI}} & 0.0282/336 & 0.031/2603 \\ \hline
Multi-view~\cite{Lin2002ECCV} & 0.0146/336 & - \\
\textbf{Multi-view (Ours)} & 0.0502/336 & 0.04/2603 \\ \hline
\end{tabular}
% }
\end{center}
\vspace{-10pt}
\end{table}

\subsection{Results and analysis}\label{sec:results}
%We evaluate against~\cite{Lin2002ECCV} based on both speed and detection accuracy while with~\cite{Souza2018SIBGRAPI} only on the latter.
%We evaluated the processing time on all scenes images %using Matlab code 
%both by taking into account the total amount of views per scene as well as only one view for a more fair comparison (see Table 1 in suppl. material). 
%This can be justified due to the face-based vs pixel-based approach that we follow.
%Moreover, our solution compares the whole image at once, while \cite{Lin2002ECCV} is iterating row by row.
We compare the methods in Table~\ref{table:accuracy} showing the accuracy of both multi-view techniques alongside the single image detector technique (\cite{Souza2018SIBGRAPI}) which was a pre-processing step in our pipeline. Due to the processing time needed for~\cite{Lin2002ECCV} we created two ablation studies showing the result of the three algorithms on 336 views (column 1) which is $25\times$  more evaluation images compared to other previous works, and on all of our dataset views which is $2603$ views in total (column 2). Our algorithm achieved higher accuracy than~\cite{Lin2002ECCV} and~\cite{Souza2018SIBGRAPI} by 3.5\% and 1.3\% respectively. Our approach also proved to be faster w.r.t. Lin et al~\cite{Lin2002ECCV} by approx. $4$k times in average (See supp. mat. for full performance comparison). 
We also show a qualitative comparison in Fig.~\ref{fig:clear_comparison} against both~\cite{Lin2002ECCV, Souza2018SIBGRAPI} where it can be seen that our approach outperforms the other approaches, especially on white objects or saturated areas which can be easily confused for specularities.
The use of faces instead of pixels in our framework has two advantages, 1) it is more robust to noise relying on the consensus across the face, and 2) using faces capitalizes on speed compared to processing individual pixels.

% %%%%%%%%%%%%%%%%%%%%%%%%%%%%%%%%%%%%%%%%%%%%%%%%%%%%

% \resizebox{6cm}{!}{

%   }

% \begin{figure}
%   \centering
%   \includegraphics[width=\linewidth]{IMG/comparison.jpg}
%   \parbox[t]{\columnwidth}{\relax}
%   \caption{
%      From left to right: rendered image, specular map, our estimation, Lin et al. \cite{Lin2002ECCV} estimation, single view estimation (Souza et al\cite{Souza2018SIBGRAPI}).
%           }
%           \label{fig:comparison}
%   \end{figure}
% %%%%%%%%%%%%%%%%%%%%%%%%%%%%%%%%%%%%%%%%%%%%%%%%%%%%
\section{Conclusion}\label{sec:conclusion}
In this paper we presented the LIGHTS dataset, which simulates real light transportation, different material and object properties as well as a wide variation of light setups.
We have shown that our dataset provides a chellange to existing methods and the proposed pipeline is able to distinguish between specular and non-specular regions that can be confusing for other methods especially for the single-image methods. % By aggregating information from multi-view images.
% The proposed method widens the future research of specular removal to deep models that can be trained on pre-processed multi-view natural images common during reconstruction capture.
Despite that our methods accuracy is limited by the size and number of faces in the scene mesh, it overcomes and performs faster than the pixel-based methods.
% While we found that the traditionally used sets of evaluation images were not reflecting the real performance of detection algorithms especially on the natural images, we presented a physically based rendering light analysis dataset, 
% possible to be used for future light analysis evaluations.
% It contains wide variations of light setup, materials, and objects.
% In future work the method can be applied to large-scale multi-view datasets (e.g. Matterport3D and Scannet) or incorporating semantic information to train deep models to overcome confusion between specularities and similar effects on single images.\\


\bibliographystyle{IEEEbib}
% \bibliography{spec}
{
% \footnotesize
\bibliography{spec}}
\clearpage
\appendix

%
% \begin{abstract}
\section{Supplementary Material}
\noindent In this document we present the following supplementary material:
\begin{itemize}
    \item[-] Extra samples and structural information of the dataset.
    \item[-] Additional results over those presented in the main paper.
\end{itemize}

\begin{figure}[H]
    \centering
    \includegraphics[width=\linewidth]{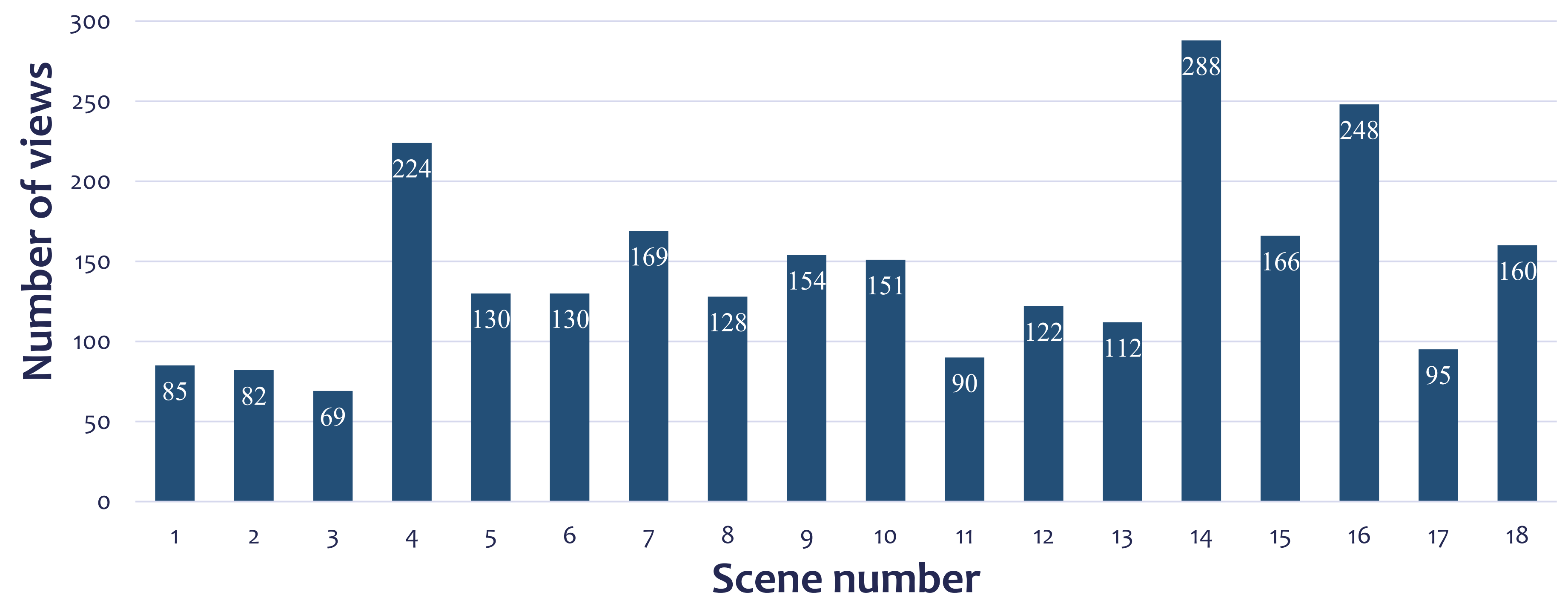}
    \caption{Number of rendered views at each scene}
    \label{fig:view_dist}
\end{figure}

First we try to illustrate a more complete overview of the whole dataset scenes. In Figure~\ref{fig:view_dist} we show the total amount of views rendered and used for each one, this varies from 69 to 288 which were randomly generated for a more realistic representation.

% additional results over those presented in the main paper, we show the performance of single image techniques on both lab-constrained and non lab-constrained images and side-by-side comparison of our proposed multi-view technique, Lin et al. \cite{Lin2002ECCV} estimation, and single view estimation of Souza et al\cite{Souza2018SIBGRAPI}. Moreover, we show a speed comparison of our face-based technique against pixel-based technique of  Lin et al. \cite{Lin2002ECCV} on LAD. We also show a sample of each scene from our presented LIGHT dataset.

% \end{abstract}
%
% \begin{keywords}
% Specular-highlights, Multi-view, Dataset, Face-based detection
% \end{keywords}
%
% \section{Dataset}\label{sec:dataset}

As we have mentioned in the main paper, each scene comes with a set of different maps which can be used for light analysis. Thus, the scenes and the corresponding extracted maps can be seen in figures 5-22.

Figure~\ref{fig:SingleImageRemovalDrawbacks} demonstrates the simplicity of the current lab-constrained state-of-the-art datasets for specular highlight detection (i.e. first three rows) and a comparison between different single image specular highlight detector and sample scenes from the LIGHTS dataset where they fail to address.

Figures 3 and 4, similarly to figure 2 in the main paper, showcase the qualitative performance of our multi-view specular detection pipeline for more scene samples of the LIGHTS dataset side-by-side with the estimation from Lin et al.~\cite{Lin2002ECCV} and of Souza et al.~\cite{Souza2018SIBGRAPI}. The sharp cuts on some of the detections are due to the face/patch segmentation of the scene each time, meaning that we a more dense patch segmentation these would be more smooth. As it can be seen in the majority of the scenes our pipeline seems to correctly focus on the patches where the specular components are contained while the other two methods usually provide a more broad specular detection as output.

Finally, table~\ref{table:timing_statistics} reports the speed comparison of our face-based multi-view detector against pixel-based technique of  Lin et al.~\cite{Lin2002ECCV} on the presented views of LIGHTS dataset for each of the 18 scenes. Considering that the method from Li is really time consuming having a comparison to all views per scene was not really feasible (columns 3 and 4). Thus, for a more fair comparison we applied the same time evaluation only on one frame from both our and Li's solutions (columns 5 and 6) and which shows us that our techniques is much faster from Li's by a big margin. As we describe in the main paper this is due to the face-based vs pixel-based approach that we follow as well as to the fact that our pipeline process the frames all at once overcoming the row-by-row approach in Li's work.

% \section{Performance}\label{sec:performance}
\begin{table}[ht]
\caption{Processing time (in sec) of the two compared techniques per number of views used. In the last row, the average time consumed for all the scenes per method.}
\label{table:timing_statistics}
\begin{center}
\resizebox{8.3cm}{!}{
\begin{tabular}{|c|c|c|c|c|c|}

\hline
\textbf{Scene ID} & \textbf{Total views} & \textbf{Ours (Time/\#views)} & { Lin~\cite{Lin2002ECCV}} \textbf{(Time/\#views)} & \textbf{Ours (Time/view)} & { Lin~\cite{Lin2002ECCV}} \textbf{(Time/view)} \\ \hline
\textbf{1} & 85 & 347/85 & 1982k/85 & 4.02/1 & 23k/1 \\ \hline
\textbf{2} & 82 & 177/82 & 1549k/82 & 2.15/1 & 18.9k/1 \\ \hline
\textbf{3} & 69 & 104/69 & 1234k/69 & 1.49/1 & 17.8k/1 \\ \hline
\textbf{4} & 224 & 377/224 & 993k/5 & 155/1 & 198k/1 \\ \hline
\textbf{5} & 130 & 504/130 & 493k/19 & 3.88/1 & 26k/1 \\ \hline
\textbf{6} & 130 & 674/130 & 162k/6 & 5.19/1 & 27k/1 \\ \hline
\textbf{7} & 169 & 1.2k/169 & 28k/1 & 7.31/1 & 28k/1 \\ \hline
\textbf{8} & 128 & 541/128 & 157k/10 & 4.23/1 & 15.7k/1 \\ \hline
\textbf{9} & 154 & 161/154 & 248k/10 & 1.05/1 & 24.8k/1 \\ \hline
\textbf{10} & 151 & 851/151 & 242k/10 & 5.64/1 & 24.3k/1 \\ \hline
\textbf{11} & 90 & 263/90 & 158k/11 & 2.92/1 & 14.4k/1 \\ \hline
\textbf{12} & 122 & 480/122 & 45.3k/4 & 3.93/1 & 11.3k/1 \\ \hline
\textbf{13} & 112 & 442/112 & 259k/10 & 3.95/1 & 25.9k/1 \\ \hline
\textbf{14} & 288 & 3k/288 & 220k/1 & 10.5/1 & 220k/1  \\ \hline
\textbf{15} & 166 & 647/166 & 28.4k/1 & 3.9/1 & 28.4k/1 \\ \hline
\textbf{16} & 248 & 1.95k/248 & 206k/1 & 7.88/1 & 206k/1 \\ \hline
\textbf{17} & 95 & 258/95 & 173k/10 & 2.72/1 & 17.3k/1 \\ \hline
\textbf{18} & 160 & 418/160 & 28k/1 & 2.61/1 & 28k/1 \\ \hline
{\textbf{Avg.}}&{145}&{687}& {456K} & 12.69 & 53.04k \\ \hline
\end{tabular}
}
\end{center}
\vspace{-7pt}
\end{table}

\begin{figure*}[!hb]
  \centering
%   \vspace{-450pt}
  \includegraphics[width=1\textwidth]{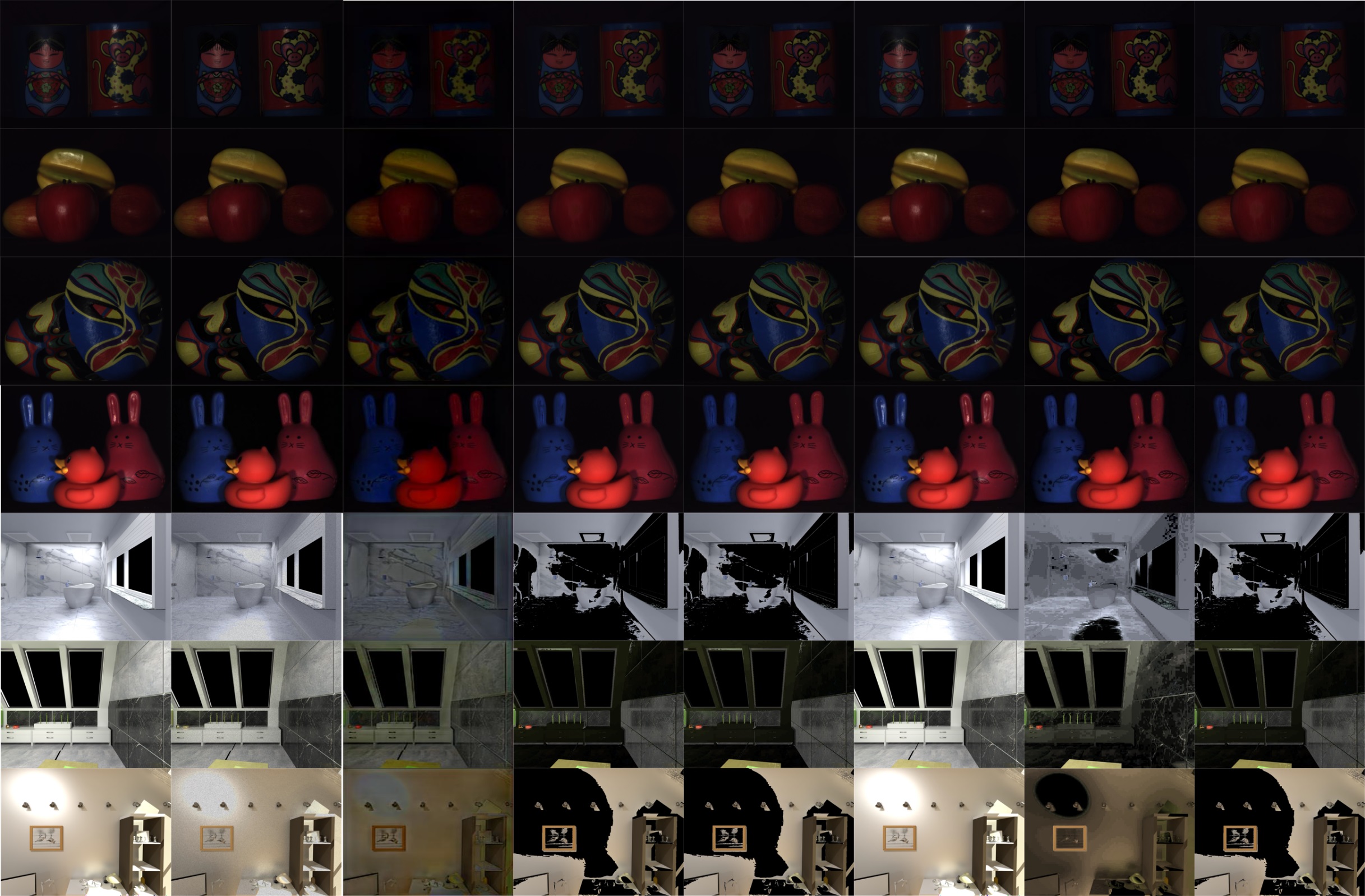}
%   \parbox[t]{\columnwidth}{\relax}
  \caption{
     From left to right: Original image, Akashi et al~\cite{AKASHI2016CVIU}, Lin et al~\cite{lin2019deep}, Souza et al\cite{Souza2018SIBGRAPI}, Shen et al~\cite{Shen2013ApplOpt}, Tan et al~\cite{Tan2005PAMI}, Yang et al~\cite{Yang2010SBH}, Yamamato et al~\cite{TakahisaYamamoto2019ITE}. Performance of single images specular high-light detection and removal algorithms on conventional test images (rows 1, 2) which we refer to it as lab-constrained images, and on normal images (rows 3, 4) from LIGHTS, it is clear that they are performing good on the conventionally used images but they have poor performance on natural (non-lab-constrained) images.}
          \label{fig:SingleImageRemovalDrawbacks}
  \end{figure*}

{\noindent\textbf{Acknowledgments:} The authors would like to list the following blender community users for publicly providing access to the initial CAD models used in the dataset and which we further modified. These are Nacimus, Cenobi
, DonCliche
, Acid\_Jazz
, SlykDrako
, Jay-Artist
, TheCGNinja
, MatteSr
, mikenx
, Ariadrianto
, Hilux
, ORBANGeoffrey
, Wig42
, Vladoffsky
, and Ermmus.
}

\begin{figure*}[!ht]
  \centering
  \includegraphics[width=\linewidth]{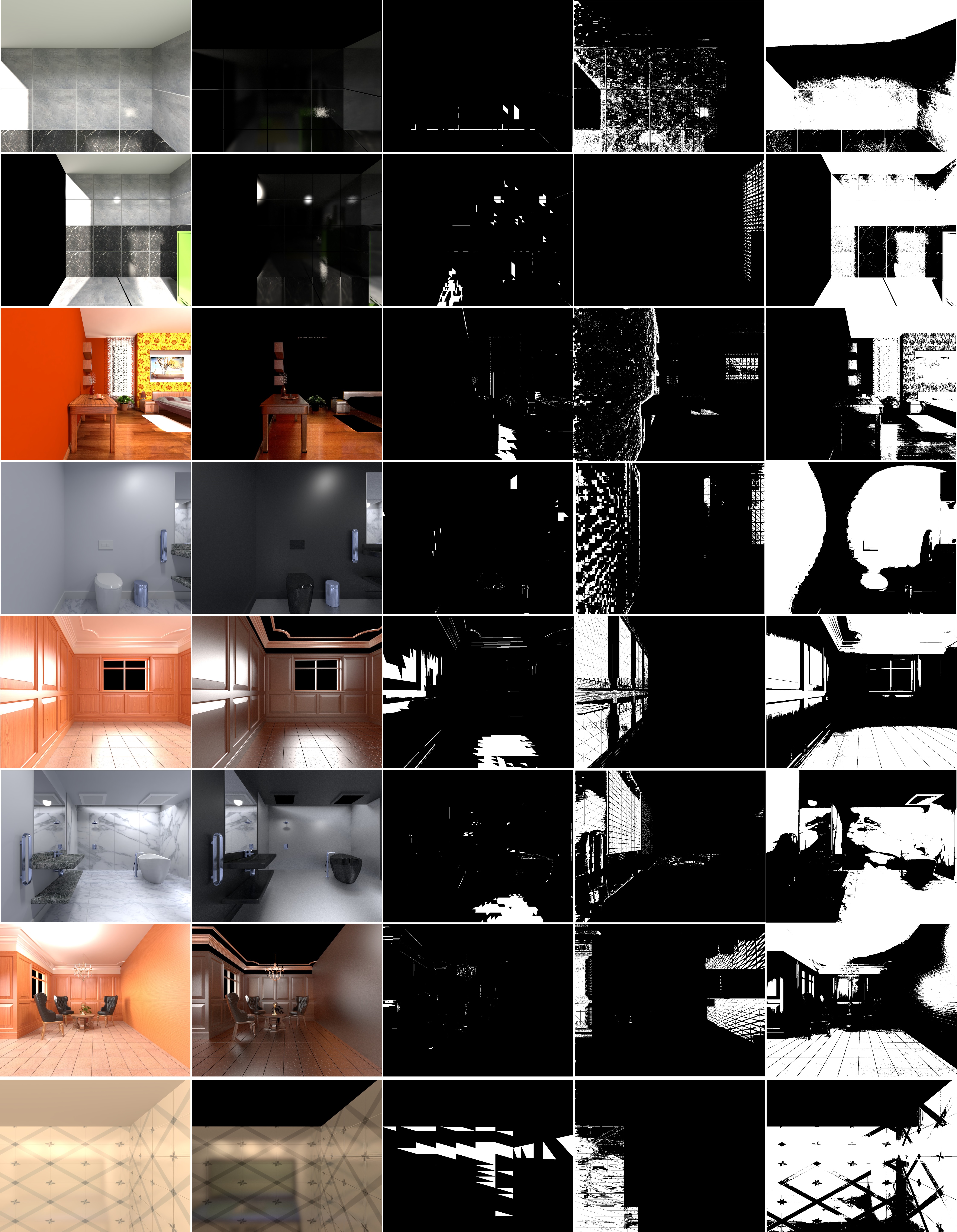}
  \parbox[t]{\columnwidth}{\relax}
  \caption{From left to right: rendered image, specular map, our estimation, Lin et al. \cite{Lin2002ECCV} estimation, and our dependant single view estimation (Souza et al\cite{Souza2018SIBGRAPI}).}
          \label{fig:SingleImageRemovalDrawbacks}
  \end{figure*}
\begin{figure*}[!ht]
  \centering
  \includegraphics[width=\linewidth]{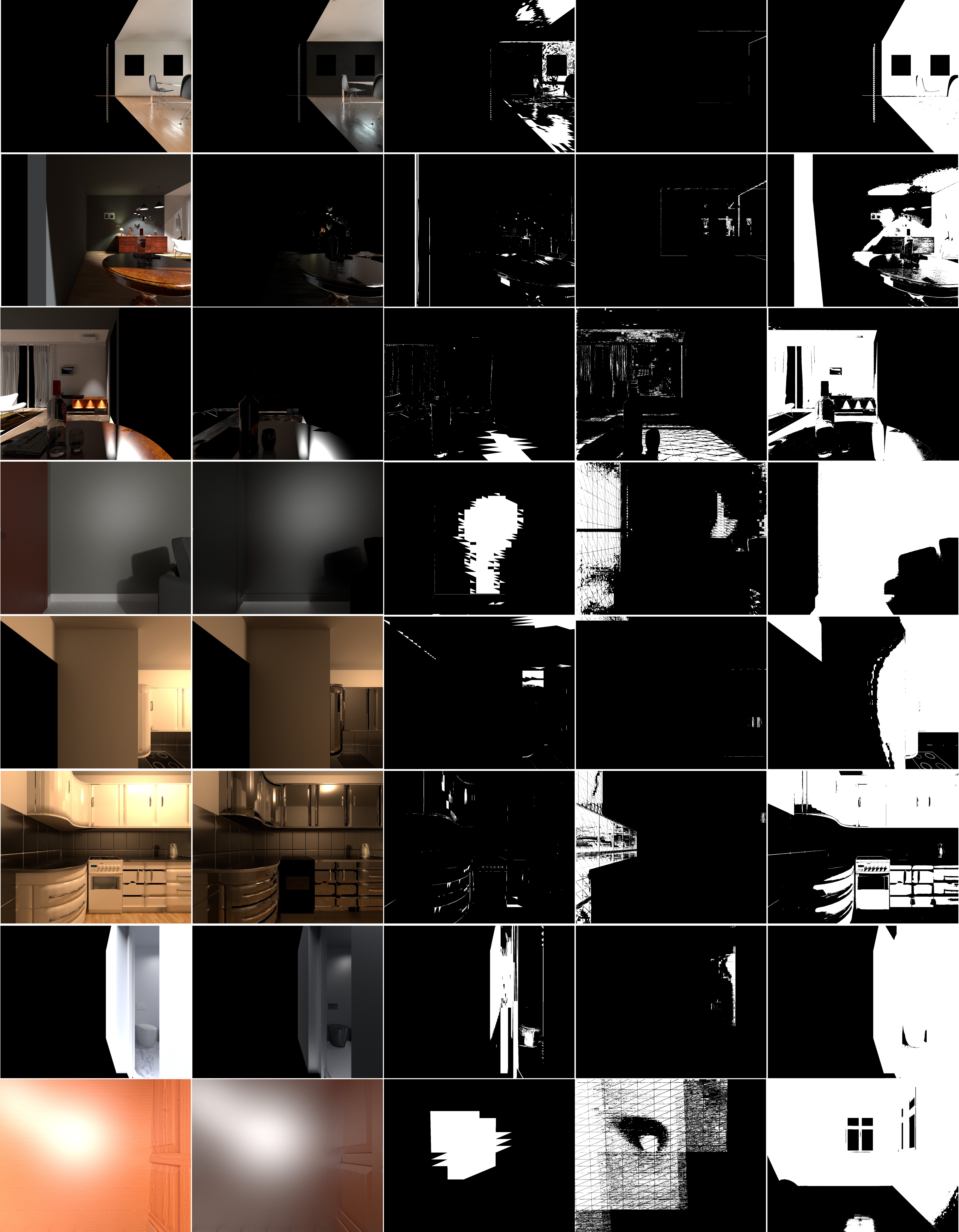}
  \parbox[t]{\columnwidth}{\relax}
  \caption{From left to right: rendered image, specular map, our estimation, Lin et al. \cite{Lin2002ECCV} estimation, and our dependant single view estimation (Souza et al\cite{Souza2018SIBGRAPI}).}
          \label{fig:SingleImageRemovalDrawbacks}
  \end{figure*}

\begin{figure}[!ht]
     \centering
     \begin{subfigure}[b]{0.49\linewidth}
         \centering
         \includegraphics[width=1\linewidth]{IMG/supp_mat/supp_mat_1/Image.jpg}
         \caption{Rendered}
         \label{fig:y equals x}
     \end{subfigure}
     \begin{subfigure}[b]{0.49\linewidth}
         \centering
         \includegraphics[width=1\linewidth]{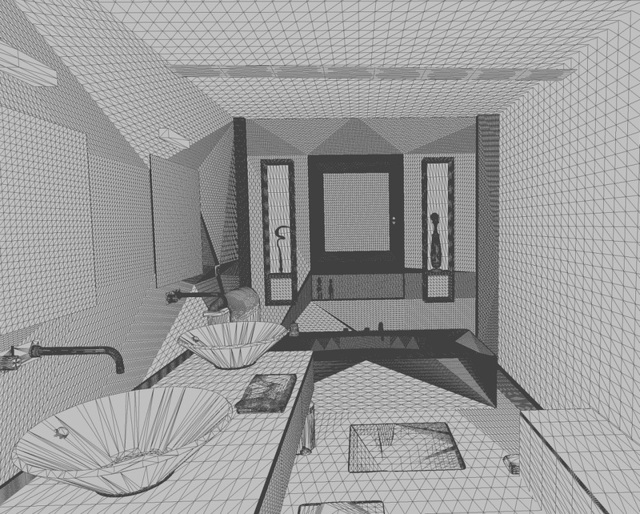}
         \caption{Triangulate mesh}
         \label{fig:y equals x}
     \end{subfigure}
     \hfill
     \begin{subfigure}[b]{0.32\linewidth}
         \centering
         \includegraphics[width=\linewidth]{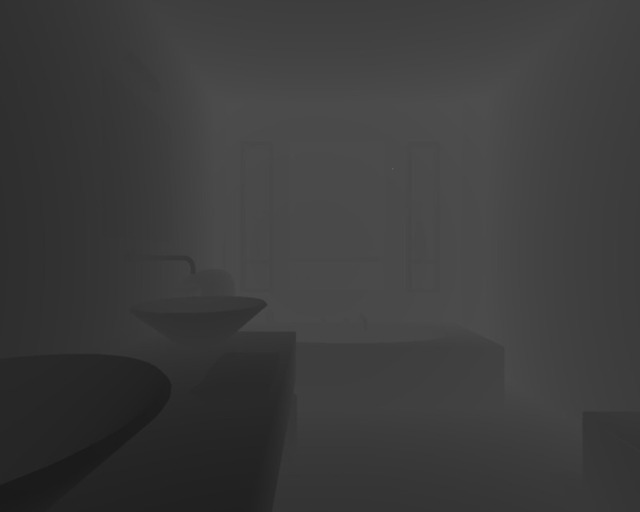}
         \caption{Depth}
         \label{fig:three sin x}
     \end{subfigure}
     \hfill
     \begin{subfigure}[b]{0.32\linewidth}
         \centering
         \includegraphics[width=\linewidth]{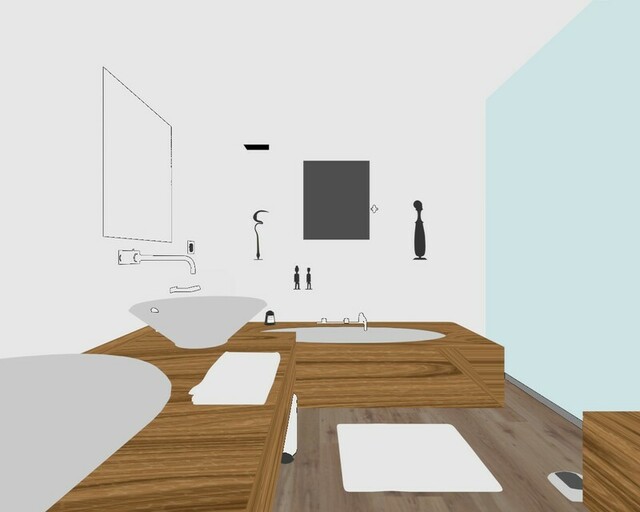}
         \caption{Albedo}
         \label{fig:five over x}
     \end{subfigure}
      \hfill
     \begin{subfigure}[b]{0.32\linewidth}
         \centering
         \includegraphics[width=\linewidth]{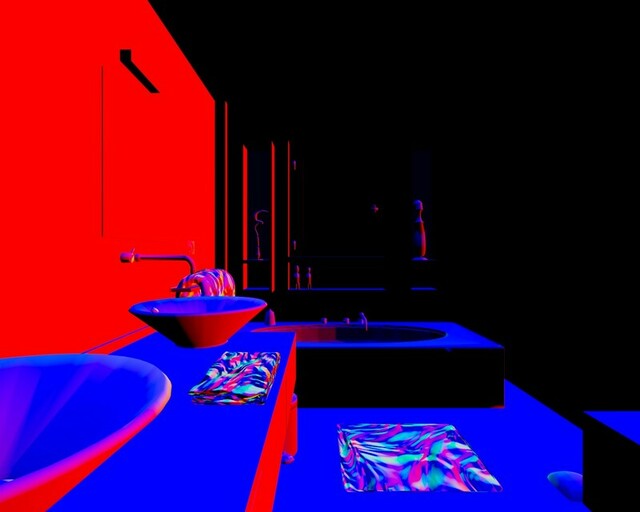}
         \caption{Normals}
         \label{fig:five over x}
     \end{subfigure}
     \begin{subfigure}[b]{0.32\linewidth}
         \centering
         \includegraphics[width=\linewidth]{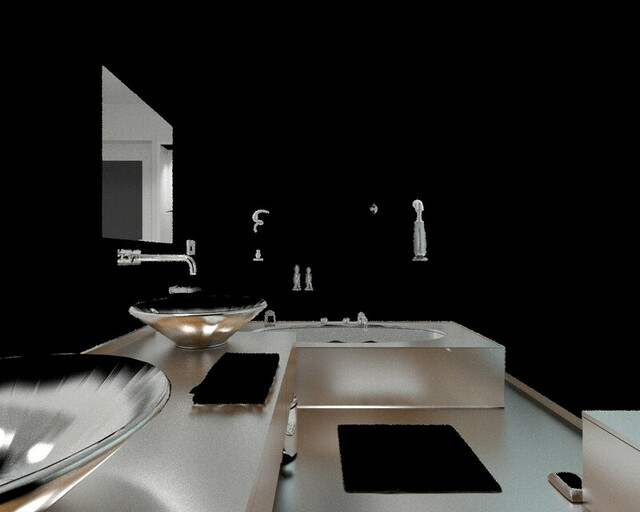}
         \caption{Specular Indirect}
         \label{fig:five over x}
     \end{subfigure}
     \begin{subfigure}[b]{0.32\linewidth}
         \centering
         \includegraphics[width=\linewidth]{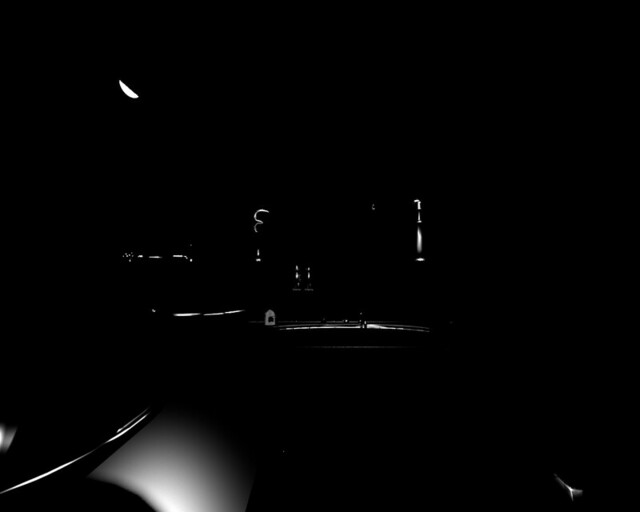}
         \caption{Specular Direct}
         \label{fig:five over x}
     \end{subfigure}
     \begin{subfigure}[b]{0.32\linewidth}
         \centering
         \includegraphics[width=\linewidth]{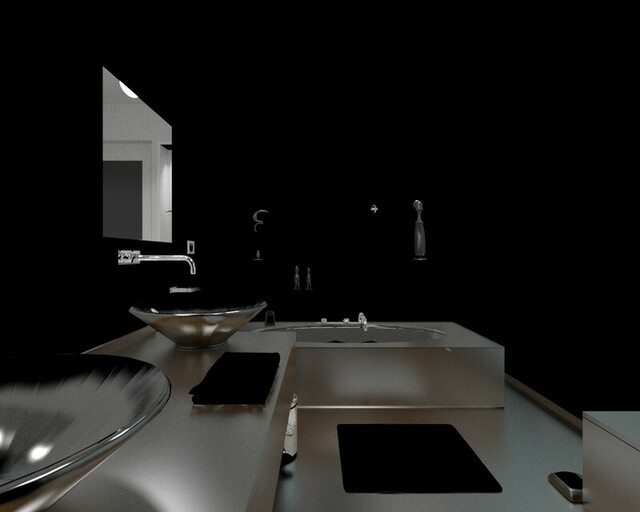}
         \caption{Specular}
         \label{fig:five over x}
     \end{subfigure}
     \begin{subfigure}[b]{0.32\linewidth}
         \centering
         \includegraphics[width=\linewidth]{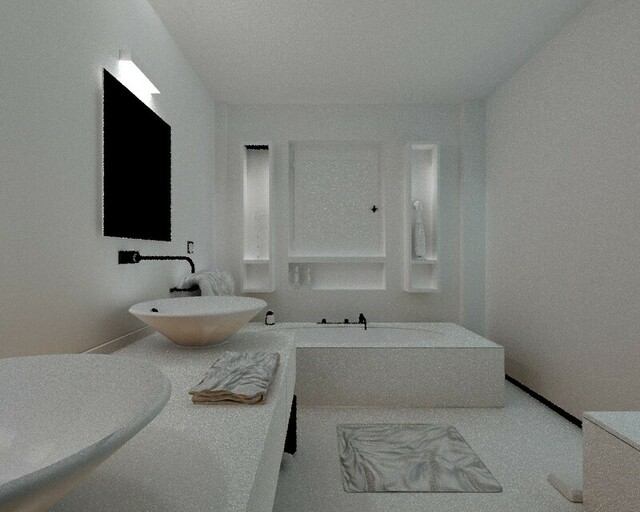}
         \caption{Diffuse Indirect}
         \label{fig:five over x}
     \end{subfigure}
     \begin{subfigure}[b]{0.32\linewidth}
         \centering
         \includegraphics[width=\linewidth]{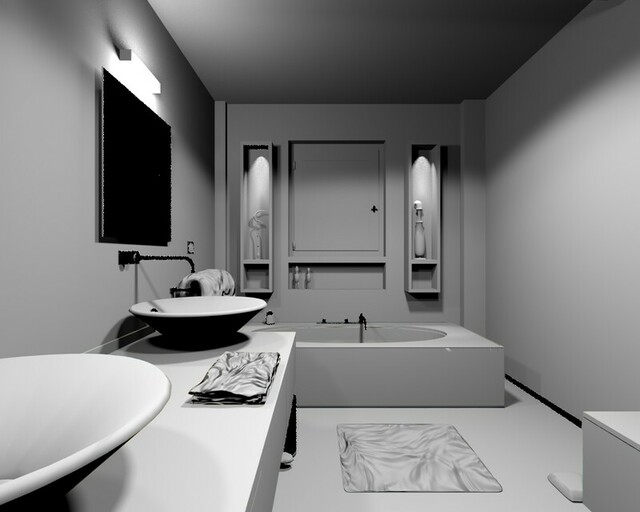}
         \caption{Diffuse Direct}
         \label{fig:five over x}
     \end{subfigure}
     \begin{subfigure}[b]{0.32\linewidth}
         \centering
         \includegraphics[width=\linewidth]{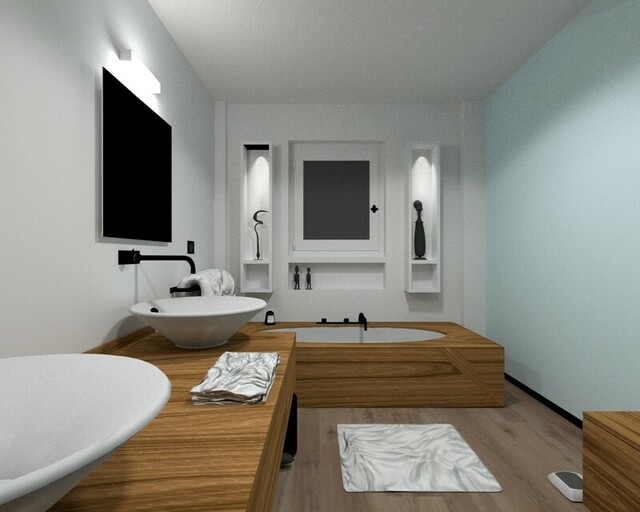}
         \caption{Diffuse}
         \label{fig:five over x}
     \end{subfigure}
     \begin{subfigure}[b]{0.32\linewidth}
         \centering
         \includegraphics[width=\linewidth]{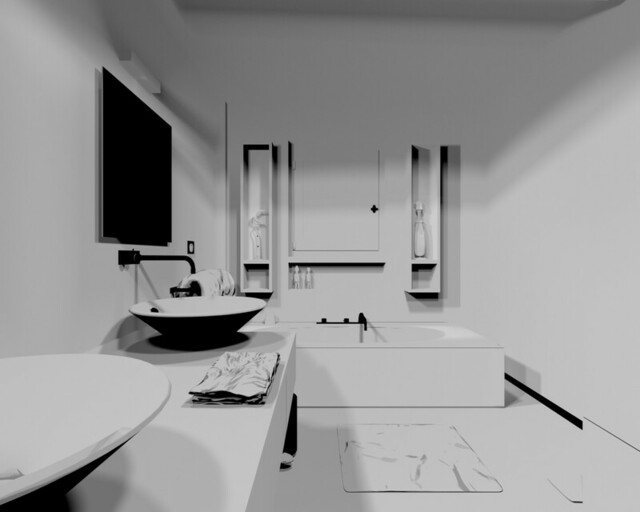}
         \caption{Shadow}
         \label{fig:five over x}
     \end{subfigure}
     \begin{subfigure}[b]{0.32\linewidth}
         \centering
         \includegraphics[width=\linewidth]{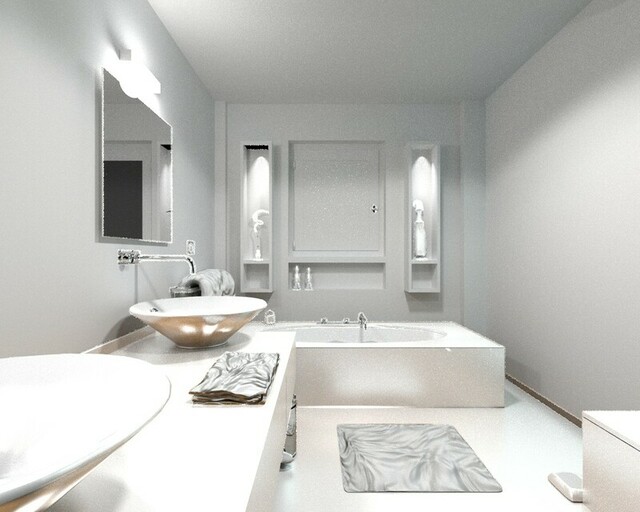}
         \caption{Lightmap}
         \label{fig:five over x}
     \end{subfigure}
     \begin{subfigure}[b]{0.32\linewidth}
         \centering
         \includegraphics[width=\linewidth]{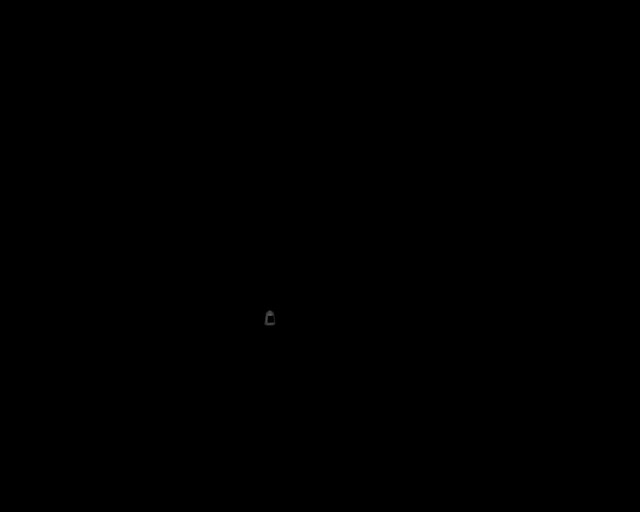}
         \caption{Transmission}
         \label{fig:five over x}
     \end{subfigure}
        \caption{Sample from scene 1}
        \label{fig:three graphs}
\end{figure}

\begin{figure}[!ht]
     \centering
     \begin{subfigure}[b]{0.49\linewidth}
         \centering
         \includegraphics[width=1\linewidth]{IMG/supp_mat/supp_mat_2/Image.jpg}
         \caption{Rendered Image}
         \label{fig:y equals x}
     \end{subfigure}
    \begin{subfigure}[b]{0.49\linewidth}
         \centering
         \includegraphics[width=1\linewidth]{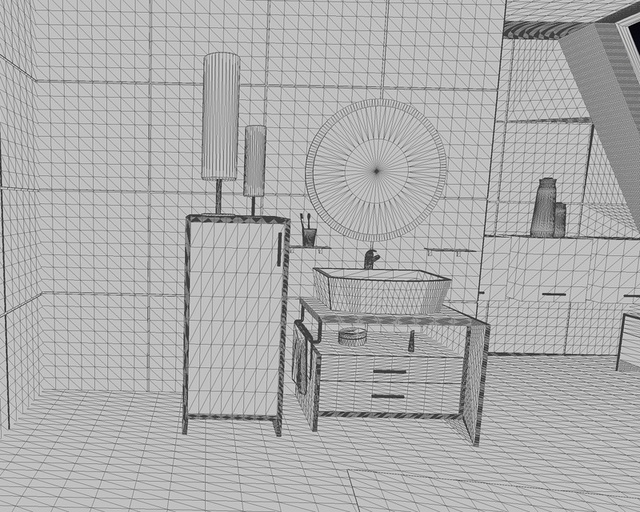}
         \caption{Triangulate mesh}
         \label{fig:y equals x}
     \end{subfigure}
     \hfill
     \begin{subfigure}[b]{0.32\linewidth}
         \centering
         \includegraphics[width=\linewidth]{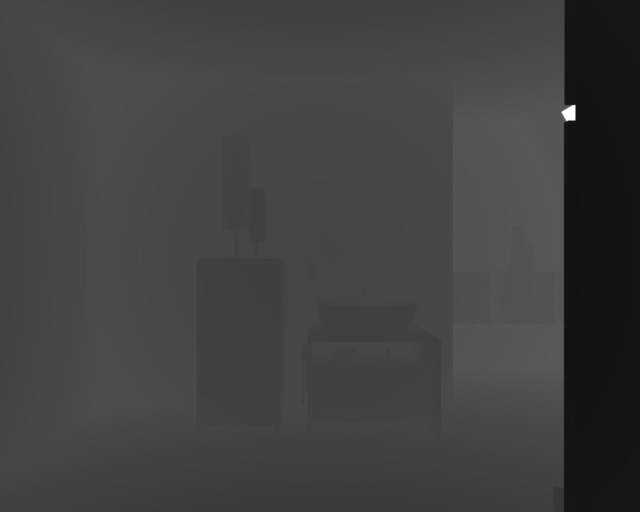}
         \caption{Depth}
         \label{fig:three sin x}
     \end{subfigure}
     \hfill
     \begin{subfigure}[b]{0.32\linewidth}
         \centering
         \includegraphics[width=\linewidth]{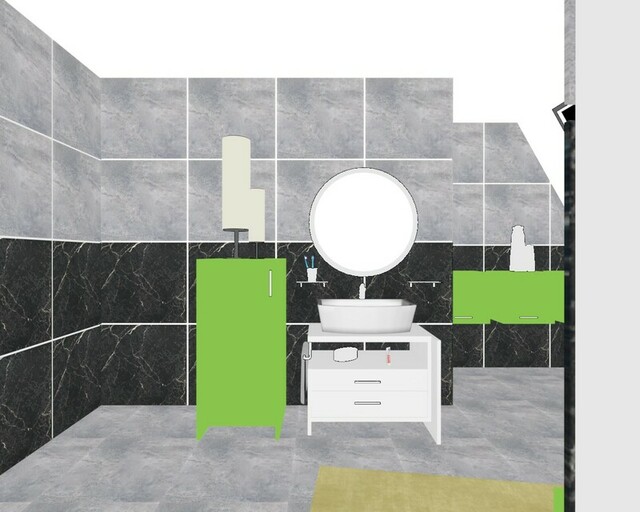}
         \caption{Albedo}
         \label{fig:five over x}
     \end{subfigure}
      \hfill
     \begin{subfigure}[b]{0.32\linewidth}
         \centering
         \includegraphics[width=\linewidth]{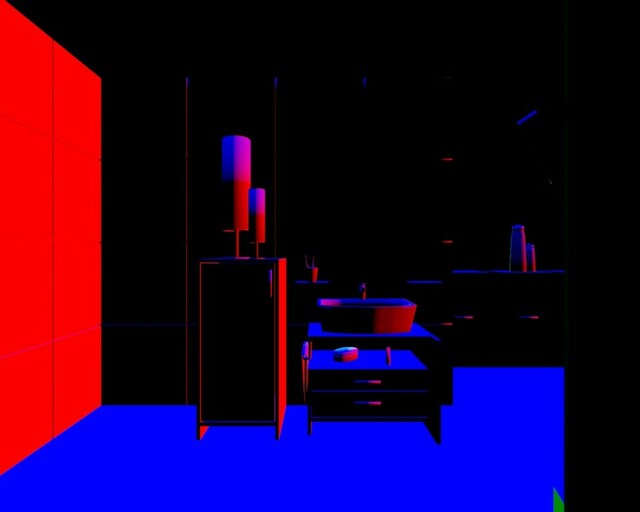}
         \caption{Normals}
         \label{fig:five over x}
     \end{subfigure}
     \begin{subfigure}[b]{0.32\linewidth}
         \centering
         \includegraphics[width=\linewidth]{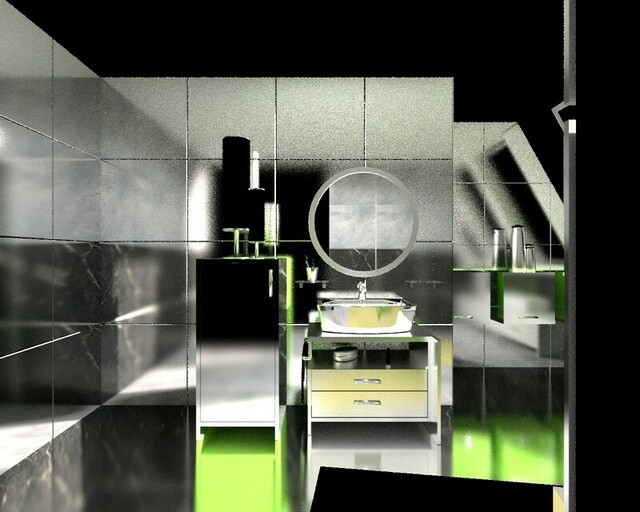}
         \caption{Specular Indirect}
         \label{fig:five over x}
     \end{subfigure}
     \begin{subfigure}[b]{0.32\linewidth}
         \centering
         \includegraphics[width=\linewidth]{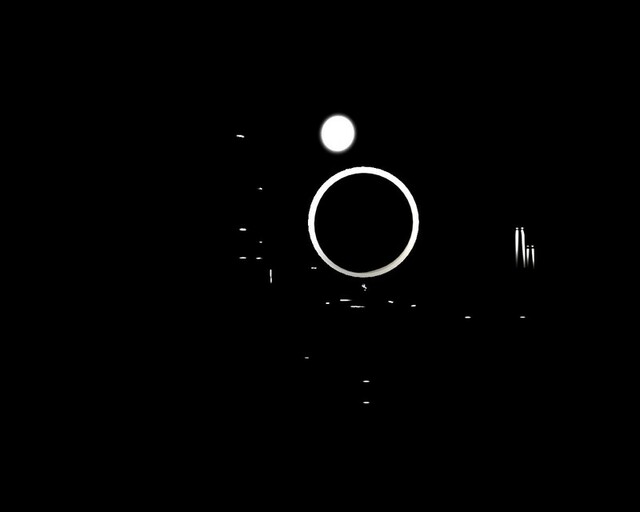}
         \caption{Specular Direct}
         \label{fig:five over x}
     \end{subfigure}
     \begin{subfigure}[b]{0.32\linewidth}
         \centering
         \includegraphics[width=\linewidth]{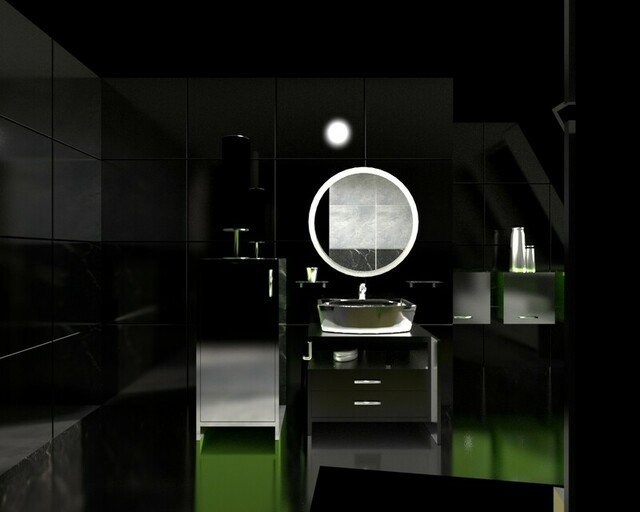}
         \caption{Specular}
         \label{fig:five over x}
     \end{subfigure}
     \begin{subfigure}[b]{0.32\linewidth}
         \centering
         \includegraphics[width=\linewidth]{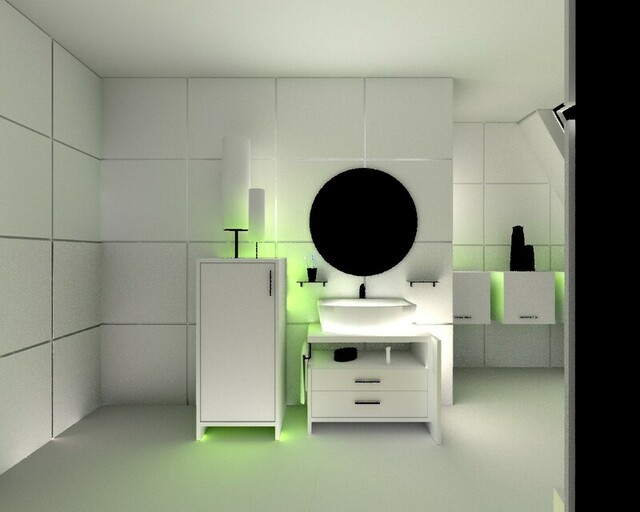}
         \caption{Diffuse Indirect}
         \label{fig:five over x}
     \end{subfigure}
     \begin{subfigure}[b]{0.32\linewidth}
         \centering
         \includegraphics[width=\linewidth]{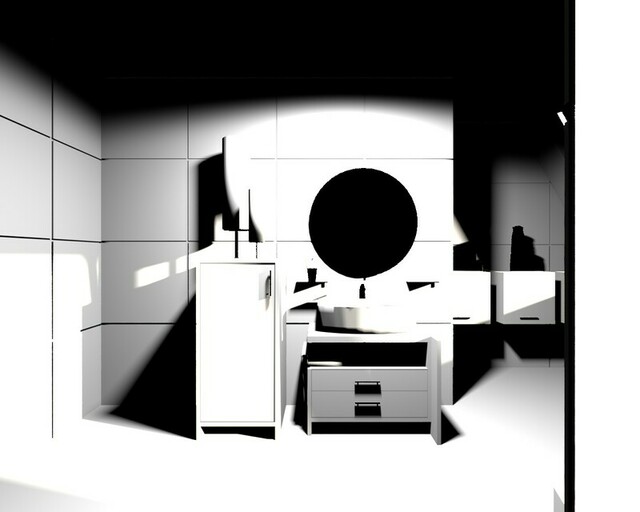}
         \caption{Diffuse Direct}
         \label{fig:five over x}
     \end{subfigure}
     \begin{subfigure}[b]{0.32\linewidth}
         \centering
         \includegraphics[width=\linewidth]{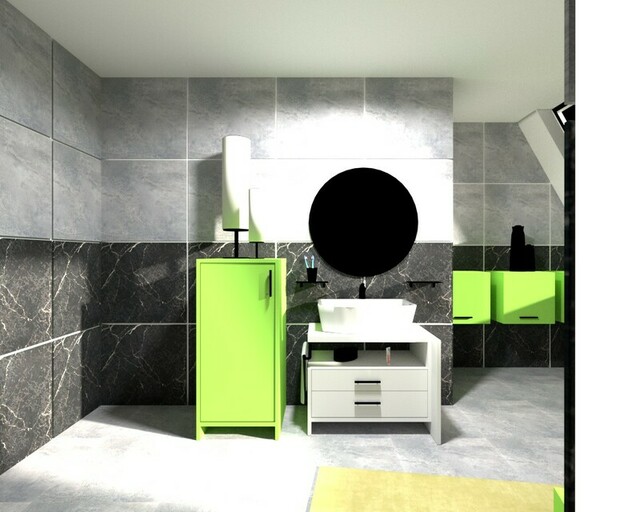}
         \caption{Diffuse}
         \label{fig:five over x}
     \end{subfigure}
     \begin{subfigure}[b]{0.32\linewidth}
         \centering
         \includegraphics[width=\linewidth]{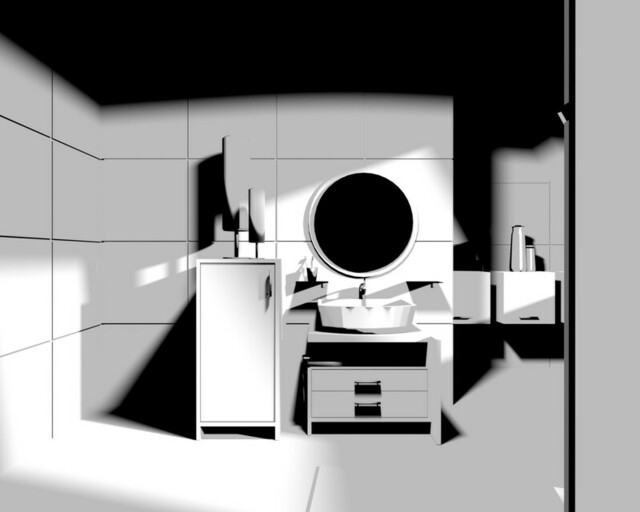}
         \caption{Shadow}
         \label{fig:five over x}
     \end{subfigure}
     \begin{subfigure}[b]{0.32\linewidth}
         \centering
         \includegraphics[width=\linewidth]{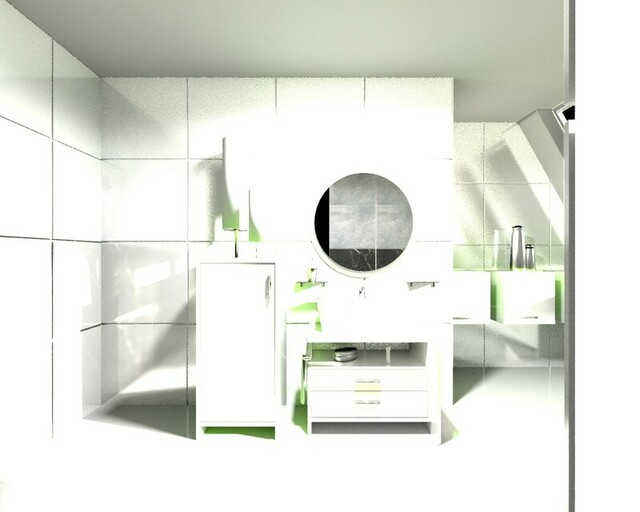}
         \caption{Lightmap}
         \label{fig:five over x}
     \end{subfigure}
     \begin{subfigure}[b]{0.32\linewidth}
         \centering
         \includegraphics[width=\linewidth]{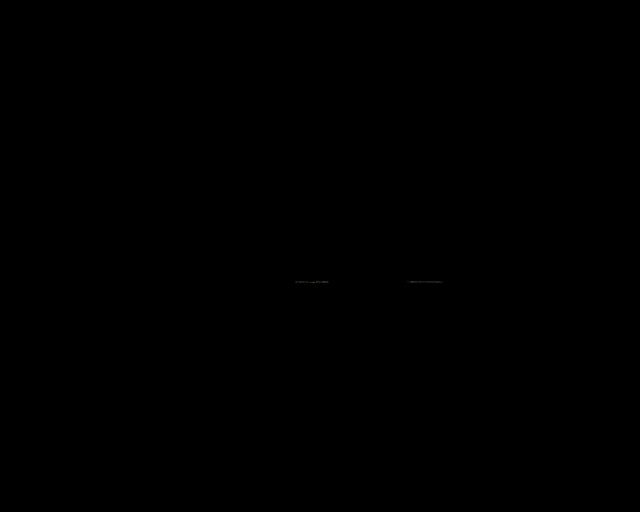}
         \caption{Transmission}
         \label{fig:five over x}
     \end{subfigure}
        \caption{Sample from scene 2}
        \label{fig:three graphs}
\end{figure}

\begin{figure}[!ht]
     \centering
     \begin{subfigure}[b]{0.49\linewidth}
         \centering
         \includegraphics[width=1\linewidth]{IMG/supp_mat/supp_mat_3/Image.jpg}
         \caption{Rendered Image}
         \label{fig:y equals x}
     \end{subfigure}
     \begin{subfigure}[b]{0.49\linewidth}
         \centering
         \includegraphics[width=1\linewidth]{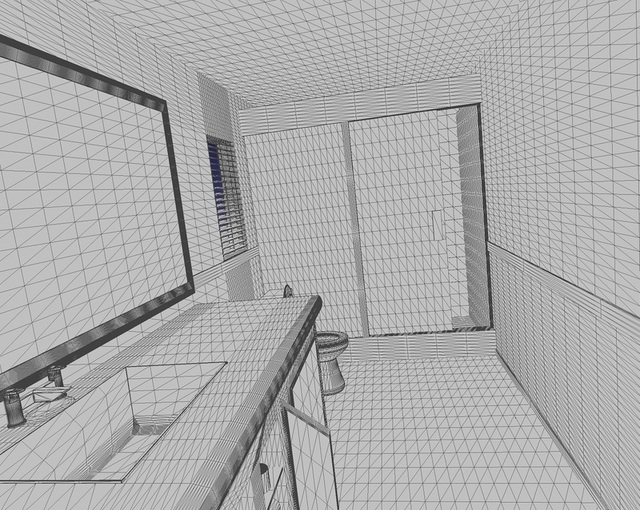}
         \caption{Triangulate mesh}
         \label{fig:y equals x}
     \end{subfigure}
     \hfill
     \begin{subfigure}[b]{0.32\linewidth}
         \centering
         \includegraphics[width=\linewidth]{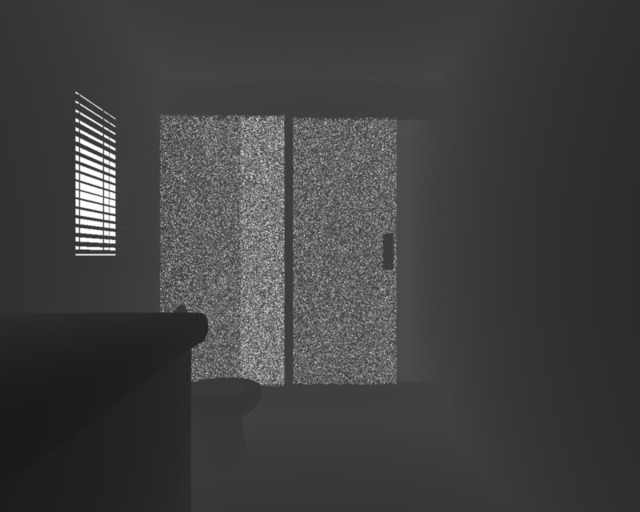}
         \caption{Depth}
         \label{fig:three sin x}
     \end{subfigure}
     \hfill
     \begin{subfigure}[b]{0.32\linewidth}
         \centering
         \includegraphics[width=\linewidth]{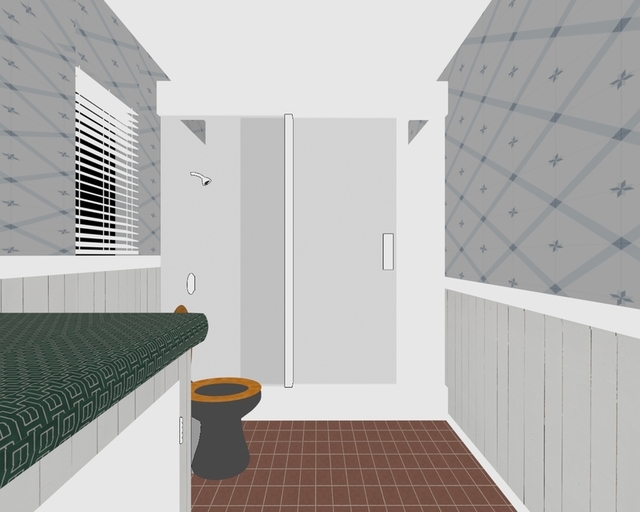}
         \caption{Albedo}
         \label{fig:five over x}
     \end{subfigure}
      \hfill
     \begin{subfigure}[b]{0.32\linewidth}
         \centering
         \includegraphics[width=\linewidth]{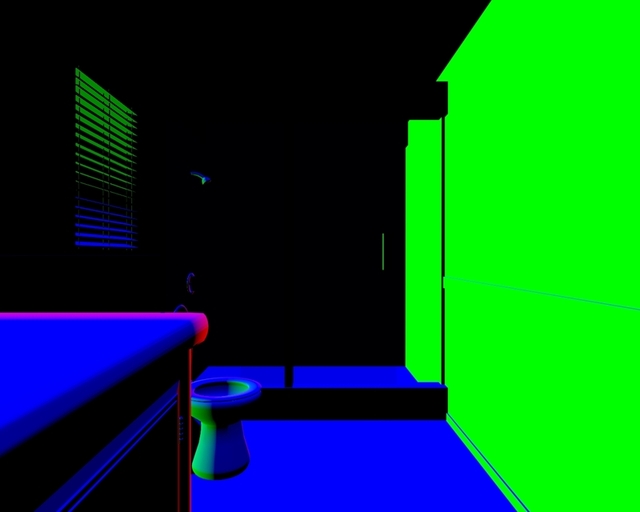}
         \caption{Normals}
         \label{fig:five over x}
     \end{subfigure}
     \begin{subfigure}[b]{0.32\linewidth}
         \centering
         \includegraphics[width=\linewidth]{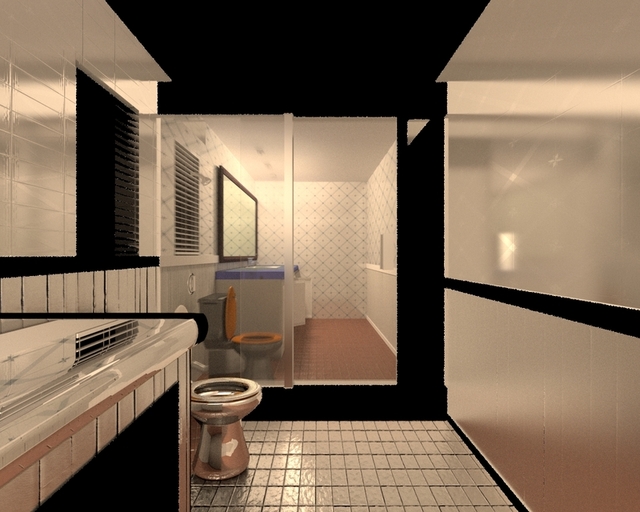}
         \caption{Specular Indirect}
         \label{fig:five over x}
     \end{subfigure}
     \begin{subfigure}[b]{0.32\linewidth}
         \centering
         \includegraphics[width=\linewidth]{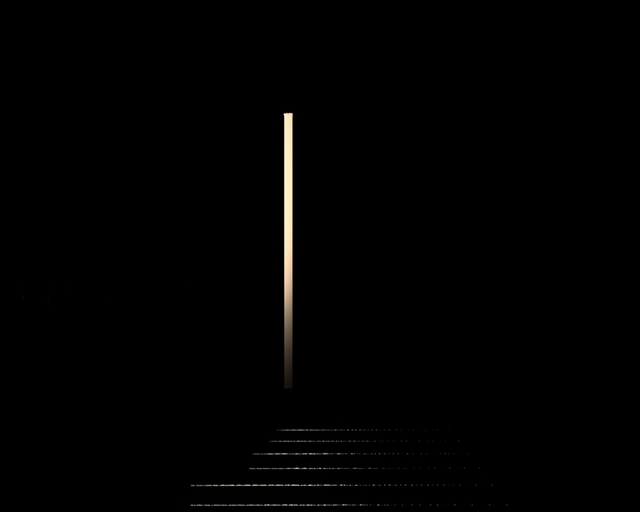}
         \caption{Specular Direct}
         \label{fig:five over x}
     \end{subfigure}
     \begin{subfigure}[b]{0.32\linewidth}
         \centering
         \includegraphics[width=\linewidth]{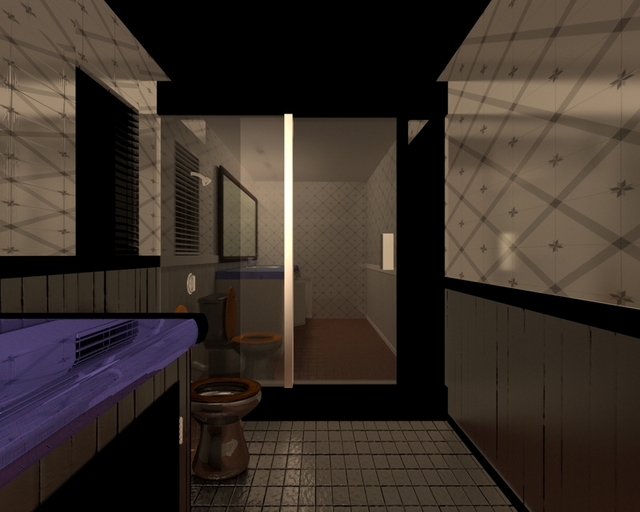}
         \caption{Specular}
         \label{fig:five over x}
     \end{subfigure}
     \begin{subfigure}[b]{0.32\linewidth}
         \centering
         \includegraphics[width=\linewidth]{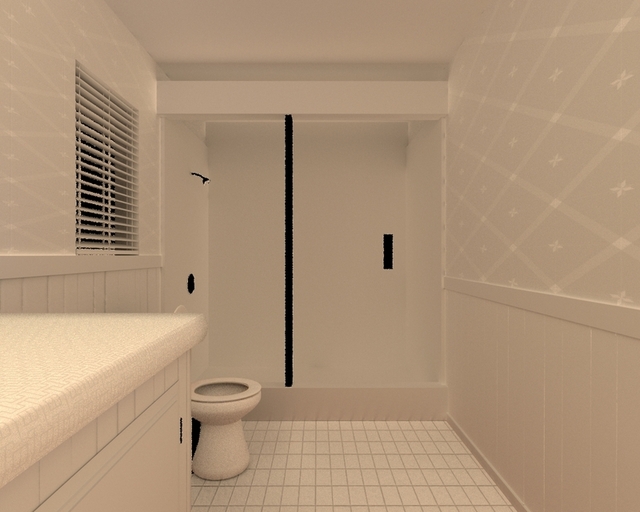}
         \caption{Diffuse Indirect}
         \label{fig:five over x}
     \end{subfigure}
     \begin{subfigure}[b]{0.32\linewidth}
         \centering
         \includegraphics[width=\linewidth]{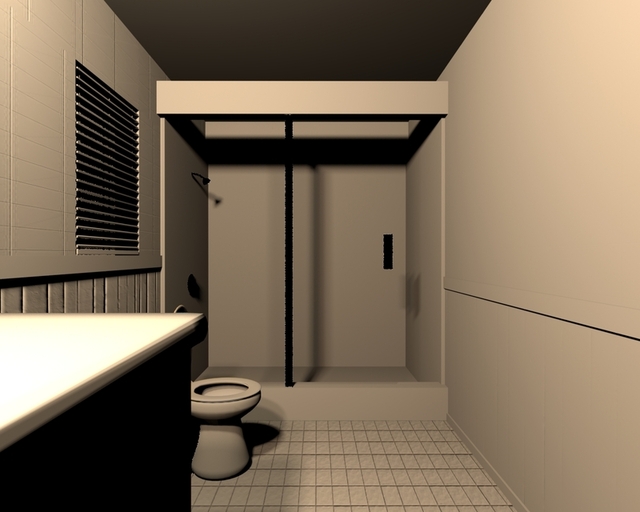}
         \caption{Diffuse Direct}
         \label{fig:five over x}
     \end{subfigure}
     \begin{subfigure}[b]{0.32\linewidth}
         \centering
         \includegraphics[width=\linewidth]{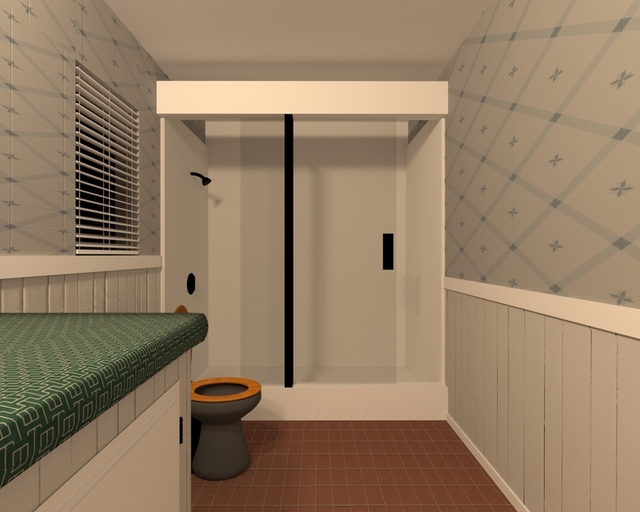}
         \caption{Diffuse}
         \label{fig:five over x}
     \end{subfigure}
     \begin{subfigure}[b]{0.32\linewidth}
         \centering
         \includegraphics[width=\linewidth]{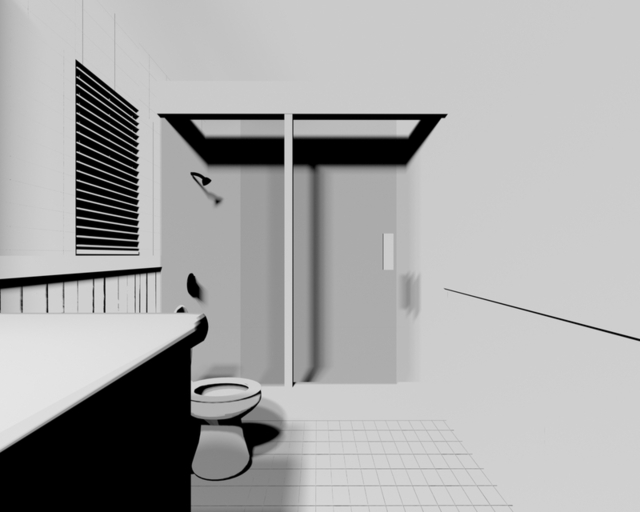}
         \caption{Shadow}
         \label{fig:five over x}
     \end{subfigure}
     \begin{subfigure}[b]{0.32\linewidth}
         \centering
         \includegraphics[width=\linewidth]{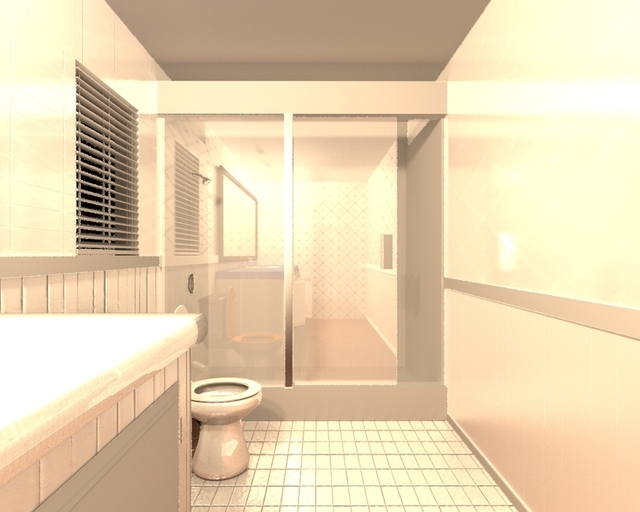}
         \caption{Lightmap}
         \label{fig:five over x}
     \end{subfigure}
     \begin{subfigure}[b]{0.32\linewidth}
         \centering
         \includegraphics[width=\linewidth]{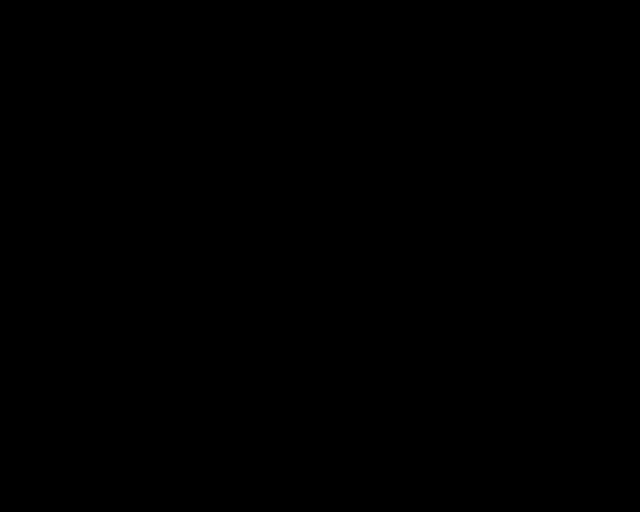}
         \caption{Transmission}
         \label{fig:five over x}
     \end{subfigure}
        \caption{Sample from scene 3}
        \label{fig:three graphs}
\end{figure}

\begin{figure}[!ht]
     \centering
     \begin{subfigure}[b]{0.49\linewidth}
         \centering
         \includegraphics[width=1\linewidth]{IMG/supp_mat/supp_mat_4/Image.jpg}
         \caption{Rendered Image}
         \label{fig:y equals x}
     \end{subfigure}
     \begin{subfigure}[b]{0.49\linewidth}
         \centering
         \includegraphics[width=1\linewidth]{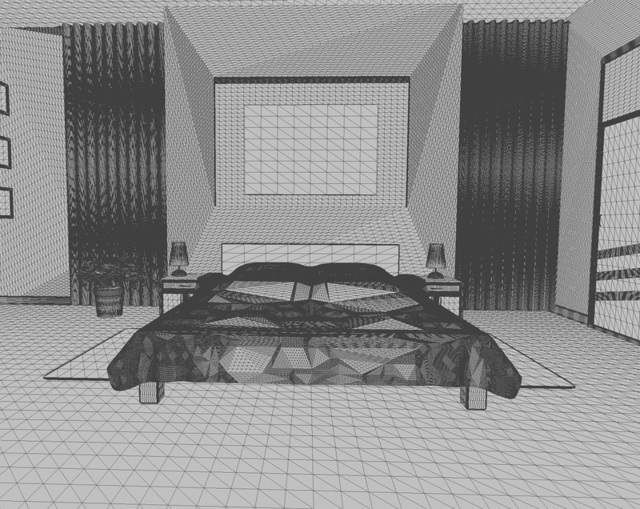}
         \caption{Triangulate mesh}
         \label{fig:y equals x}
     \end{subfigure}
     \hfill
     \begin{subfigure}[b]{0.32\linewidth}
         \centering
         \includegraphics[width=\linewidth]{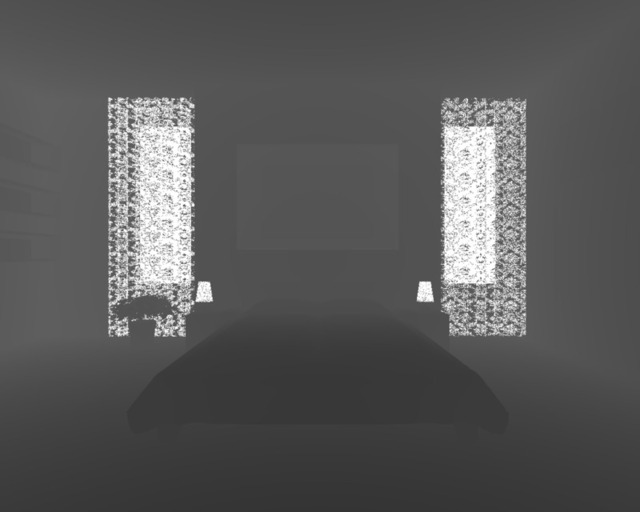}
         \caption{Depth}
         \label{fig:three sin x}
     \end{subfigure}
     \hfill
     \begin{subfigure}[b]{0.32\linewidth}
         \centering
         \includegraphics[width=\linewidth]{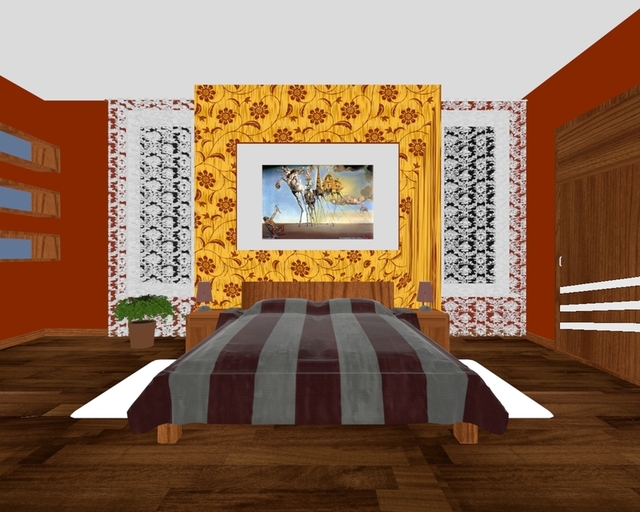}
         \caption{Albedo}
         \label{fig:five over x}
     \end{subfigure}
      \hfill
     \begin{subfigure}[b]{0.32\linewidth}
         \centering
         \includegraphics[width=\linewidth]{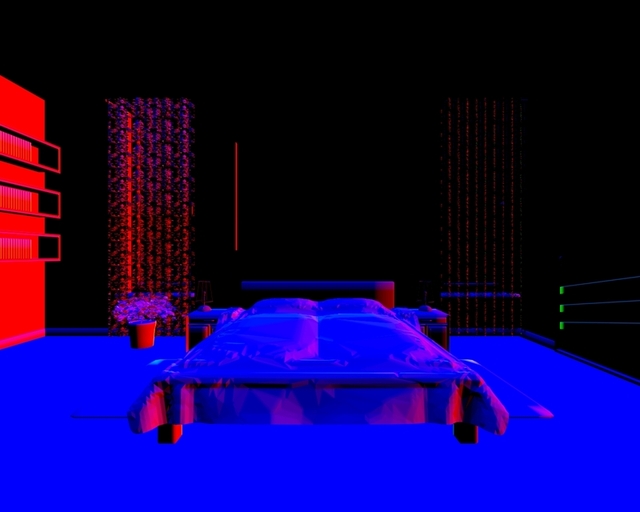}
         \caption{Normals}
         \label{fig:five over x}
     \end{subfigure}
     \begin{subfigure}[b]{0.32\linewidth}
         \centering
         \includegraphics[width=\linewidth]{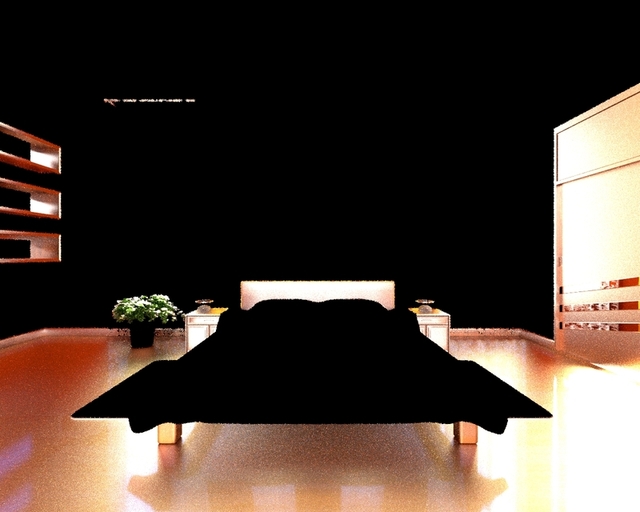}
         \caption{Specular Indirect}
         \label{fig:five over x}
     \end{subfigure}
     \begin{subfigure}[b]{0.32\linewidth}
         \centering
         \includegraphics[width=\linewidth]{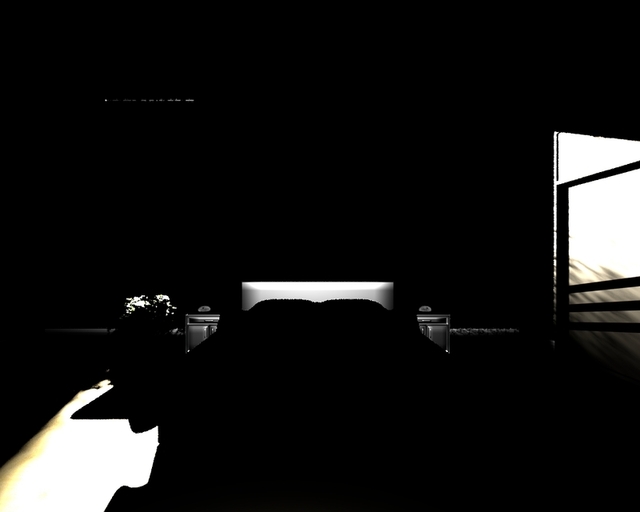}
         \caption{Specular Direct}
         \label{fig:five over x}
     \end{subfigure}
     \begin{subfigure}[b]{0.32\linewidth}
         \centering
         \includegraphics[width=\linewidth]{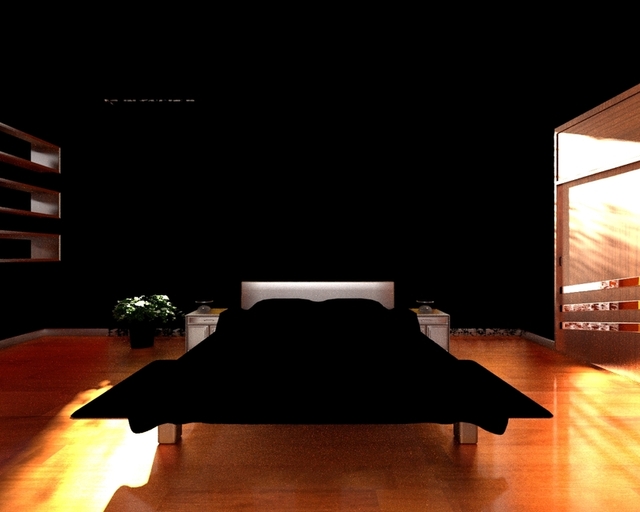}
         \caption{Specular}
         \label{fig:five over x}
     \end{subfigure}
     \begin{subfigure}[b]{0.32\linewidth}
         \centering
         \includegraphics[width=\linewidth]{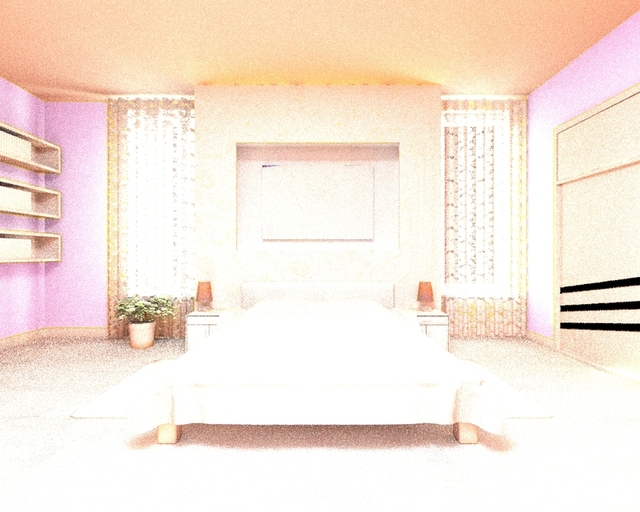}
         \caption{Diffuse Indirect}
         \label{fig:five over x}
     \end{subfigure}
     \begin{subfigure}[b]{0.32\linewidth}
         \centering
         \includegraphics[width=\linewidth]{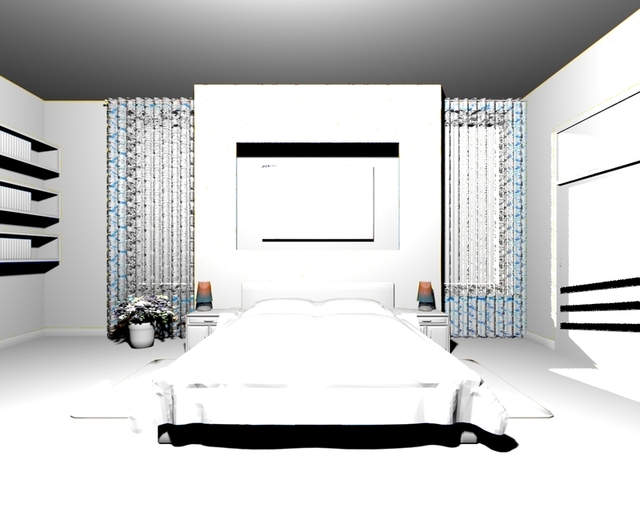}
         \caption{Diffuse Direct}
         \label{fig:five over x}
     \end{subfigure}
     \begin{subfigure}[b]{0.32\linewidth}
         \centering
         \includegraphics[width=\linewidth]{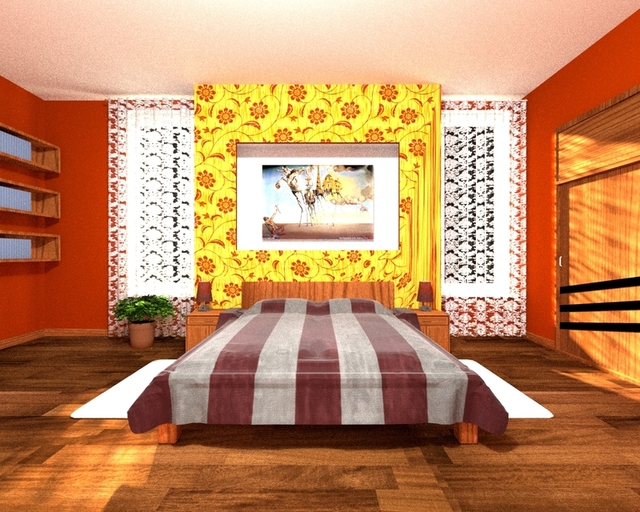}
         \caption{Diffuse}
         \label{fig:five over x}
     \end{subfigure}
     \begin{subfigure}[b]{0.32\linewidth}
         \centering
         \includegraphics[width=\linewidth]{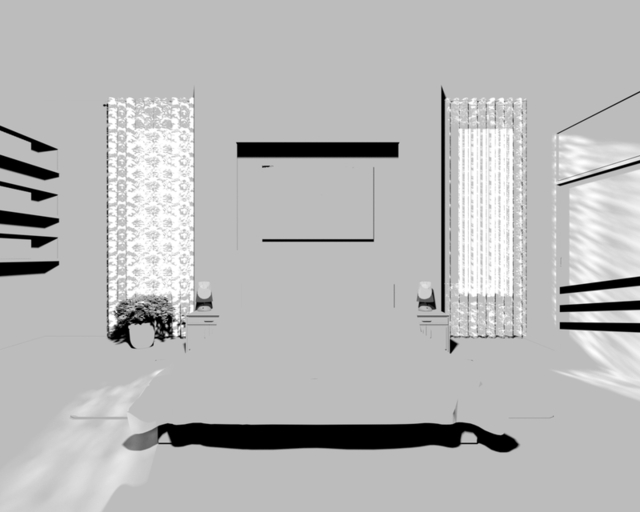}
         \caption{Shadow}
         \label{fig:five over x}
     \end{subfigure}
     \begin{subfigure}[b]{0.32\linewidth}
         \centering
         \includegraphics[width=\linewidth]{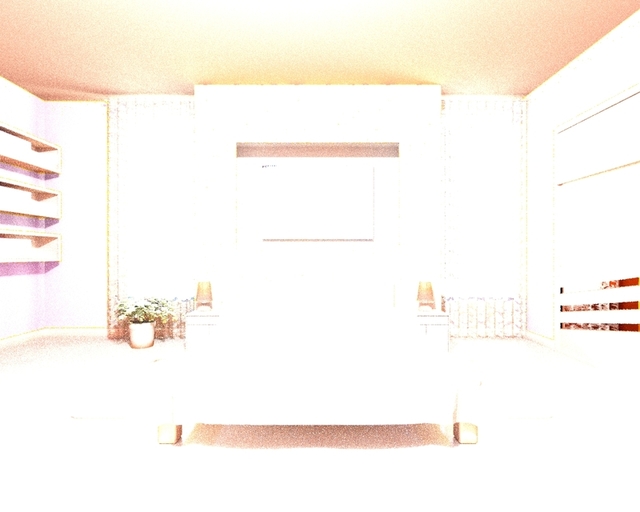}
         \caption{Lightmap}
         \label{fig:five over x}
     \end{subfigure}
     \begin{subfigure}[b]{0.32\linewidth}
         \centering
         \includegraphics[width=\linewidth]{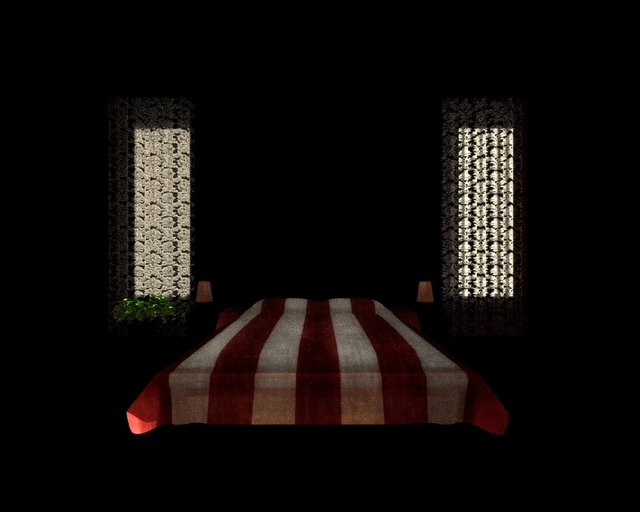}
         \caption{Transmission}
         \label{fig:five over x}
     \end{subfigure}
        \caption{Sample from scene 4}
        \label{fig:three graphs}
\end{figure}

\begin{figure}[!ht]
     \centering
     \begin{subfigure}[b]{0.49\linewidth}
         \centering
         \includegraphics[width=1\linewidth]{IMG/supp_mat/supp_mat_5/Image.jpg}
         \caption{Rendered Image}
         \label{fig:y equals x}
     \end{subfigure}
     \begin{subfigure}[b]{0.49\linewidth}
         \centering
         \includegraphics[width=1\linewidth]{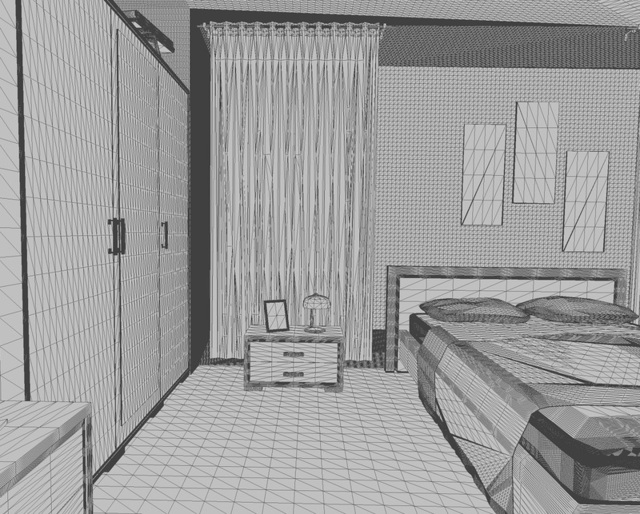}
         \caption{Triangulate mesh}
         \label{fig:y equals x}
     \end{subfigure}
     \hfill
     \begin{subfigure}[b]{0.32\linewidth}
         \centering
         \includegraphics[width=\linewidth]{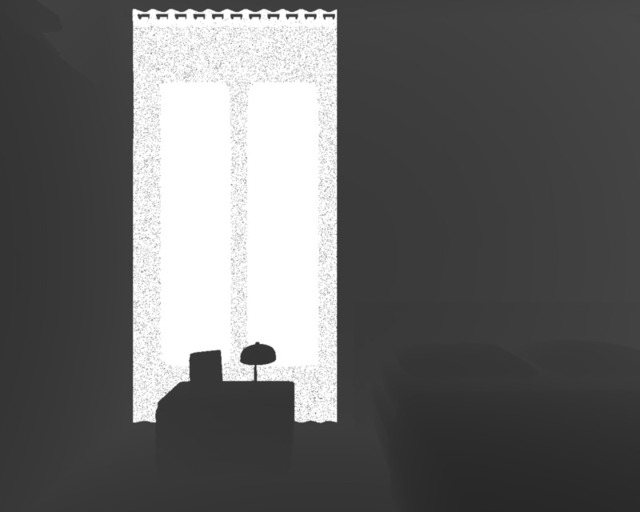}
         \caption{Depth}
         \label{fig:three sin x}
     \end{subfigure}
     \hfill
     \begin{subfigure}[b]{0.32\linewidth}
         \centering
         \includegraphics[width=\linewidth]{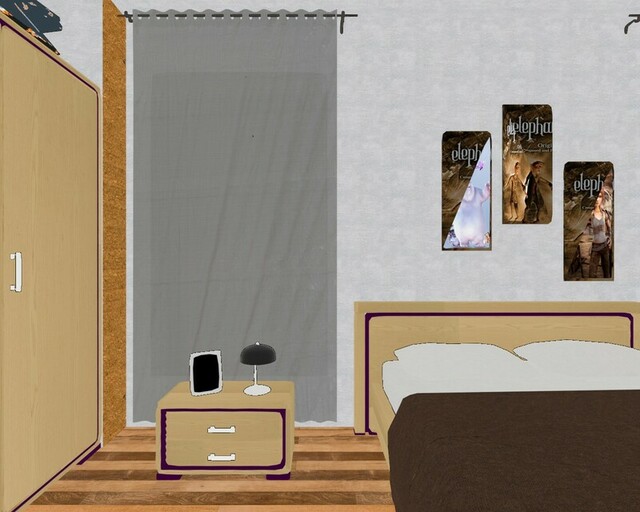}
         \caption{Albedo}
         \label{fig:five over x}
     \end{subfigure}
      \hfill
     \begin{subfigure}[b]{0.32\linewidth}
         \centering
         \includegraphics[width=\linewidth]{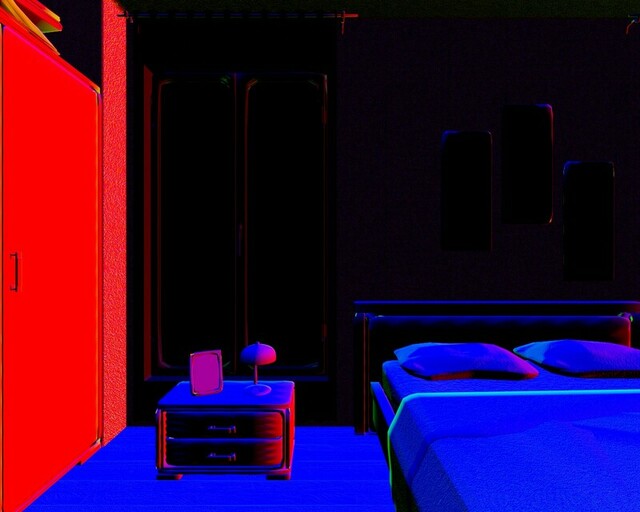}
         \caption{Normals}
         \label{fig:five over x}
     \end{subfigure}
     \begin{subfigure}[b]{0.32\linewidth}
         \centering
         \includegraphics[width=\linewidth]{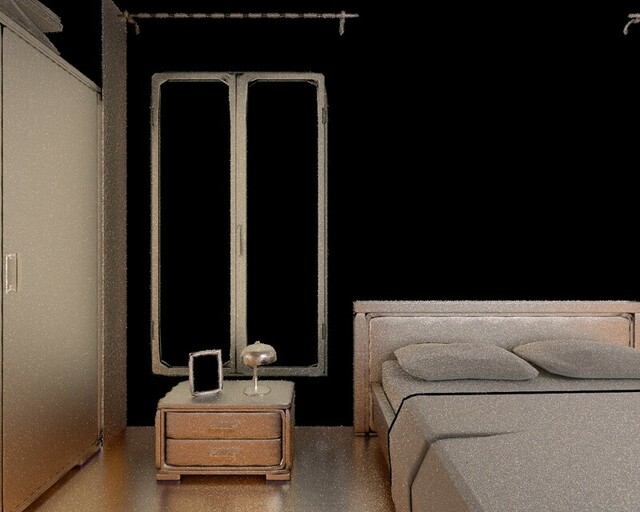}
         \caption{Specular Indirect}
         \label{fig:five over x}
     \end{subfigure}
     \begin{subfigure}[b]{0.32\linewidth}
         \centering
         \includegraphics[width=\linewidth]{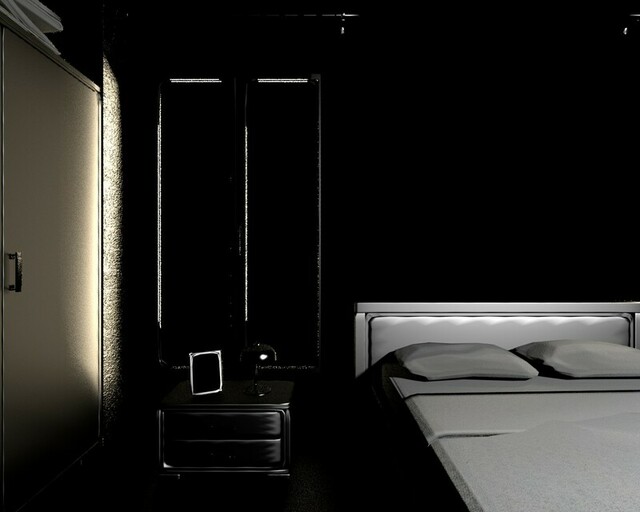}
         \caption{Specular Direct}
         \label{fig:five over x}
     \end{subfigure}
     \begin{subfigure}[b]{0.32\linewidth}
         \centering
         \includegraphics[width=\linewidth]{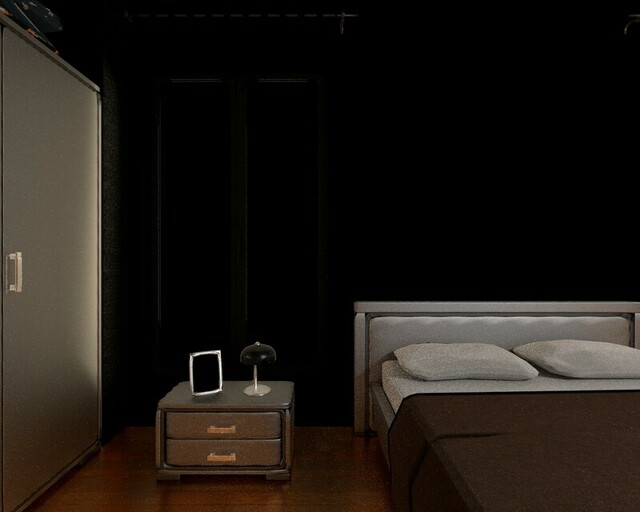}
         \caption{Specular}
         \label{fig:five over x}
     \end{subfigure}
     \begin{subfigure}[b]{0.32\linewidth}
         \centering
         \includegraphics[width=\linewidth]{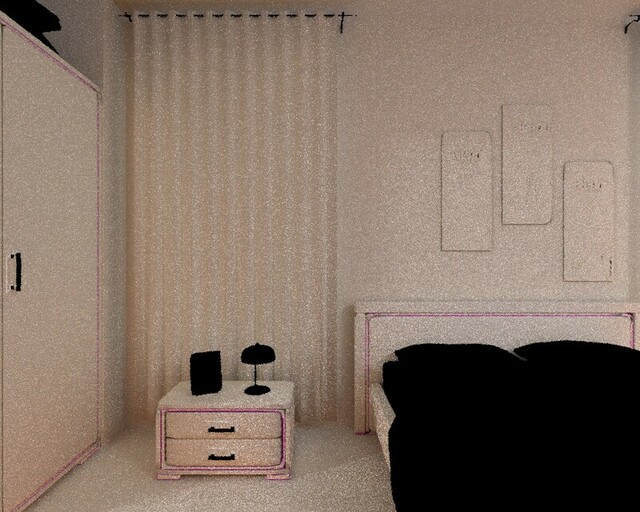}
         \caption{Diffuse Indirect}
         \label{fig:five over x}
     \end{subfigure}
     \begin{subfigure}[b]{0.32\linewidth}
         \centering
         \includegraphics[width=\linewidth]{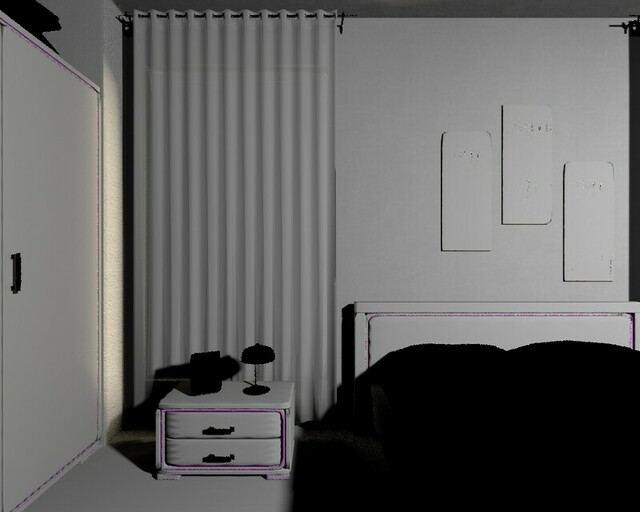}
         \caption{Diffuse Direct}
         \label{fig:five over x}
     \end{subfigure}
     \begin{subfigure}[b]{0.32\linewidth}
         \centering
         \includegraphics[width=\linewidth]{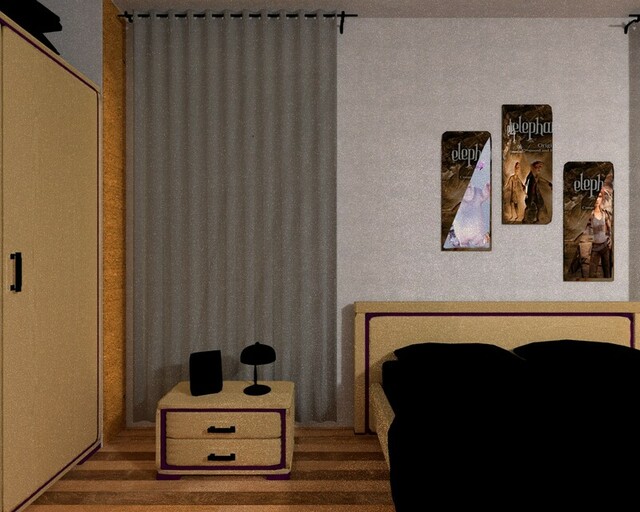}
         \caption{Diffuse}
         \label{fig:five over x}
     \end{subfigure}
     \begin{subfigure}[b]{0.32\linewidth}
         \centering
         \includegraphics[width=\linewidth]{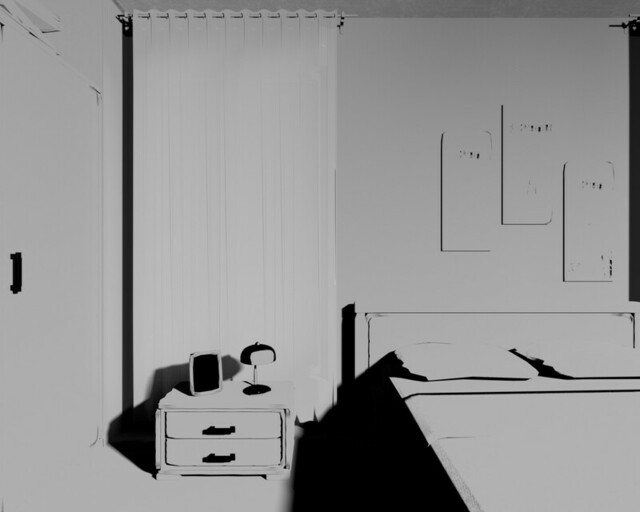}
         \caption{Shadow}
         \label{fig:five over x}
     \end{subfigure}
     \begin{subfigure}[b]{0.32\linewidth}
         \centering
         \includegraphics[width=\linewidth]{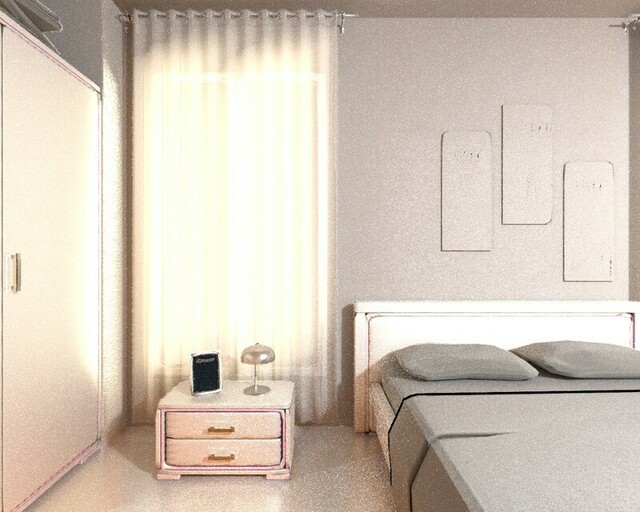}
         \caption{Lightmap}
         \label{fig:five over x}
     \end{subfigure}
     \begin{subfigure}[b]{0.32\linewidth}
         \centering
         \includegraphics[width=\linewidth]{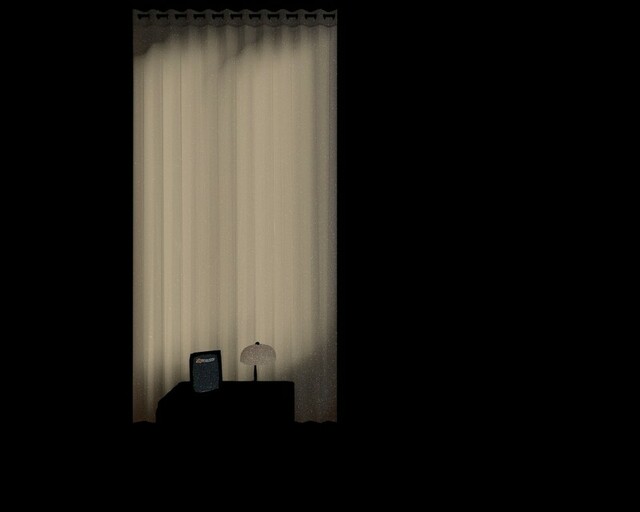}
         \caption{Transmission}
         \label{fig:five over x}
     \end{subfigure}
        \caption{Sample from scene 5}
        \label{fig:three graphs}
\end{figure}

\begin{figure}[!ht]
     \centering
     \begin{subfigure}[b]{0.49\linewidth}
         \centering
         \includegraphics[width=1\linewidth]{IMG/supp_mat/supp_mat_6/Image.jpg}
         \caption{Rendered Image}
         \label{fig:y equals x}
     \end{subfigure}
     \begin{subfigure}[b]{0.49\linewidth}
         \centering
         \includegraphics[width=1\linewidth]{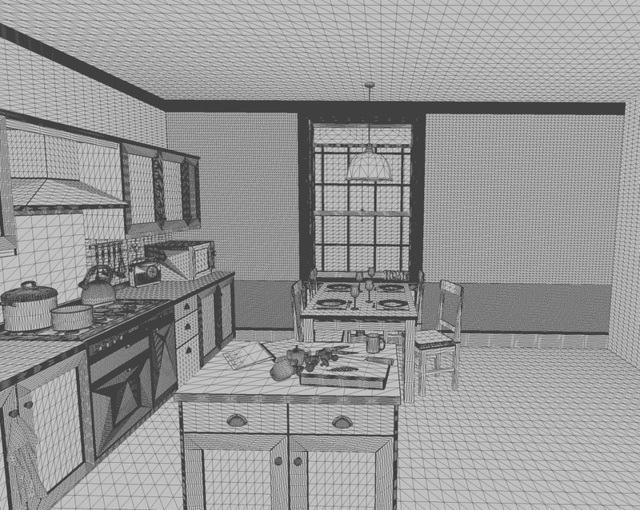}
         \caption{Triangulate mesh}
         \label{fig:y equals x}
     \end{subfigure}
     \hfill
     \begin{subfigure}[b]{0.32\linewidth}
         \centering
         \includegraphics[width=\linewidth]{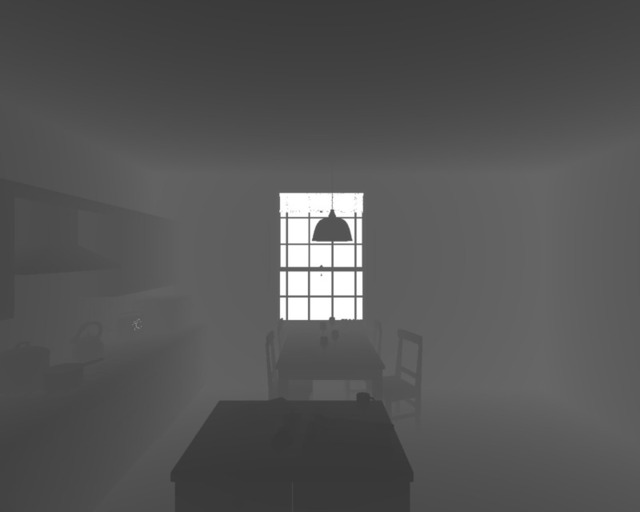}
         \caption{Depth}
         \label{fig:three sin x}
     \end{subfigure}
     \hfill
     \begin{subfigure}[b]{0.32\linewidth}
         \centering
         \includegraphics[width=\linewidth]{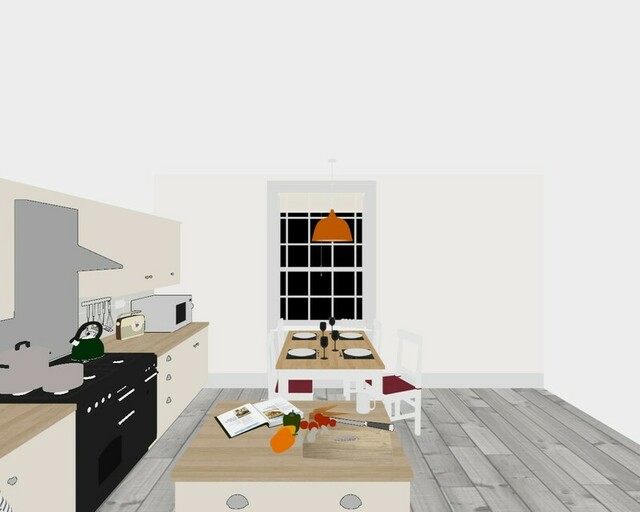}
         \caption{Albedo}
         \label{fig:five over x}
     \end{subfigure}
      \hfill
     \begin{subfigure}[b]{0.32\linewidth}
         \centering
         \includegraphics[width=\linewidth]{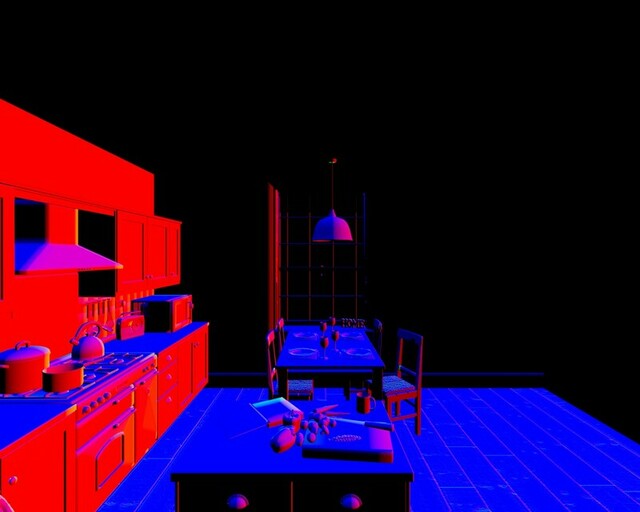}
         \caption{Normals}
         \label{fig:five over x}
     \end{subfigure}
     \begin{subfigure}[b]{0.32\linewidth}
         \centering
         \includegraphics[width=\linewidth]{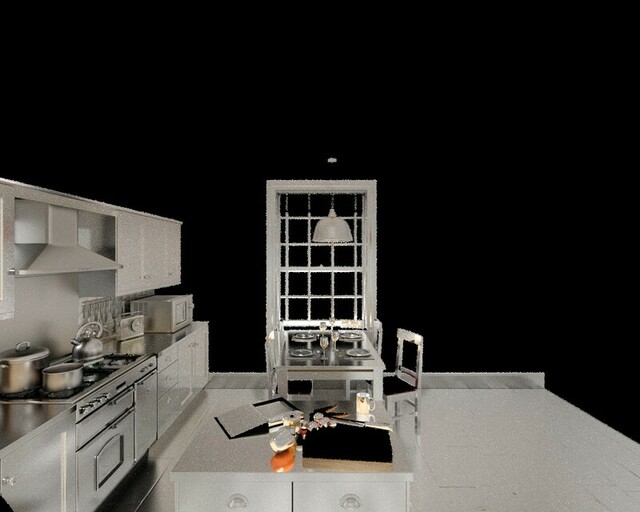}
         \caption{Specular Indirect}
         \label{fig:five over x}
     \end{subfigure}
     \begin{subfigure}[b]{0.32\linewidth}
         \centering
         \includegraphics[width=\linewidth]{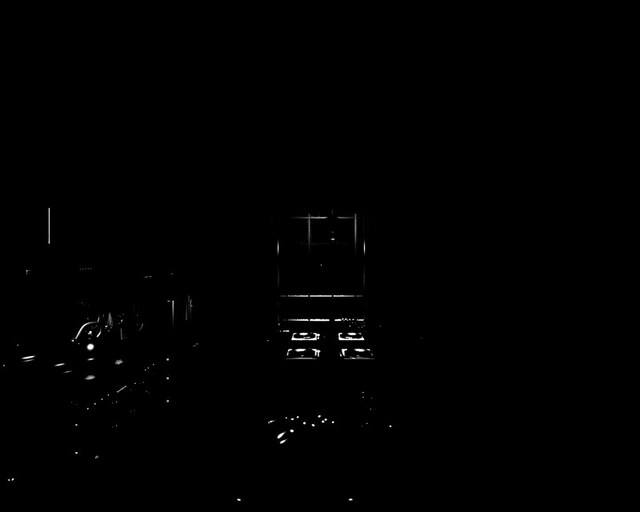}
         \caption{Specular Direct}
         \label{fig:five over x}
     \end{subfigure}
     \begin{subfigure}[b]{0.32\linewidth}
         \centering
         \includegraphics[width=\linewidth]{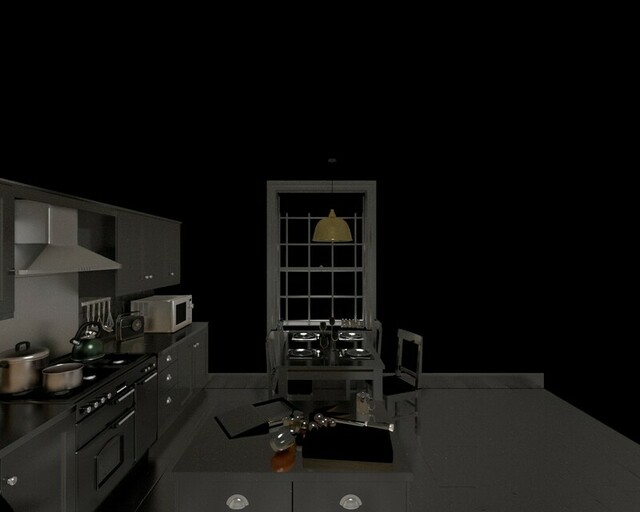}
         \caption{Specular}
         \label{fig:five over x}
     \end{subfigure}
     \begin{subfigure}[b]{0.32\linewidth}
         \centering
         \includegraphics[width=\linewidth]{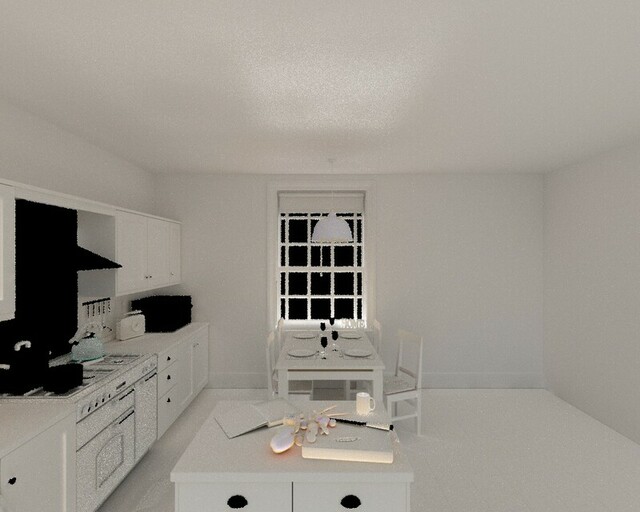}
         \caption{Diffuse Indirect}
         \label{fig:five over x}
     \end{subfigure}
     \begin{subfigure}[b]{0.32\linewidth}
         \centering
         \includegraphics[width=\linewidth]{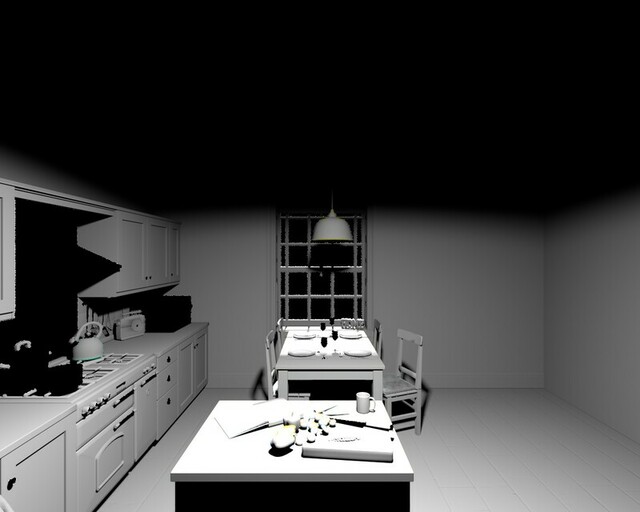}
         \caption{Diffuse Direct}
         \label{fig:five over x}
     \end{subfigure}
     \begin{subfigure}[b]{0.32\linewidth}
         \centering
         \includegraphics[width=\linewidth]{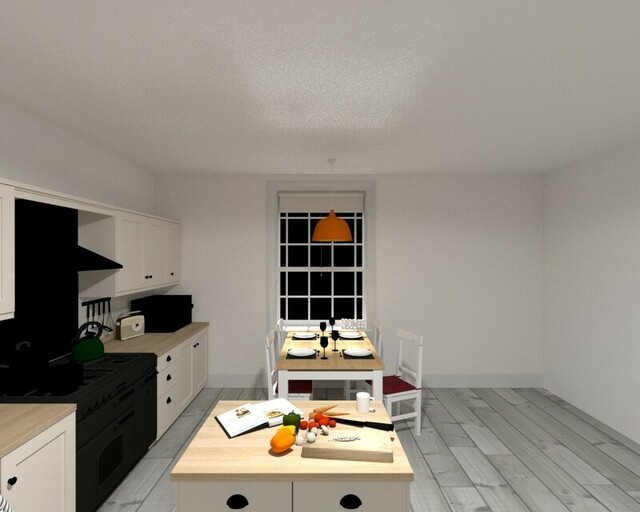}
         \caption{Diffuse}
         \label{fig:five over x}
     \end{subfigure}
     \begin{subfigure}[b]{0.32\linewidth}
         \centering
         \includegraphics[width=\linewidth]{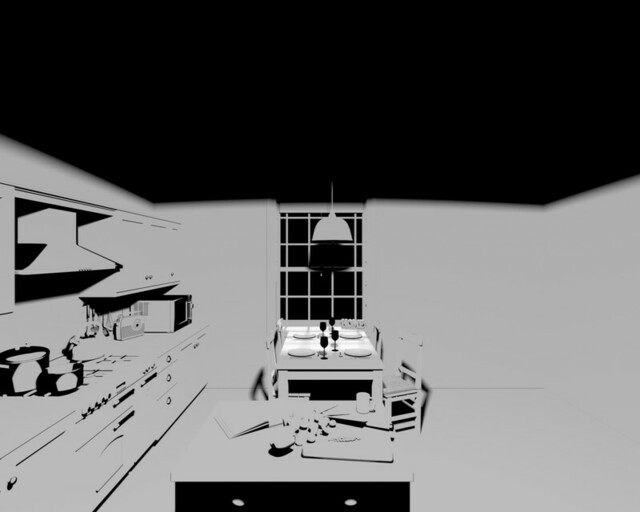}
         \caption{Shadow}
         \label{fig:five over x}
     \end{subfigure}
     \begin{subfigure}[b]{0.32\linewidth}
         \centering
         \includegraphics[width=\linewidth]{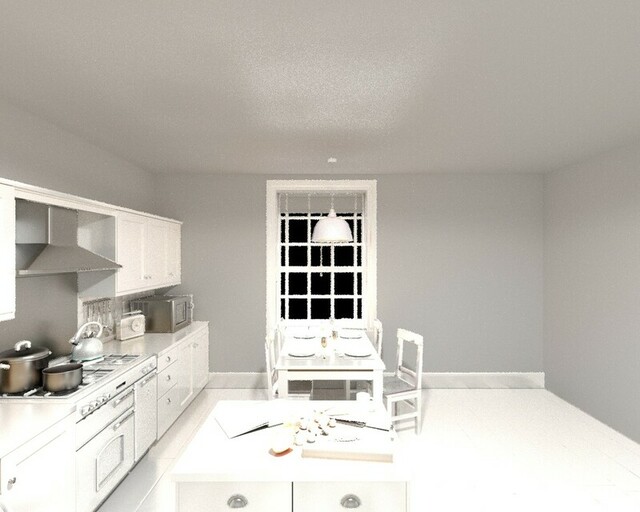}
         \caption{Lightmap}
         \label{fig:five over x}
     \end{subfigure}
     \begin{subfigure}[b]{0.32\linewidth}
         \centering
         \includegraphics[width=\linewidth]{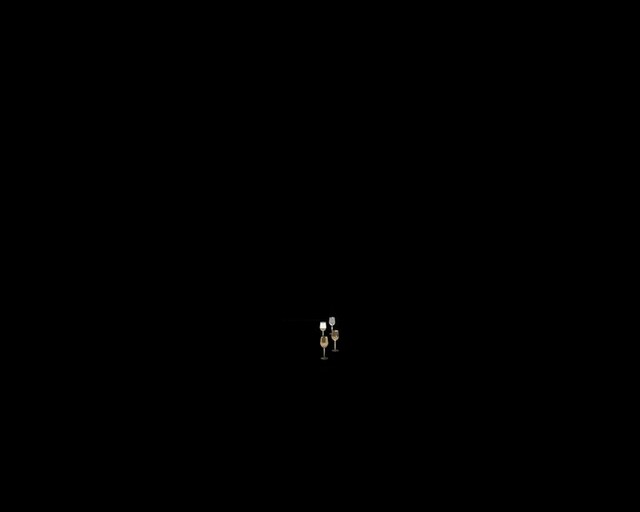}
         \caption{Transmission}
         \label{fig:five over x}
     \end{subfigure}
        \caption{Sample from scene 6}
        \label{fig:three graphs}
\end{figure}

\begin{figure}[!ht]
     \centering
     \begin{subfigure}[b]{0.49\linewidth}
         \centering
         \includegraphics[width=1\linewidth]{IMG/supp_mat/supp_mat_7/Image.jpg}
         \caption{Rendered Image}
         \label{fig:y equals x}
     \end{subfigure}
     \begin{subfigure}[b]{0.49\linewidth}
         \centering
         \includegraphics[width=1\linewidth]{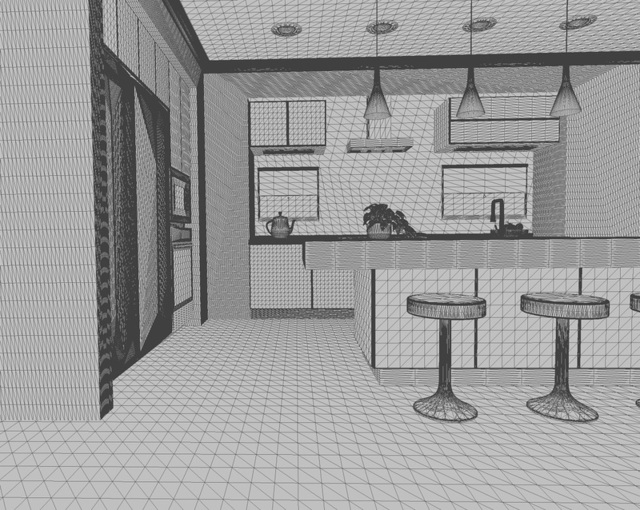}
         \caption{Triangulate mesh}
         \label{fig:y equals x}
     \end{subfigure}
     \hfill
     \begin{subfigure}[b]{0.32\linewidth}
         \centering
         \includegraphics[width=\linewidth]{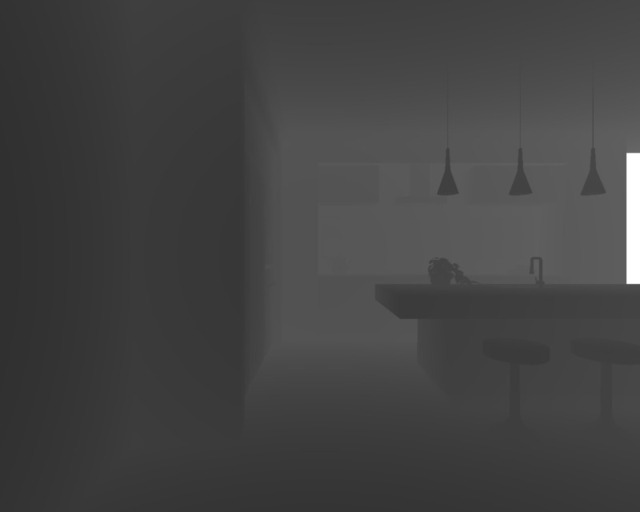}
         \caption{Depth}
         \label{fig:three sin x}
     \end{subfigure}
     \hfill
     \begin{subfigure}[b]{0.32\linewidth}
         \centering
         \includegraphics[width=\linewidth]{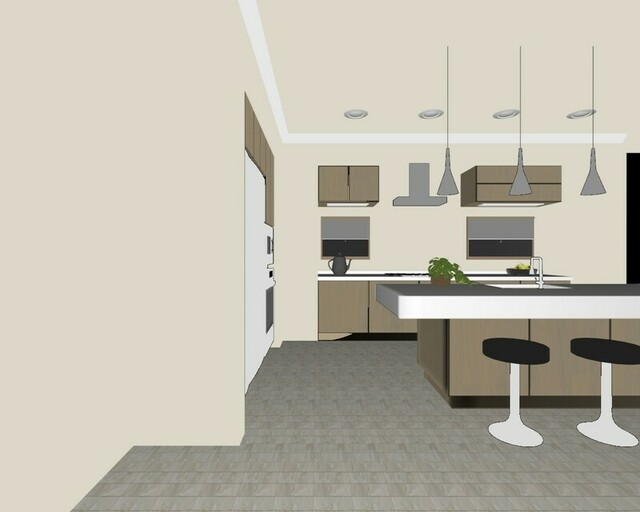}
         \caption{Albedo}
         \label{fig:five over x}
     \end{subfigure}
      \hfill
     \begin{subfigure}[b]{0.32\linewidth}
         \centering
         \includegraphics[width=\linewidth]{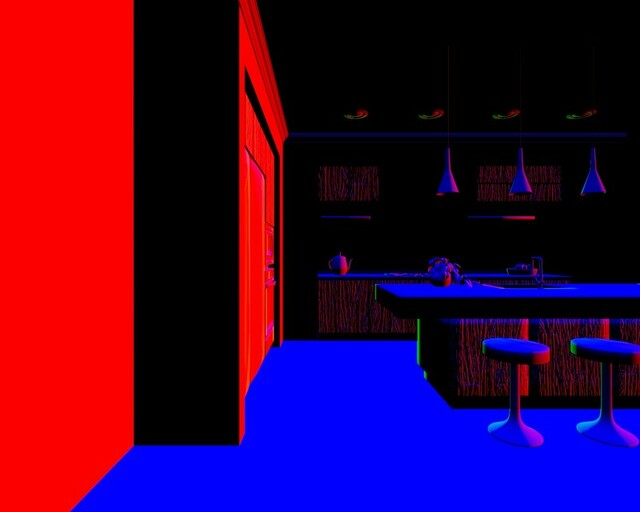}
         \caption{Normals}
         \label{fig:five over x}
     \end{subfigure}
     \begin{subfigure}[b]{0.32\linewidth}
         \centering
         \includegraphics[width=\linewidth]{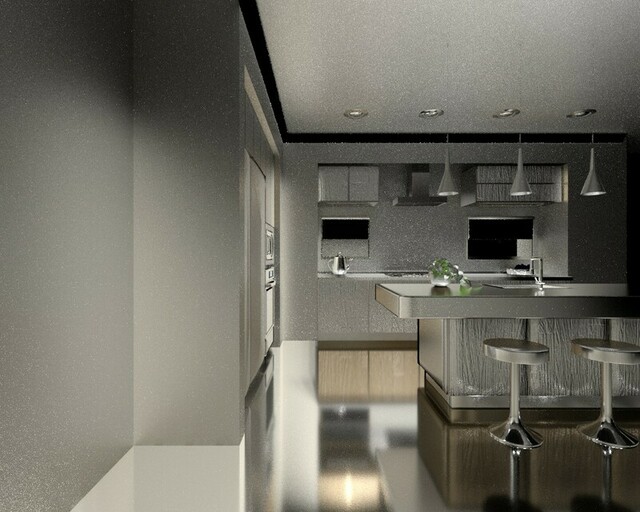}
         \caption{Specular Indirect}
         \label{fig:five over x}
     \end{subfigure}
     \begin{subfigure}[b]{0.32\linewidth}
         \centering
         \includegraphics[width=\linewidth]{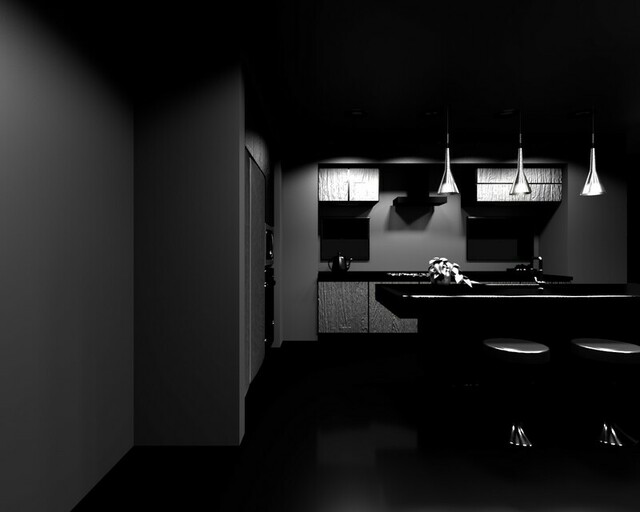}
         \caption{Specular Direct}
         \label{fig:five over x}
     \end{subfigure}
     \begin{subfigure}[b]{0.32\linewidth}
         \centering
         \includegraphics[width=\linewidth]{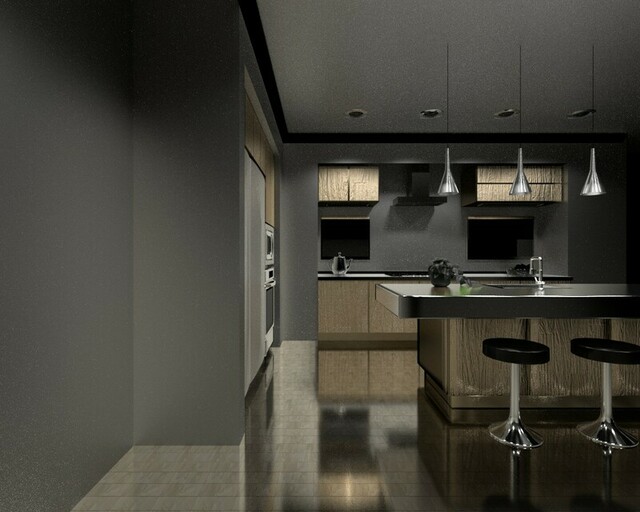}
         \caption{Specular}
         \label{fig:five over x}
     \end{subfigure}
     \begin{subfigure}[b]{0.32\linewidth}
         \centering
         \includegraphics[width=\linewidth]{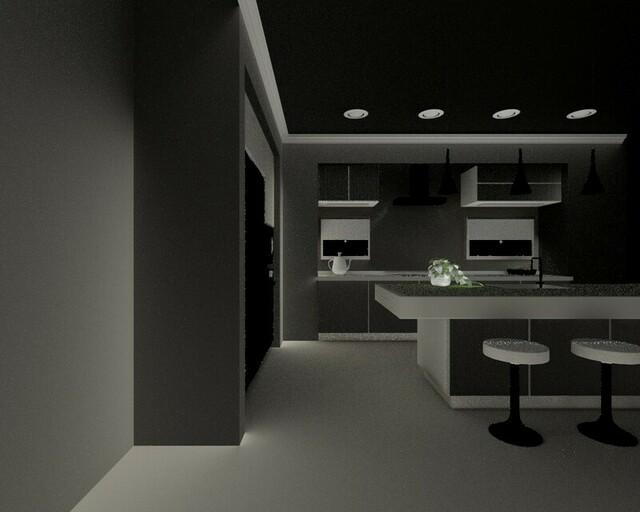}
         \caption{Diffuse Indirect}
         \label{fig:five over x}
     \end{subfigure}
     \begin{subfigure}[b]{0.32\linewidth}
         \centering
         \includegraphics[width=\linewidth]{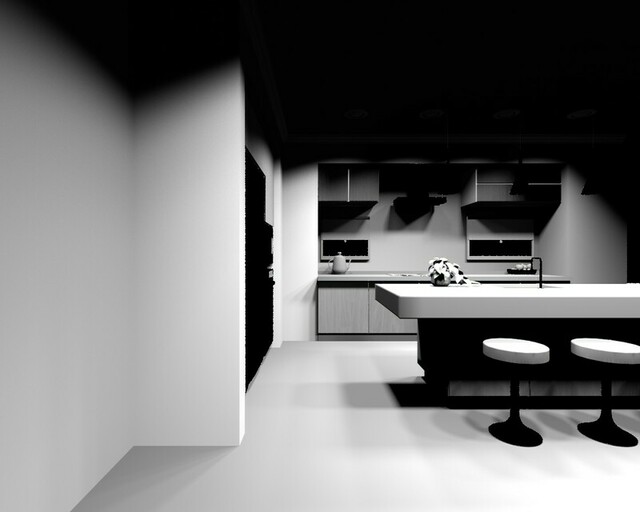}
         \caption{Diffuse Direct}
         \label{fig:five over x}
     \end{subfigure}
     \begin{subfigure}[b]{0.32\linewidth}
         \centering
         \includegraphics[width=\linewidth]{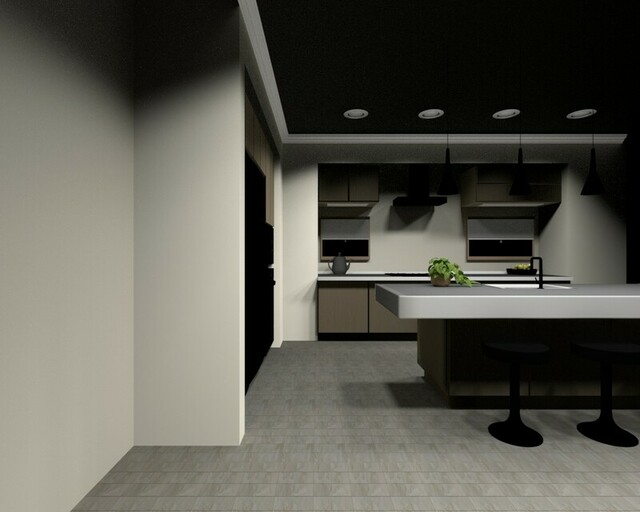}
         \caption{Diffuse}
         \label{fig:five over x}
     \end{subfigure}
     \begin{subfigure}[b]{0.32\linewidth}
         \centering
         \includegraphics[width=\linewidth]{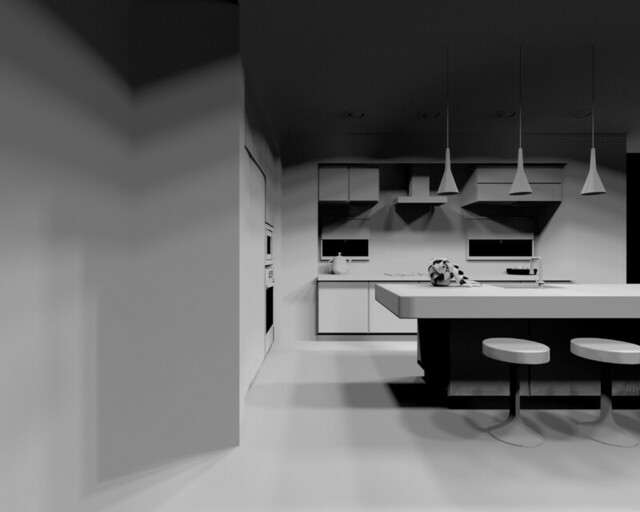}
         \caption{Shadow}
         \label{fig:five over x}
     \end{subfigure}
     \begin{subfigure}[b]{0.32\linewidth}
         \centering
         \includegraphics[width=\linewidth]{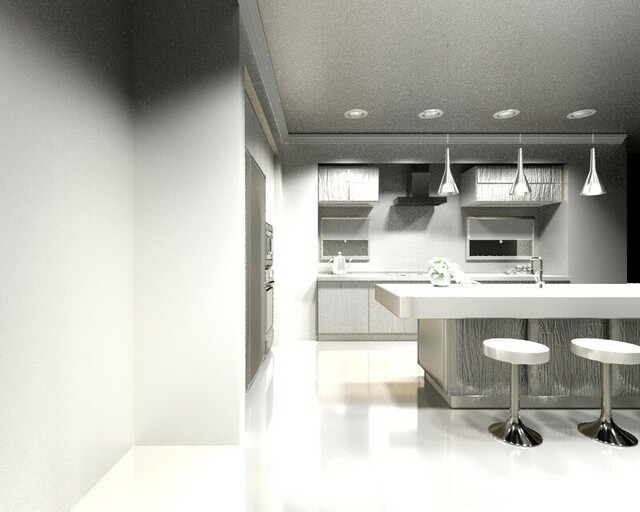}
         \caption{Lightmap}
         \label{fig:five over x}
     \end{subfigure}
     \begin{subfigure}[b]{0.32\linewidth}
         \centering
         \includegraphics[width=\linewidth]{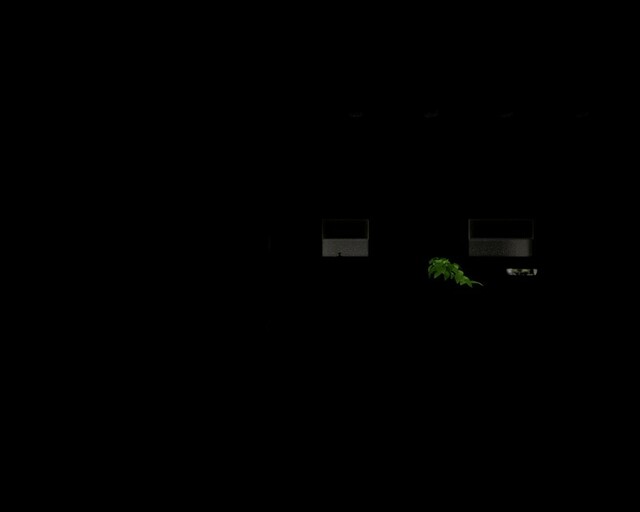}
         \caption{Transmission}
         \label{fig:five over x}
     \end{subfigure}
        \caption{Sample from scene 7}
        \label{fig:three graphs}
\end{figure}

\begin{figure}[!ht]
     \centering
     \begin{subfigure}[b]{0.49\linewidth}
         \centering
         \includegraphics[width=1\linewidth]{IMG/supp_mat/supp_mat_8/Image.jpg}
         \caption{Rendered Image}
         \label{fig:y equals x}
     \end{subfigure}
    \begin{subfigure}[b]{0.49\linewidth}
         \centering
         \includegraphics[width=1\linewidth]{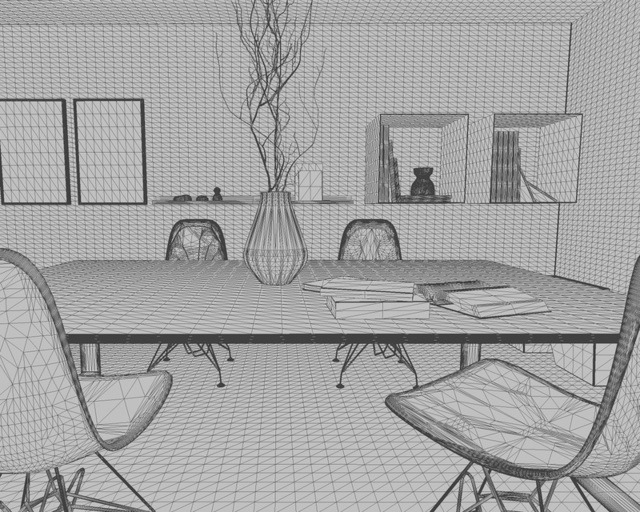}
         \caption{Triangulate mesh}
         \label{fig:y equals x}
     \end{subfigure}
     \hfill
     \begin{subfigure}[b]{0.32\linewidth}
         \centering
         \includegraphics[width=\linewidth]{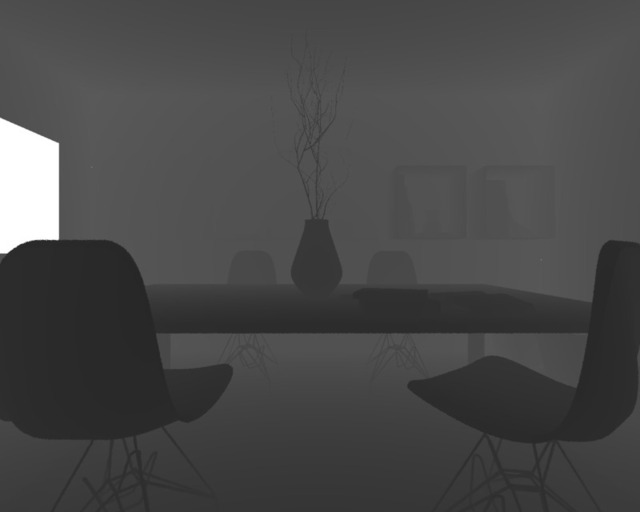}
         \caption{Depth}
         \label{fig:three sin x}
     \end{subfigure}
     \hfill
     \begin{subfigure}[b]{0.32\linewidth}
         \centering
         \includegraphics[width=\linewidth]{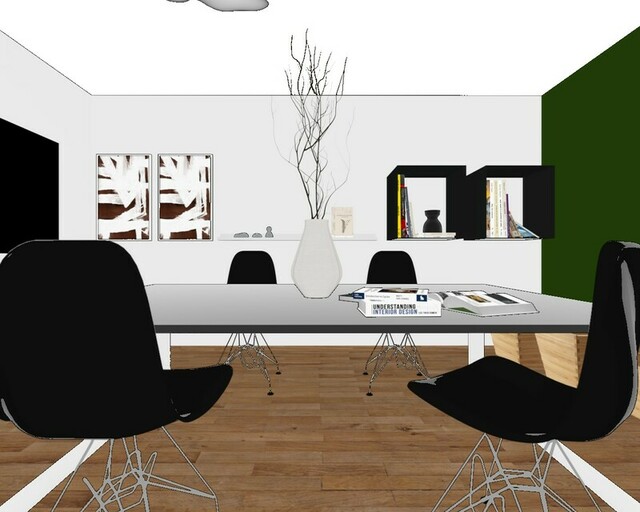}
         \caption{Albedo}
         \label{fig:five over x}
     \end{subfigure}
      \hfill
     \begin{subfigure}[b]{0.32\linewidth}
         \centering
         \includegraphics[width=\linewidth]{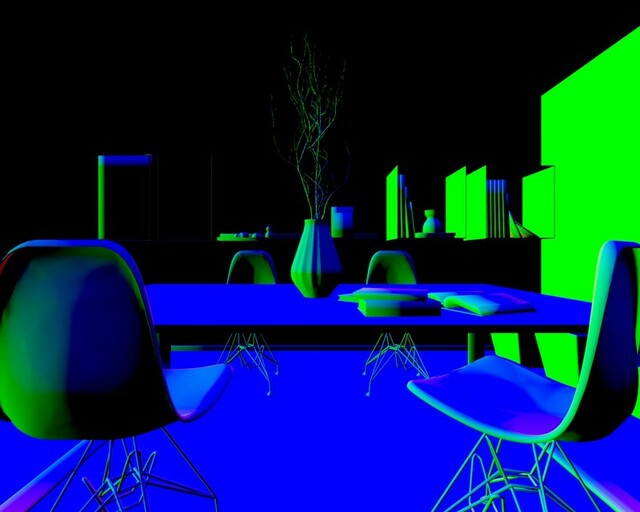}
         \caption{Normals}
         \label{fig:five over x}
     \end{subfigure}
     \begin{subfigure}[b]{0.32\linewidth}
         \centering
         \includegraphics[width=\linewidth]{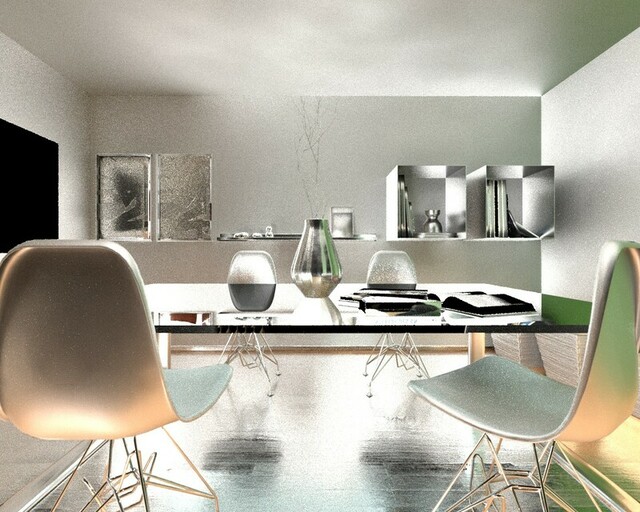}
         \caption{Specular Indirect}
         \label{fig:five over x}
     \end{subfigure}
     \begin{subfigure}[b]{0.32\linewidth}
         \centering
         \includegraphics[width=\linewidth]{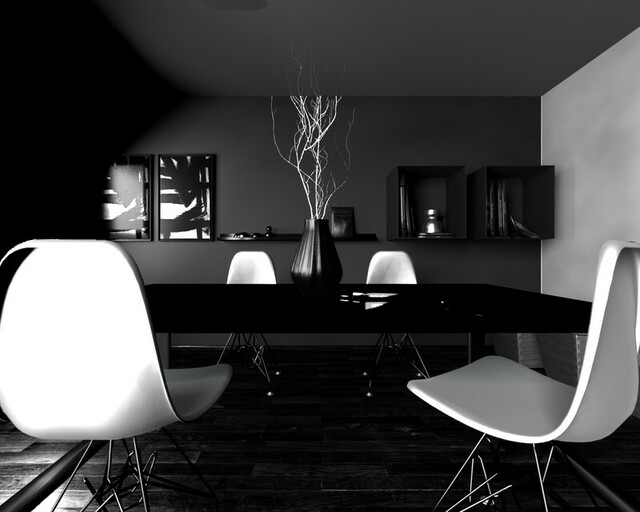}
         \caption{Specular Direct}
         \label{fig:five over x}
     \end{subfigure}
     \begin{subfigure}[b]{0.32\linewidth}
         \centering
         \includegraphics[width=\linewidth]{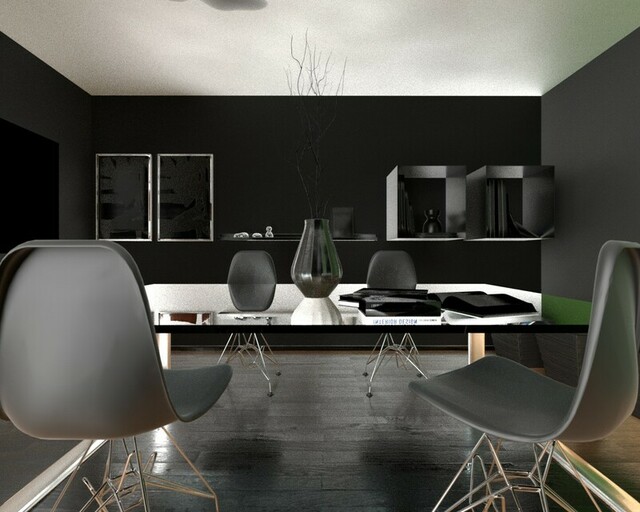}
         \caption{Specular}
         \label{fig:five over x}
     \end{subfigure}
     \begin{subfigure}[b]{0.32\linewidth}
         \centering
         \includegraphics[width=\linewidth]{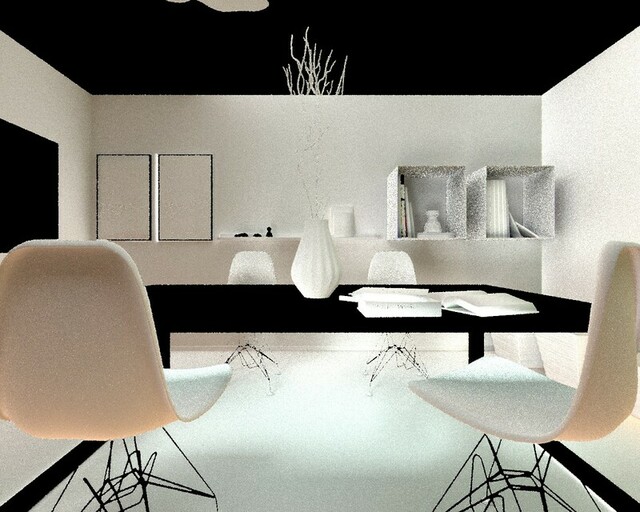}
         \caption{Diffuse Indirect}
         \label{fig:five over x}
     \end{subfigure}
     \begin{subfigure}[b]{0.32\linewidth}
         \centering
         \includegraphics[width=\linewidth]{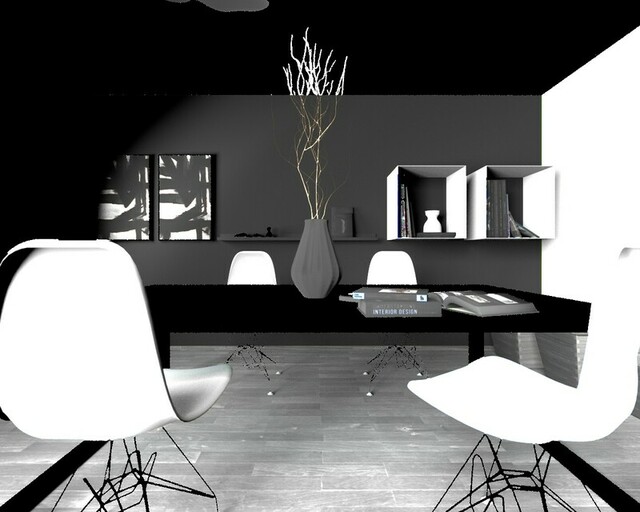}
         \caption{Diffuse Direct}
         \label{fig:five over x}
     \end{subfigure}
     \begin{subfigure}[b]{0.32\linewidth}
         \centering
         \includegraphics[width=\linewidth]{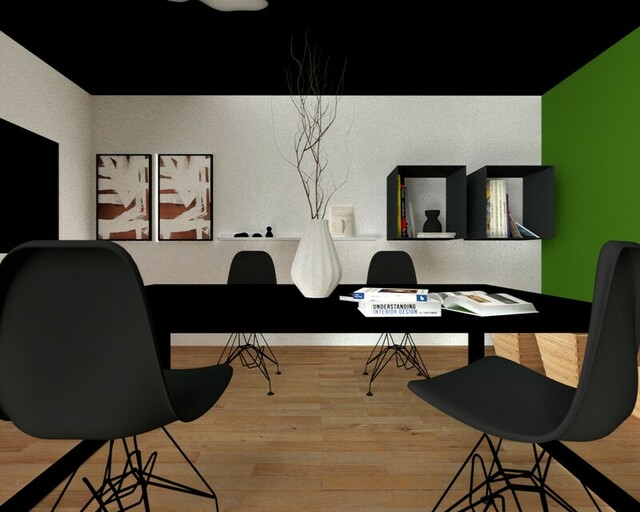}
         \caption{Diffuse}
         \label{fig:five over x}
     \end{subfigure}
     \begin{subfigure}[b]{0.32\linewidth}
         \centering
         \includegraphics[width=\linewidth]{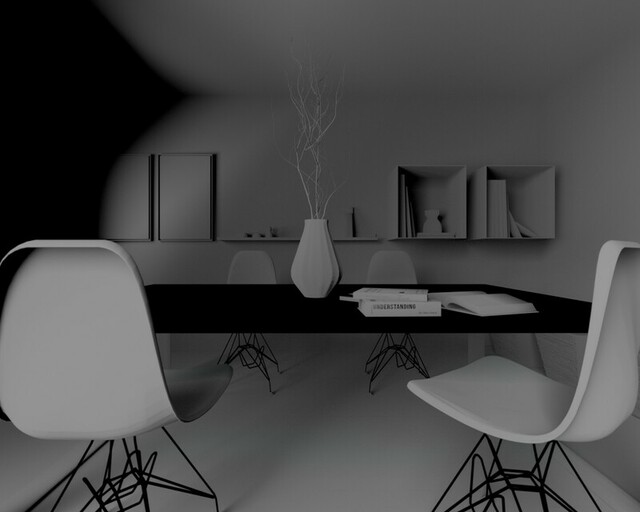}
         \caption{Shadow}
         \label{fig:five over x}
     \end{subfigure}
     \begin{subfigure}[b]{0.32\linewidth}
         \centering
         \includegraphics[width=\linewidth]{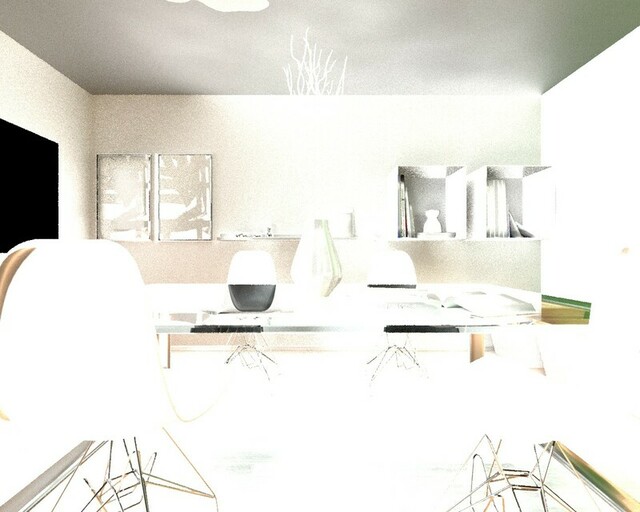}
         \caption{Lightmap}
         \label{fig:five over x}
     \end{subfigure}
     \begin{subfigure}[b]{0.32\linewidth}
         \centering
         \includegraphics[width=\linewidth]{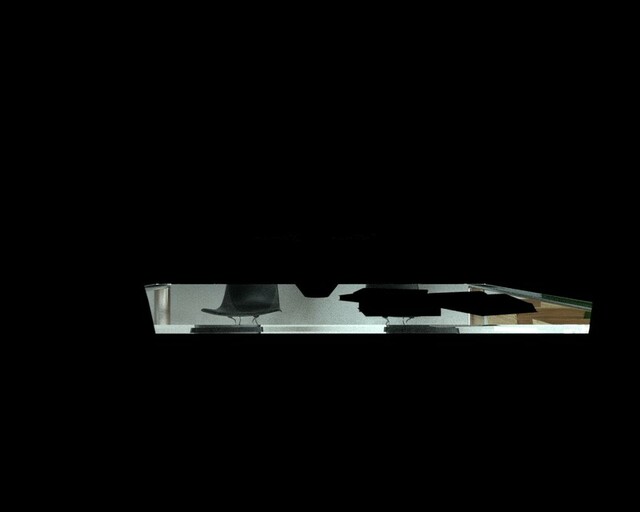}
         \caption{Transmission}
         \label{fig:five over x}
     \end{subfigure}
        \caption{Sample from scene 8}
        \label{fig:three graphs}
\end{figure}

\begin{figure}[!ht]
     \centering
     \begin{subfigure}[b]{0.49\linewidth}
         \centering
         \includegraphics[width=1\linewidth]{IMG/supp_mat/supp_mat_9/Image.jpg}
         \caption{Rendered Image}
         \label{fig:y equals x}
     \end{subfigure}
     \begin{subfigure}[b]{0.49\linewidth}
         \centering
         \includegraphics[width=1\linewidth]{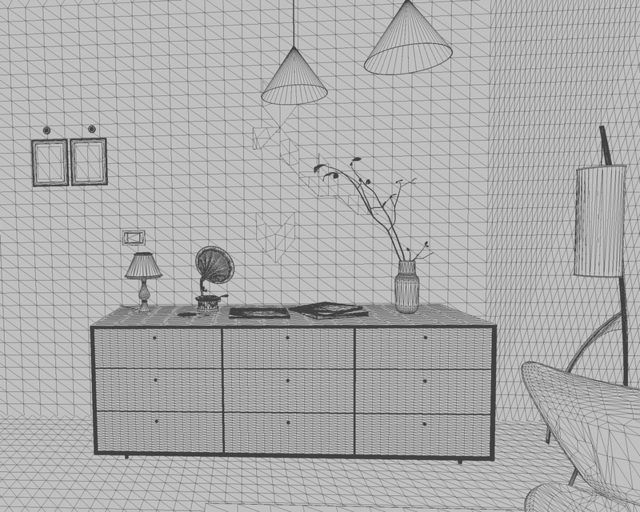}
         \caption{Triangulate mesh}
         \label{fig:y equals x}
     \end{subfigure}
     \hfill
     \begin{subfigure}[b]{0.32\linewidth}
         \centering
         \includegraphics[width=\linewidth]{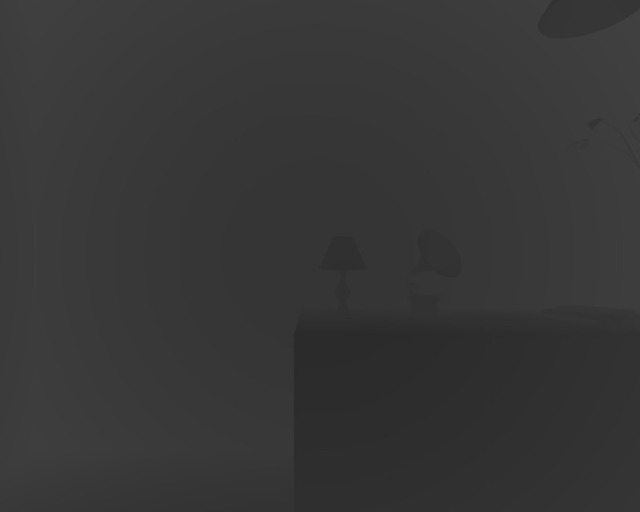}
         \caption{Depth}
         \label{fig:three sin x}
     \end{subfigure}
     \hfill
     \begin{subfigure}[b]{0.32\linewidth}
         \centering
         \includegraphics[width=\linewidth]{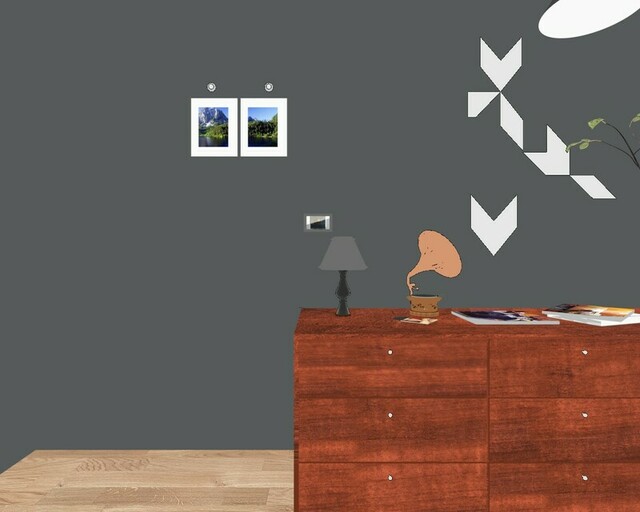}
         \caption{Albedo}
         \label{fig:five over x}
     \end{subfigure}
      \hfill
     \begin{subfigure}[b]{0.32\linewidth}
         \centering
         \includegraphics[width=\linewidth]{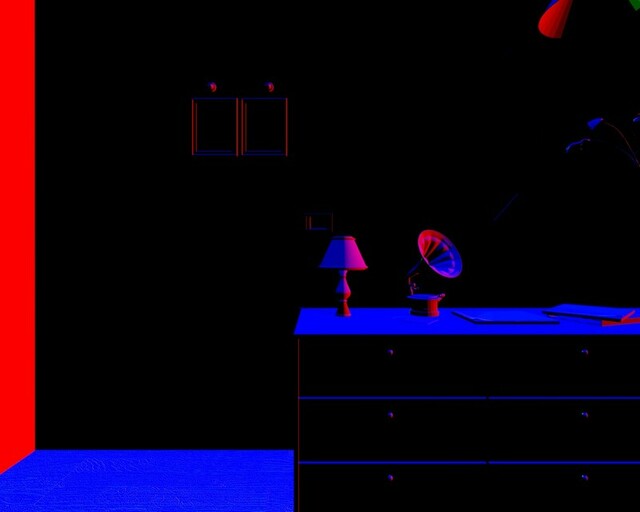}
         \caption{Normals}
         \label{fig:five over x}
     \end{subfigure}
     \begin{subfigure}[b]{0.32\linewidth}
         \centering
         \includegraphics[width=\linewidth]{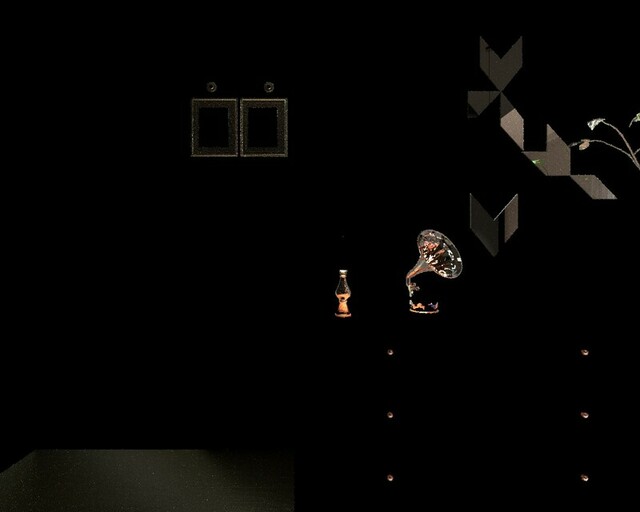}
         \caption{Specular Indirect}
         \label{fig:five over x}
     \end{subfigure}
     \begin{subfigure}[b]{0.32\linewidth}
         \centering
         \includegraphics[width=\linewidth]{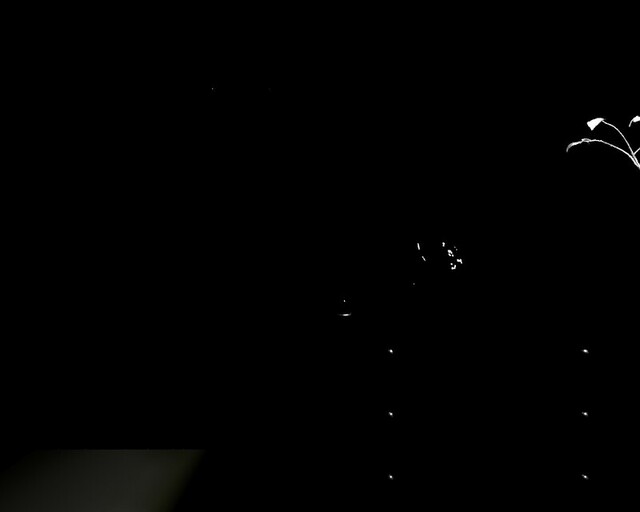}
         \caption{Specular Direct}
         \label{fig:five over x}
     \end{subfigure}
     \begin{subfigure}[b]{0.32\linewidth}
         \centering
         \includegraphics[width=\linewidth]{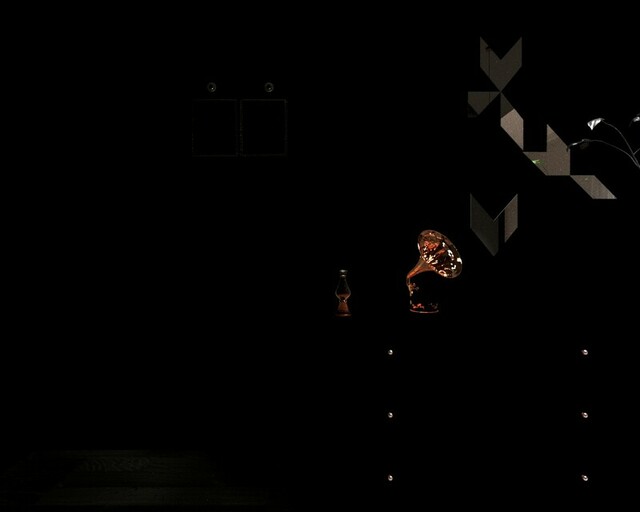}
         \caption{Specular}
         \label{fig:five over x}
     \end{subfigure}
     \begin{subfigure}[b]{0.32\linewidth}
         \centering
         \includegraphics[width=\linewidth]{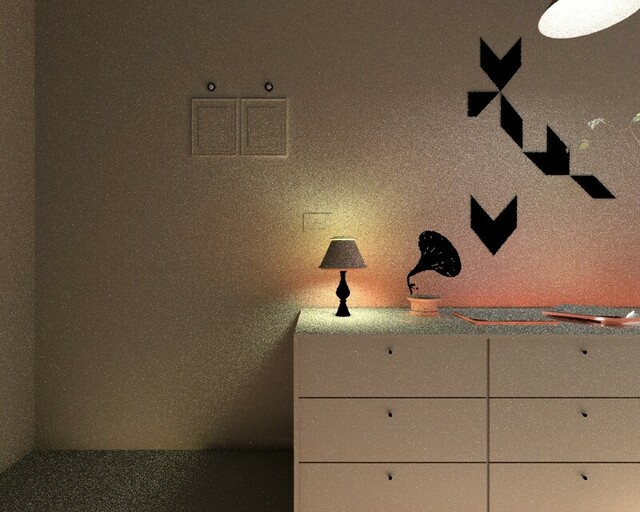}
         \caption{Diffuse Indirect}
         \label{fig:five over x}
     \end{subfigure}
     \begin{subfigure}[b]{0.32\linewidth}
         \centering
         \includegraphics[width=\linewidth]{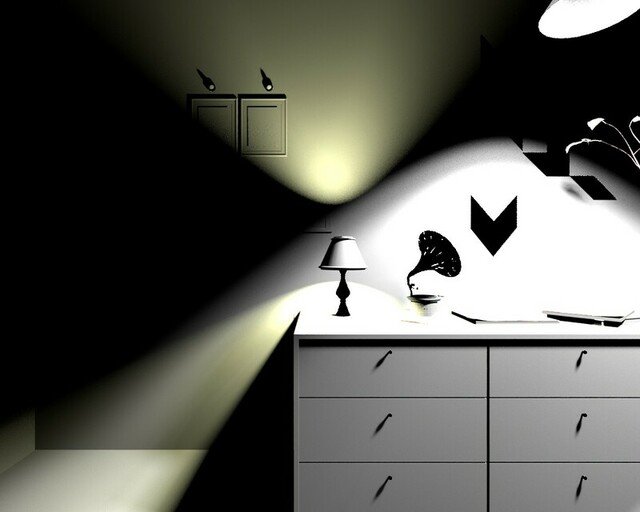}
         \caption{Diffuse Direct}
         \label{fig:five over x}
     \end{subfigure}
     \begin{subfigure}[b]{0.32\linewidth}
         \centering
         \includegraphics[width=\linewidth]{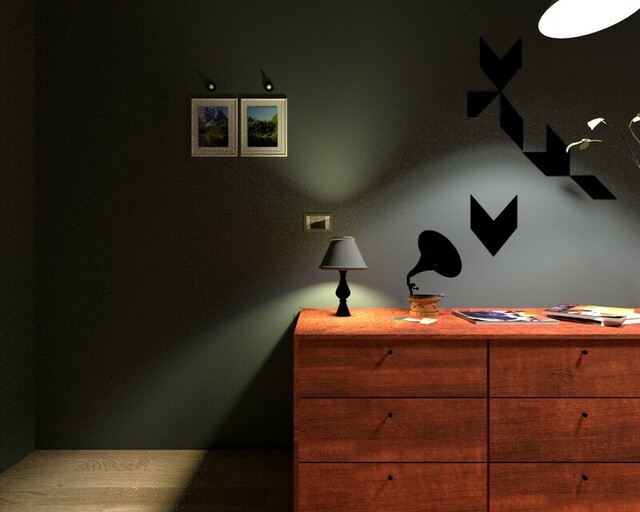}
         \caption{Diffuse}
         \label{fig:five over x}
     \end{subfigure}
     \begin{subfigure}[b]{0.32\linewidth}
         \centering
         \includegraphics[width=\linewidth]{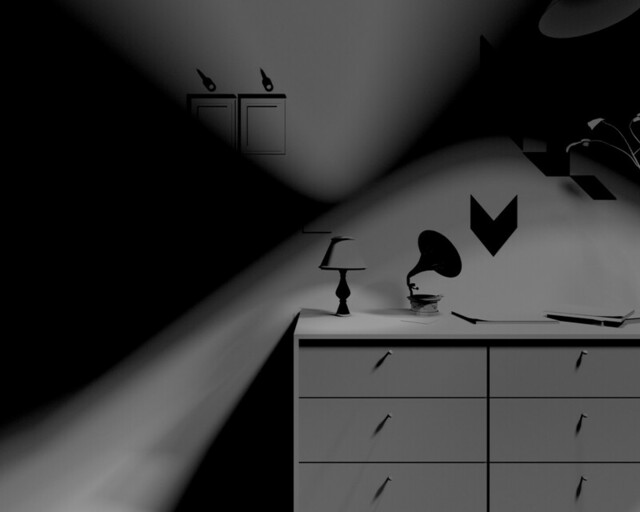}
         \caption{Shadow}
         \label{fig:five over x}
     \end{subfigure}
     \begin{subfigure}[b]{0.32\linewidth}
         \centering
         \includegraphics[width=\linewidth]{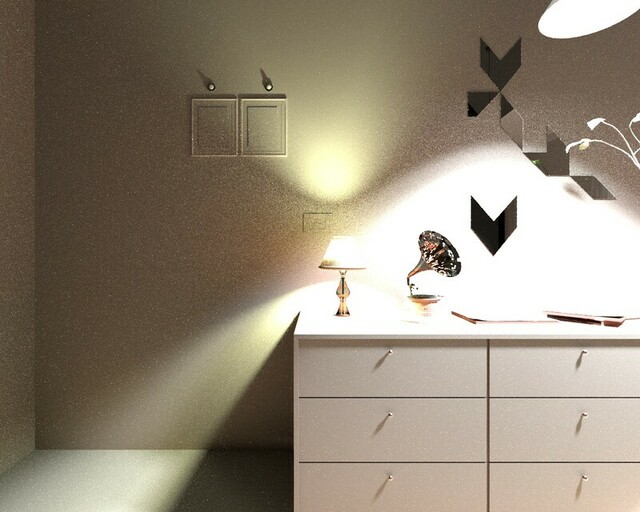}
         \caption{Lightmap}
         \label{fig:five over x}
     \end{subfigure}
     \begin{subfigure}[b]{0.32\linewidth}
         \centering
         \includegraphics[width=\linewidth]{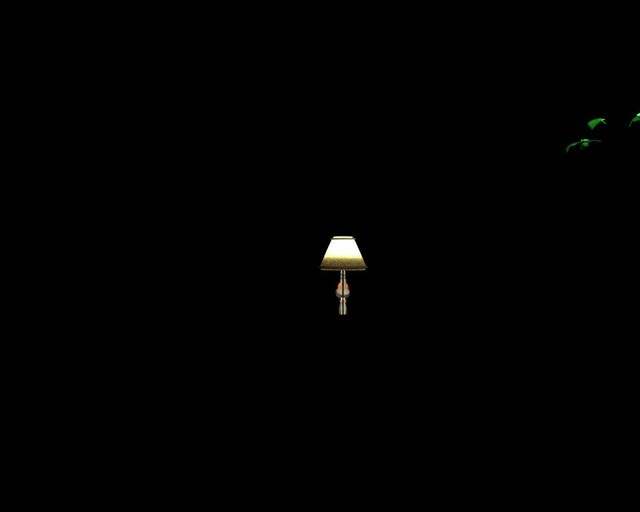}
         \caption{Transmission}
         \label{fig:five over x}
     \end{subfigure}
        \caption{Sample from scene 9}
        \label{fig:three graphs}
\end{figure}

\begin{figure}[!ht]
     \centering
     \begin{subfigure}[b]{0.49\linewidth}
         \centering
         \includegraphics[width=1\linewidth]{IMG/supp_mat/supp_mat_10/Image.jpg}
         \caption{Rendered Image}
         \label{fig:y equals x}
     \end{subfigure}
     \begin{subfigure}[b]{0.49\linewidth}
         \centering
         \includegraphics[width=1\linewidth]{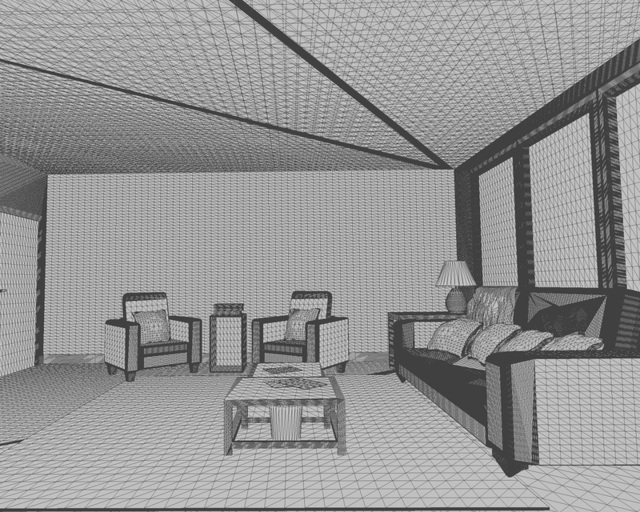}
         \caption{Triangulate mesh}
         \label{fig:y equals x}
     \end{subfigure}
     \hfill
     \begin{subfigure}[b]{0.32\linewidth}
         \centering
         \includegraphics[width=\linewidth]{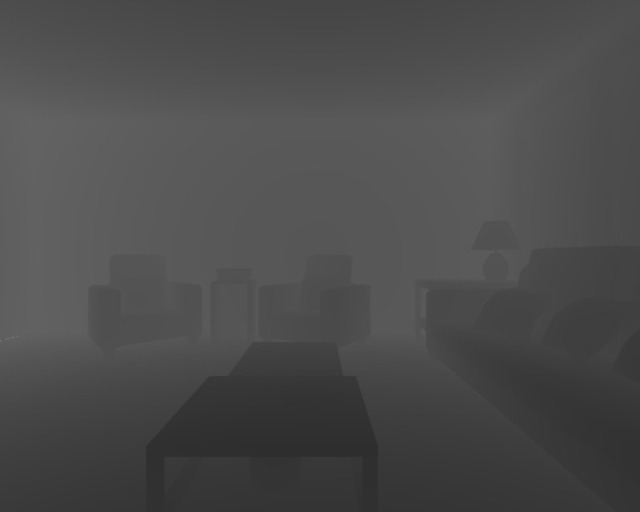}
         \caption{Depth}
         \label{fig:three sin x}
     \end{subfigure}
     \hfill
     \begin{subfigure}[b]{0.32\linewidth}
         \centering
         \includegraphics[width=\linewidth]{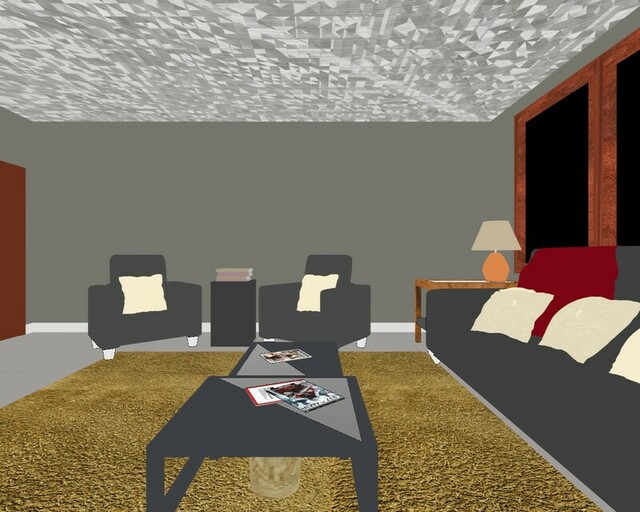}
         \caption{Albedo}
         \label{fig:five over x}
     \end{subfigure}
      \hfill
     \begin{subfigure}[b]{0.32\linewidth}
         \centering
         \includegraphics[width=\linewidth]{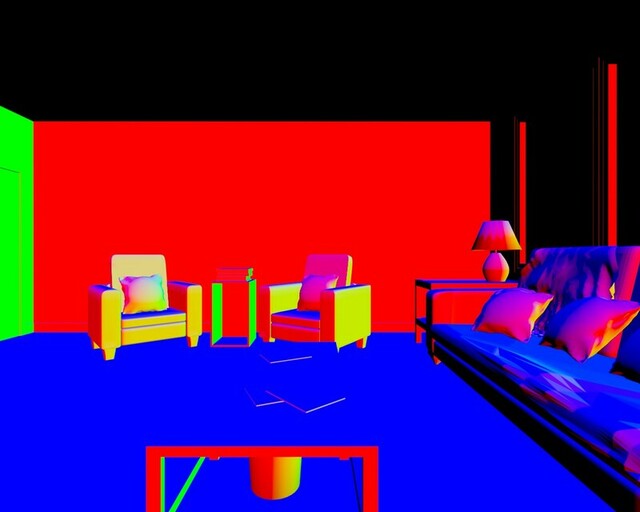}
         \caption{Normals}
         \label{fig:five over x}
     \end{subfigure}
     \begin{subfigure}[b]{0.32\linewidth}
         \centering
         \includegraphics[width=\linewidth]{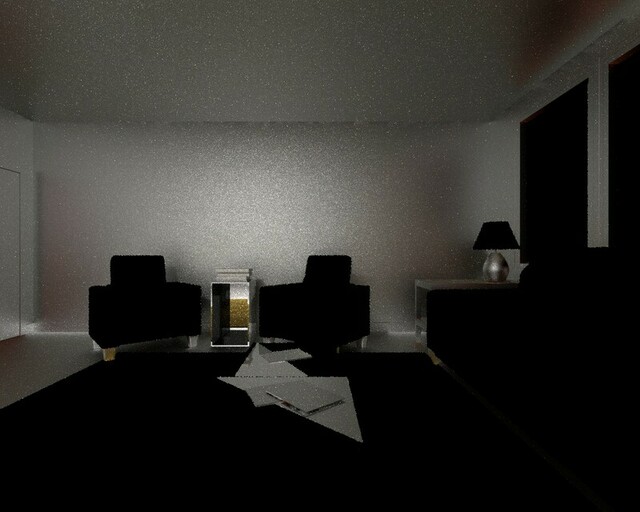}
         \caption{Specular Indirect}
         \label{fig:five over x}
     \end{subfigure}
     \begin{subfigure}[b]{0.32\linewidth}
         \centering
         \includegraphics[width=\linewidth]{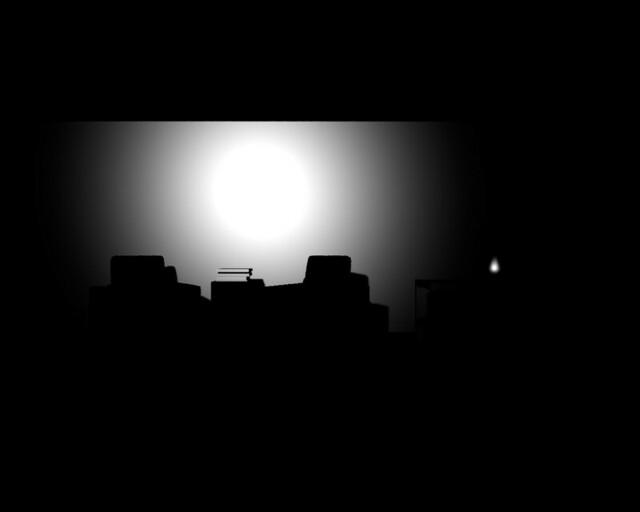}
         \caption{Specular Direct}
         \label{fig:five over x}
     \end{subfigure}
     \begin{subfigure}[b]{0.32\linewidth}
         \centering
         \includegraphics[width=\linewidth]{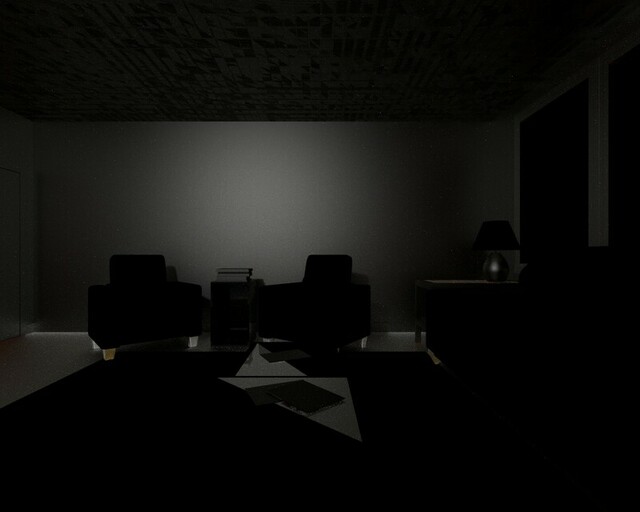}
         \caption{Specular}
         \label{fig:five over x}
     \end{subfigure}
     \begin{subfigure}[b]{0.32\linewidth}
         \centering
         \includegraphics[width=\linewidth]{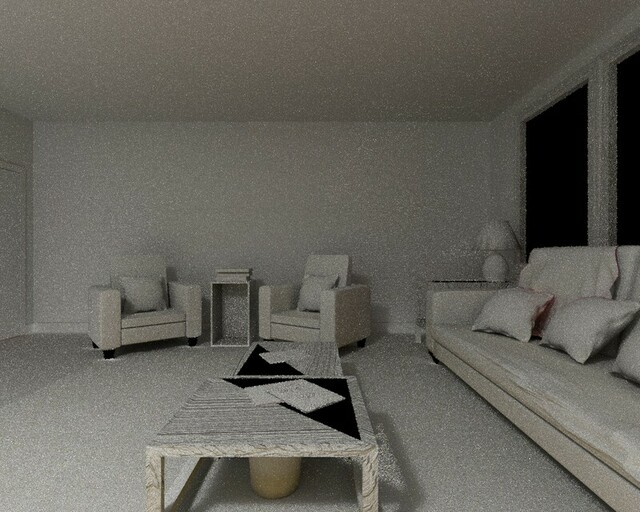}
         \caption{Diffuse Indirect}
         \label{fig:five over x}
     \end{subfigure}
     \begin{subfigure}[b]{0.32\linewidth}
         \centering
         \includegraphics[width=\linewidth]{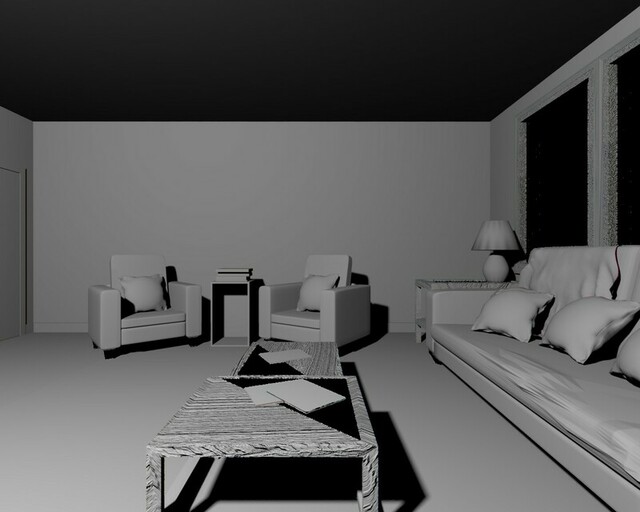}
         \caption{Diffuse Direct}
         \label{fig:five over x}
     \end{subfigure}
     \begin{subfigure}[b]{0.32\linewidth}
         \centering
         \includegraphics[width=\linewidth]{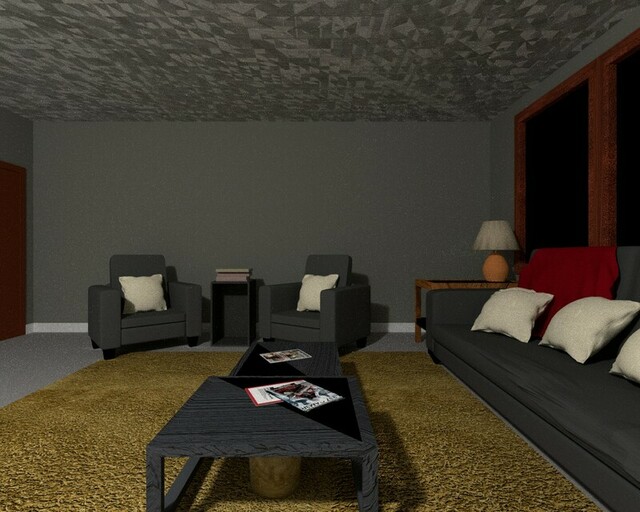}
         \caption{Diffuse}
         \label{fig:five over x}
     \end{subfigure}
     \begin{subfigure}[b]{0.32\linewidth}
         \centering
         \includegraphics[width=\linewidth]{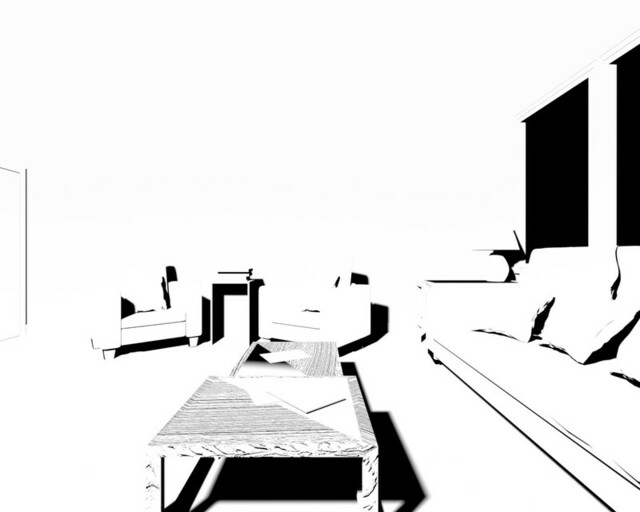}
         \caption{Shadow}
         \label{fig:five over x}
     \end{subfigure}
     \begin{subfigure}[b]{0.32\linewidth}
         \centering
         \includegraphics[width=\linewidth]{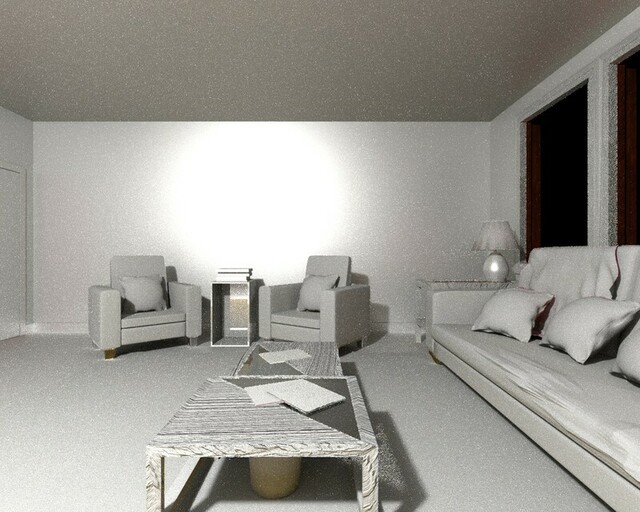}
         \caption{Lightmap}
         \label{fig:five over x}
     \end{subfigure}
     \begin{subfigure}[b]{0.32\linewidth}
         \centering
         \includegraphics[width=\linewidth]{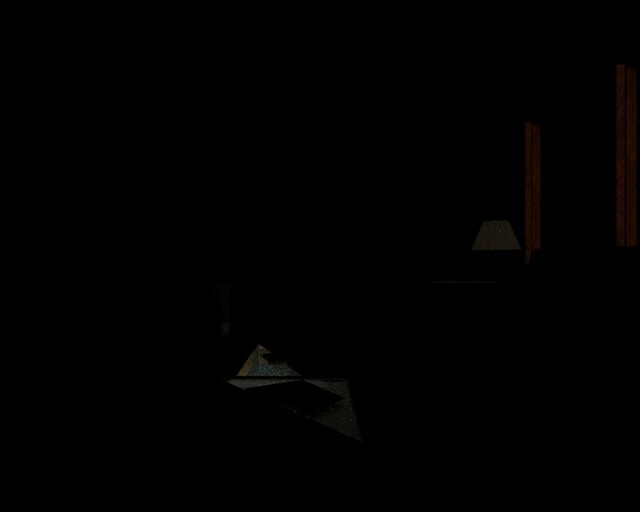}
         \caption{Transmission}
         \label{fig:five over x}
     \end{subfigure}
        \caption{Sample from scene 10}
        \label{fig:three graphs}
\end{figure}

\begin{figure}[!ht]
     \centering
     \begin{subfigure}[b]{0.49\linewidth}
         \centering
         \includegraphics[width=1\linewidth]{IMG/supp_mat/supp_mat_11/Image.jpg}
         \caption{Rendered Image}
         \label{fig:y equals x}
     \end{subfigure}
     \begin{subfigure}[b]{0.49\linewidth}
         \centering
         \includegraphics[width=1\linewidth]{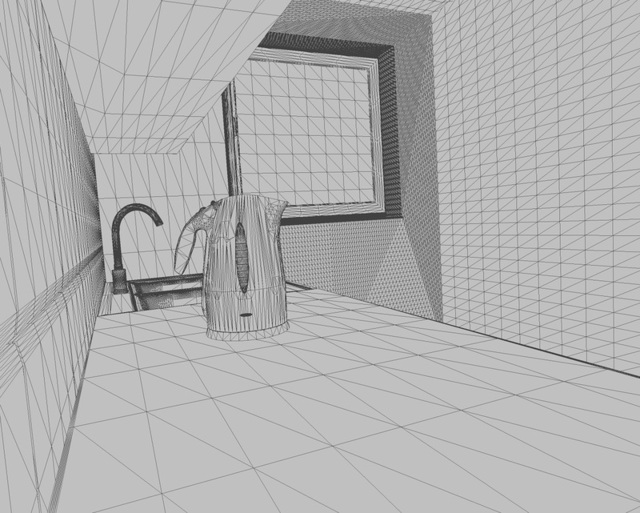}
         \caption{Triangulate mesh}
         \label{fig:y equals x}
     \end{subfigure}
     \hfill
     \begin{subfigure}[b]{0.32\linewidth}
         \centering
         \includegraphics[width=\linewidth]{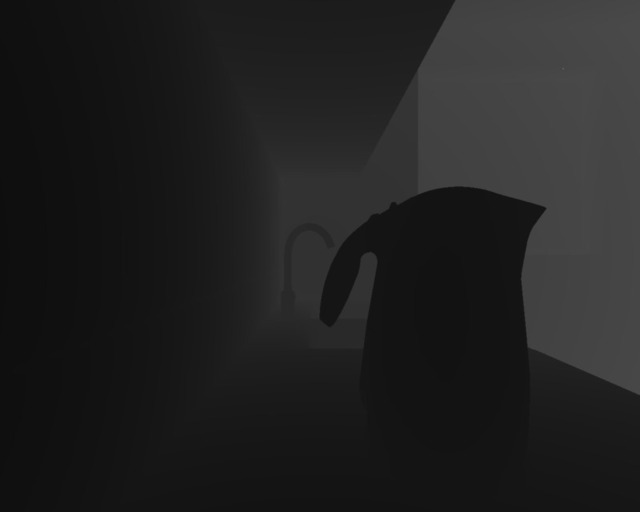}
         \caption{Depth}
         \label{fig:three sin x}
     \end{subfigure}
     \hfill
     \begin{subfigure}[b]{0.32\linewidth}
         \centering
         \includegraphics[width=\linewidth]{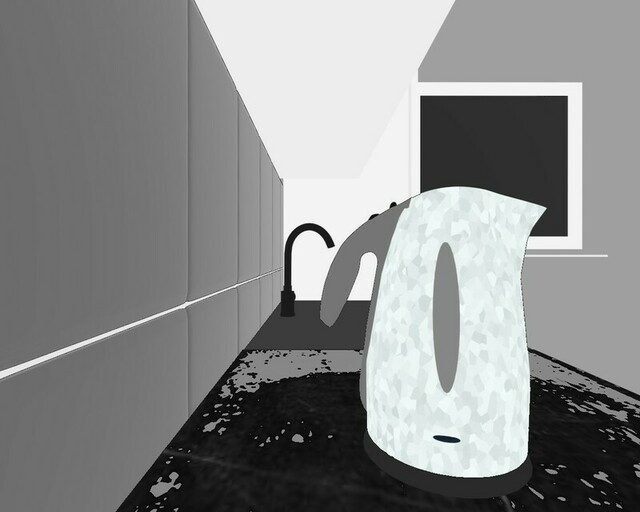}
         \caption{Albedo}
         \label{fig:five over x}
     \end{subfigure}
      \hfill
     \begin{subfigure}[b]{0.32\linewidth}
         \centering
         \includegraphics[width=\linewidth]{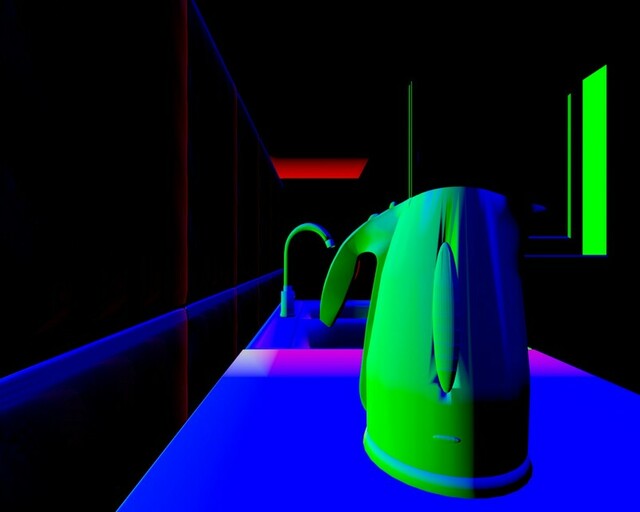}
         \caption{Normals}
         \label{fig:five over x}
     \end{subfigure}
     \begin{subfigure}[b]{0.32\linewidth}
         \centering
         \includegraphics[width=\linewidth]{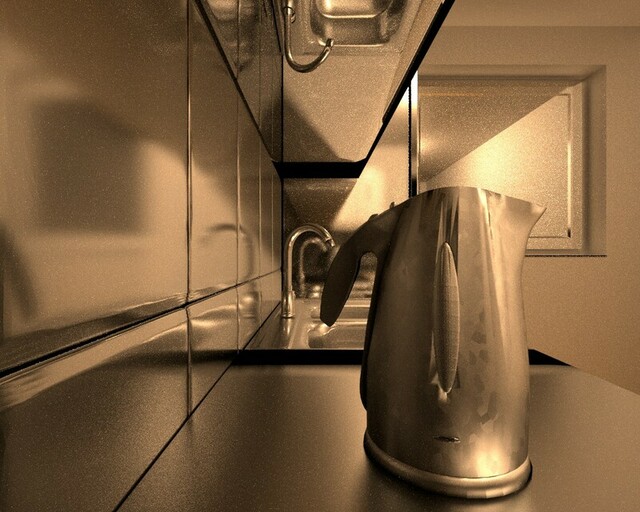}
         \caption{Specular Indirect}
         \label{fig:five over x}
     \end{subfigure}
     \begin{subfigure}[b]{0.32\linewidth}
         \centering
         \includegraphics[width=\linewidth]{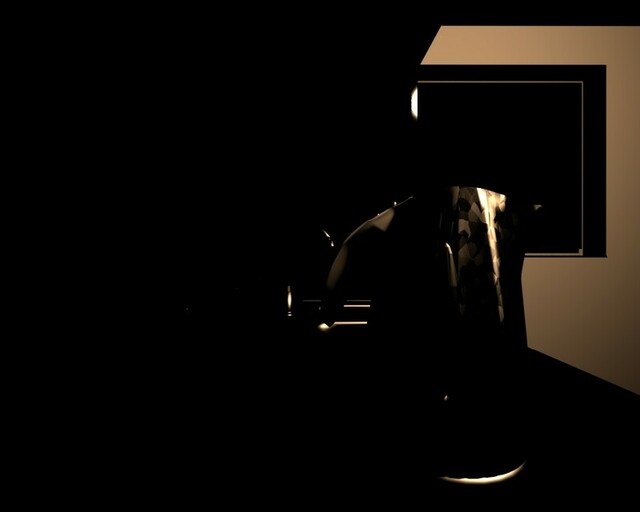}
         \caption{Specular Direct}
         \label{fig:five over x}
     \end{subfigure}
     \begin{subfigure}[b]{0.32\linewidth}
         \centering
         \includegraphics[width=\linewidth]{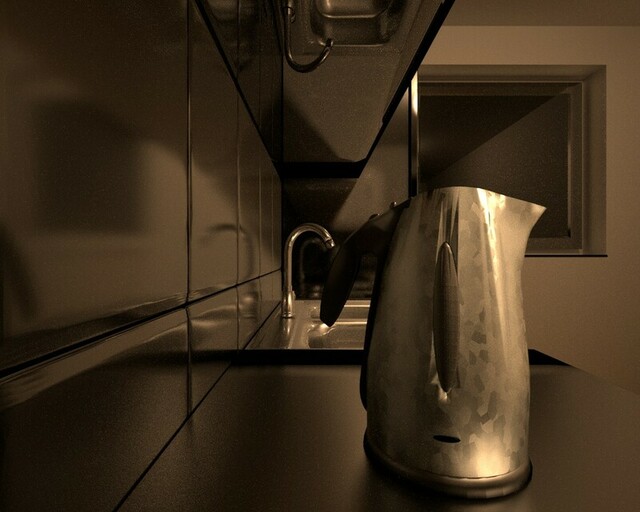}
         \caption{Specular}
         \label{fig:five over x}
     \end{subfigure}
     \begin{subfigure}[b]{0.32\linewidth}
         \centering
         \includegraphics[width=\linewidth]{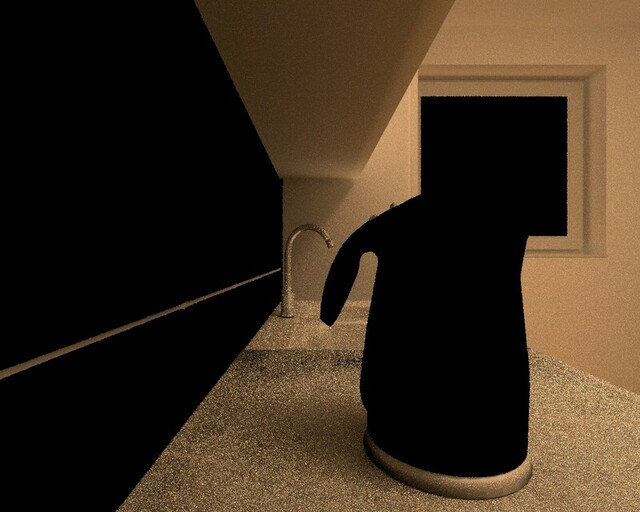}
         \caption{Diffuse Indirect}
         \label{fig:five over x}
     \end{subfigure}
     \begin{subfigure}[b]{0.32\linewidth}
         \centering
         \includegraphics[width=\linewidth]{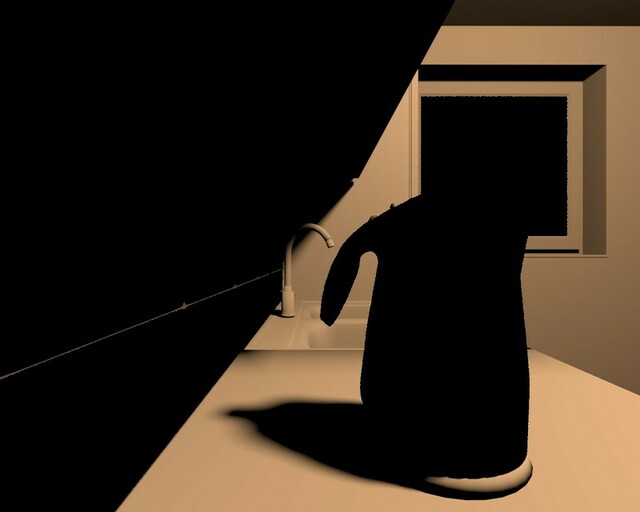}
         \caption{Diffuse Direct}
         \label{fig:five over x}
     \end{subfigure}
     \begin{subfigure}[b]{0.32\linewidth}
         \centering
         \includegraphics[width=\linewidth]{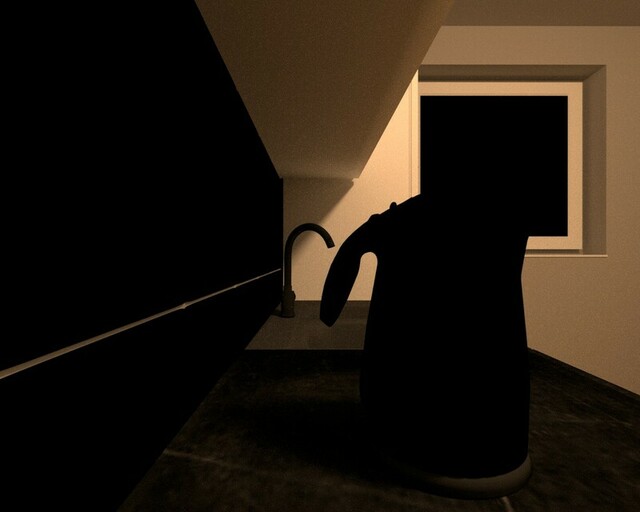}
         \caption{Diffuse}
         \label{fig:five over x}
     \end{subfigure}
     \begin{subfigure}[b]{0.32\linewidth}
         \centering
         \includegraphics[width=\linewidth]{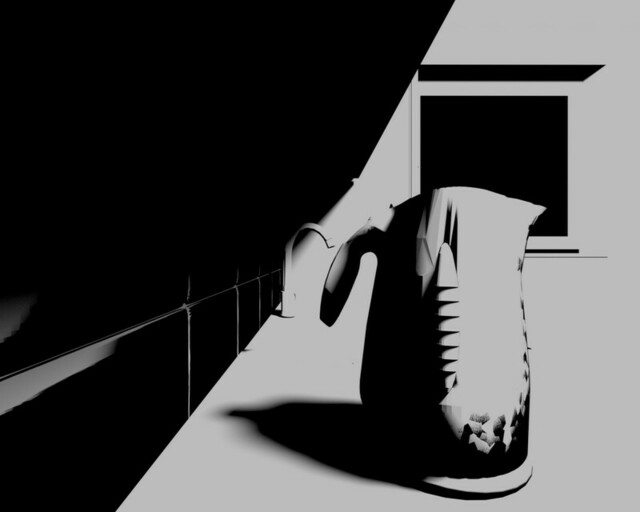}
         \caption{Shadow}
         \label{fig:five over x}
     \end{subfigure}
     \begin{subfigure}[b]{0.32\linewidth}
         \centering
         \includegraphics[width=\linewidth]{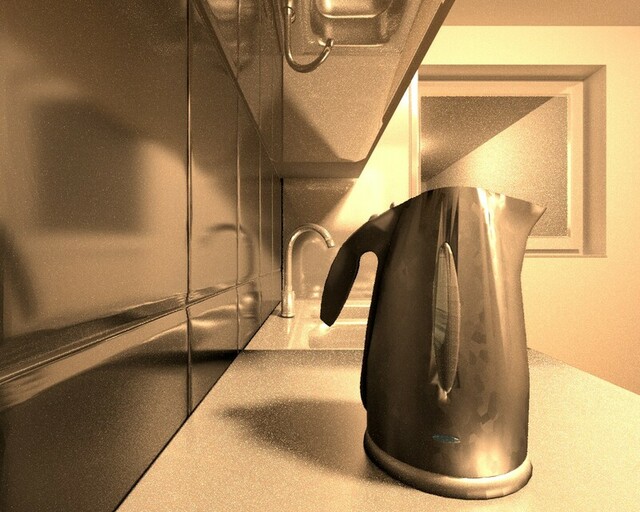}
         \caption{Lightmap}
         \label{fig:five over x}
     \end{subfigure}
     \begin{subfigure}[b]{0.32\linewidth}
         \centering
         \includegraphics[width=\linewidth]{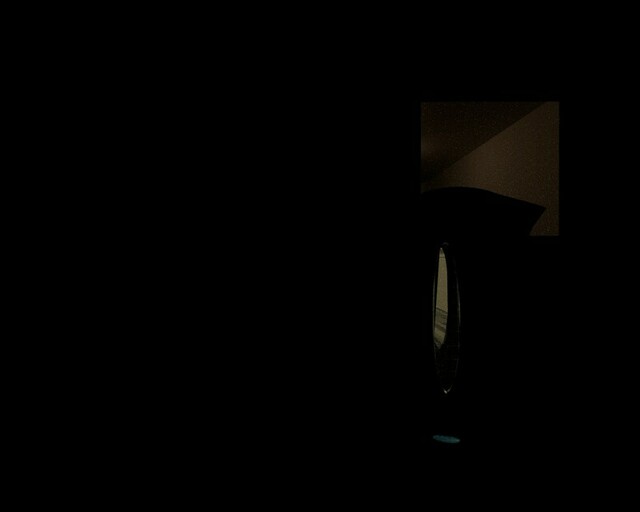}
         \caption{Transmission}
         \label{fig:five over x}
     \end{subfigure}
        \caption{Sample from scene 11}
        \label{fig:three graphs}
\end{figure}

\begin{figure}[!ht]
     \centering
     \begin{subfigure}[b]{0.49\linewidth}
         \centering
         \includegraphics[width=1\linewidth]{IMG/supp_mat/supp_mat_12/Image.jpg}
         \caption{Rendered Image}
         \label{fig:y equals x}
     \end{subfigure}
     \begin{subfigure}[b]{0.49\linewidth}
         \centering
         \includegraphics[width=1\linewidth]{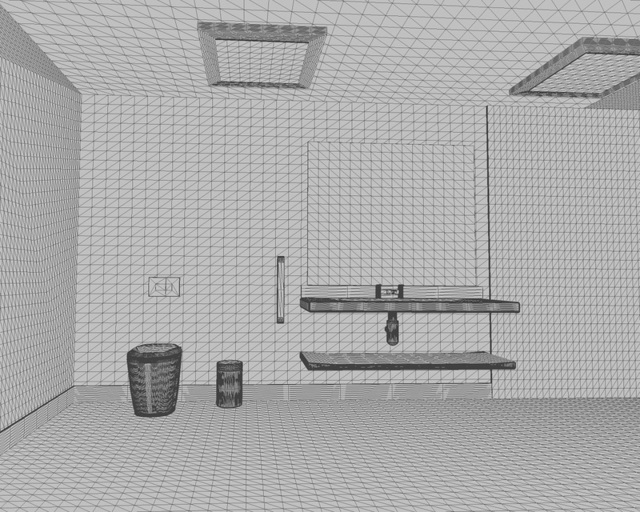}
         \caption{Triangulate mesh}
         \label{fig:y equals x}
     \end{subfigure}
     \hfill
     \begin{subfigure}[b]{0.32\linewidth}
         \centering
         \includegraphics[width=\linewidth]{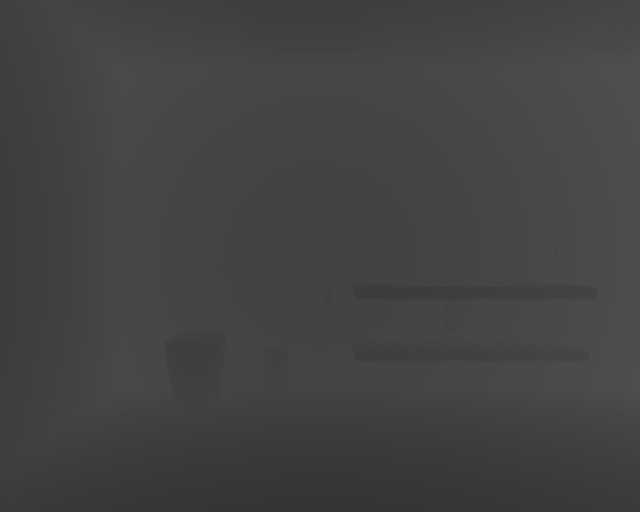}
         \caption{Depth}
         \label{fig:three sin x}
     \end{subfigure}
     \hfill
     \begin{subfigure}[b]{0.32\linewidth}
         \centering
         \includegraphics[width=\linewidth]{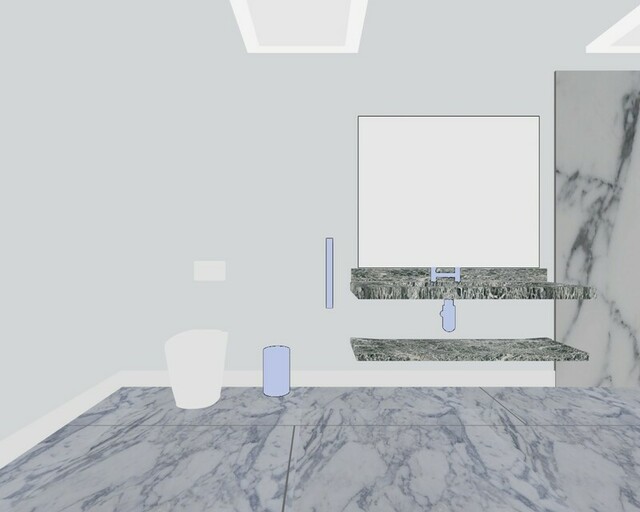}
         \caption{Albedo}
         \label{fig:five over x}
     \end{subfigure}
      \hfill
     \begin{subfigure}[b]{0.32\linewidth}
         \centering
         \includegraphics[width=\linewidth]{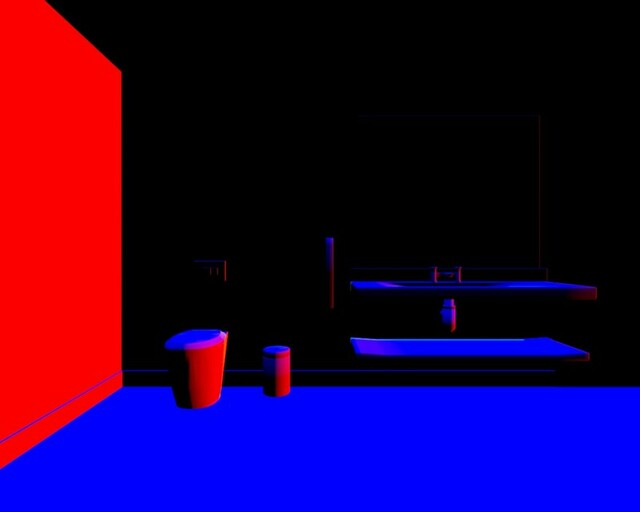}
         \caption{Normals}
         \label{fig:five over x}
     \end{subfigure}
     \begin{subfigure}[b]{0.32\linewidth}
         \centering
         \includegraphics[width=\linewidth]{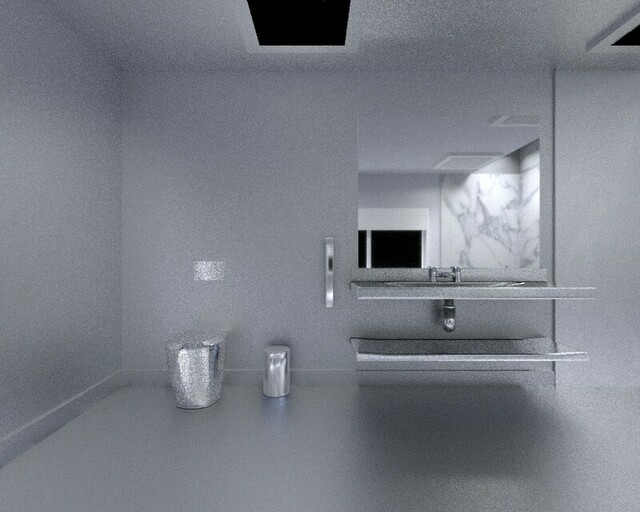}
         \caption{Specular Indirect}
         \label{fig:five over x}
     \end{subfigure}
     \begin{subfigure}[b]{0.32\linewidth}
         \centering
         \includegraphics[width=\linewidth]{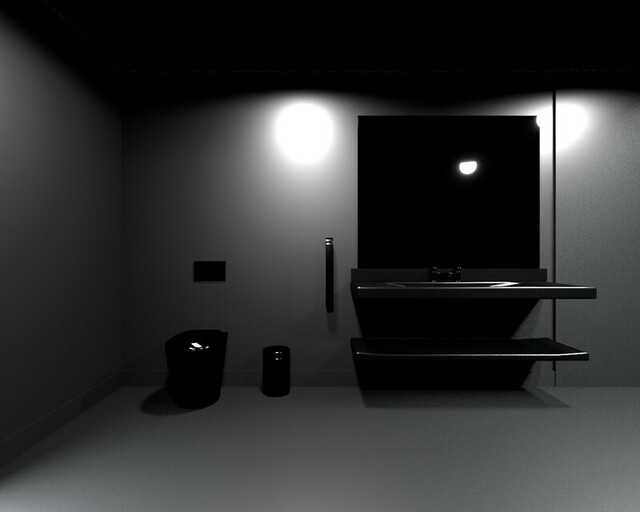}
         \caption{Specular Direct}
         \label{fig:five over x}
     \end{subfigure}
     \begin{subfigure}[b]{0.32\linewidth}
         \centering
         \includegraphics[width=\linewidth]{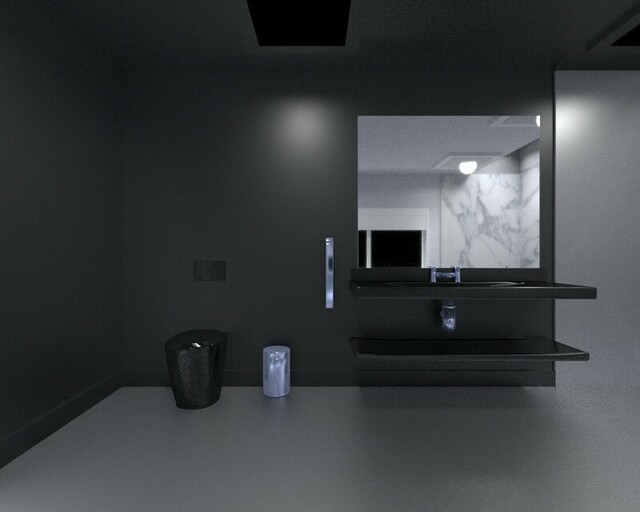}
         \caption{Specular}
         \label{fig:five over x}
     \end{subfigure}
     \begin{subfigure}[b]{0.32\linewidth}
         \centering
         \includegraphics[width=\linewidth]{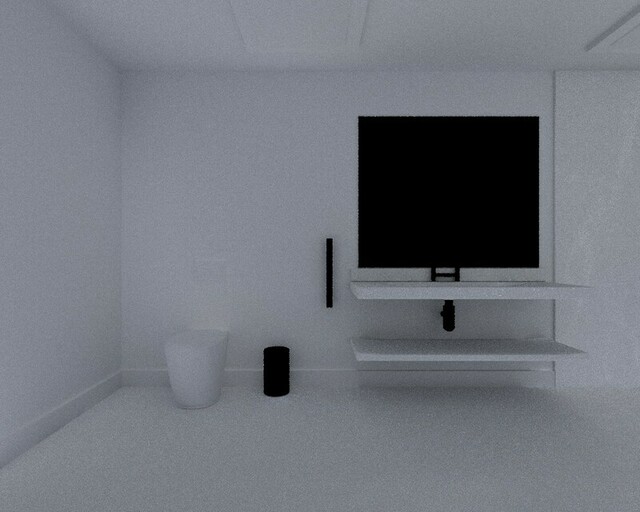}
         \caption{Diffuse Indirect}
         \label{fig:five over x}
     \end{subfigure}
     \begin{subfigure}[b]{0.32\linewidth}
         \centering
         \includegraphics[width=\linewidth]{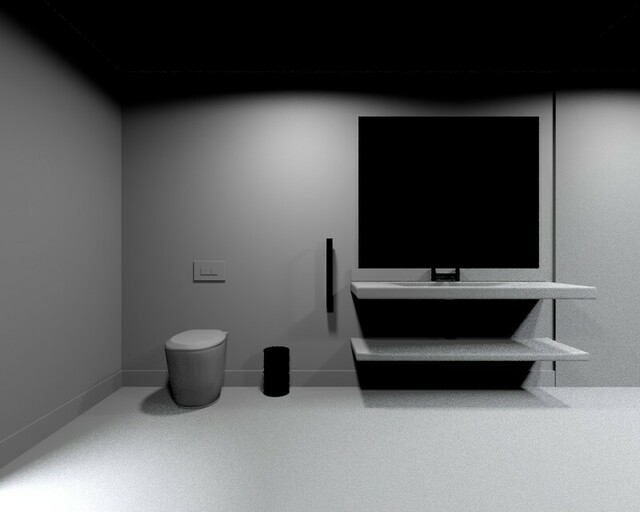}
         \caption{Diffuse Direct}
         \label{fig:five over x}
     \end{subfigure}
     \begin{subfigure}[b]{0.32\linewidth}
         \centering
         \includegraphics[width=\linewidth]{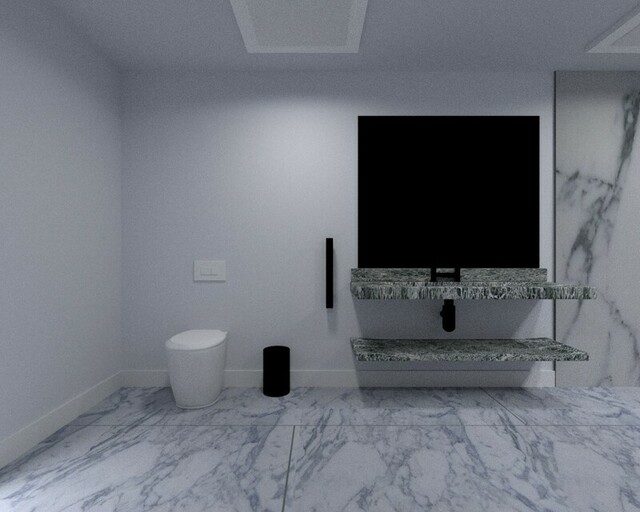}
         \caption{Diffuse}
         \label{fig:five over x}
     \end{subfigure}
     \begin{subfigure}[b]{0.32\linewidth}
         \centering
         \includegraphics[width=\linewidth]{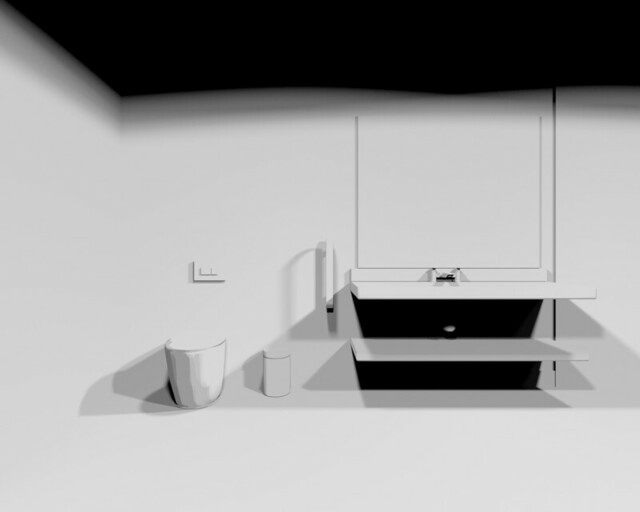}
         \caption{Shadow}
         \label{fig:five over x}
     \end{subfigure}
     \begin{subfigure}[b]{0.32\linewidth}
         \centering
         \includegraphics[width=\linewidth]{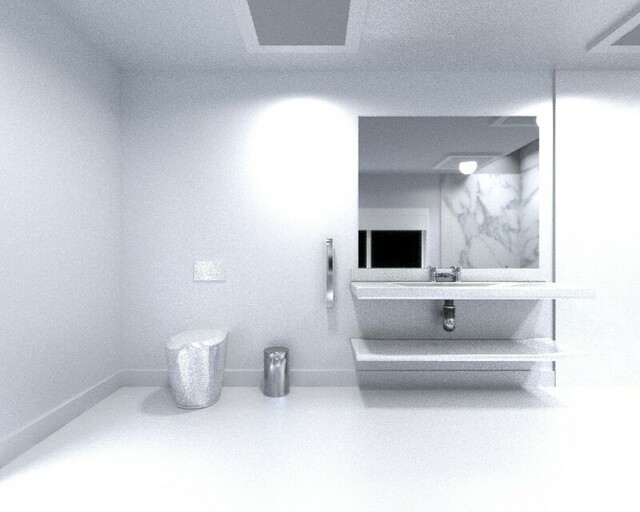}
         \caption{Lightmap}
         \label{fig:five over x}
     \end{subfigure}
     \begin{subfigure}[b]{0.32\linewidth}
         \centering
         \includegraphics[width=\linewidth]{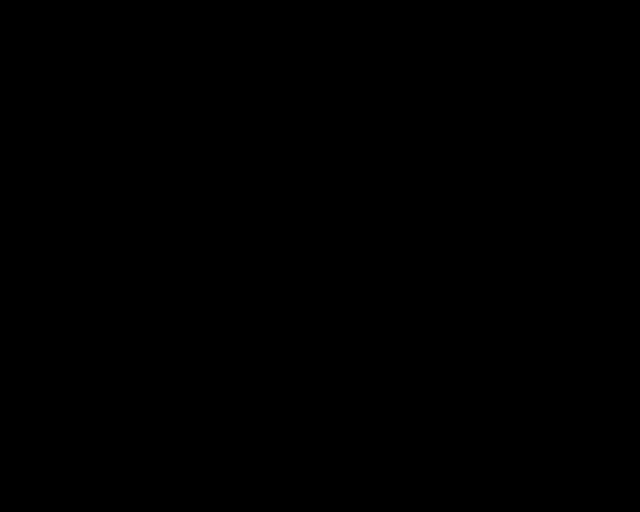}
         \caption{Transmission}
         \label{fig:five over x}
     \end{subfigure}
        \caption{Sample from scene 12}
        \label{fig:three graphs}
\end{figure}

\begin{figure}[!ht]
     \centering
     \begin{subfigure}[b]{0.49\linewidth}
         \centering
         \includegraphics[width=1\linewidth]{IMG/supp_mat/supp_mat_13/Image.jpg}
         \caption{Rendered Image}
         \label{fig:y equals x}
     \end{subfigure}
     \begin{subfigure}[b]{0.49\linewidth}
         \centering
         \includegraphics[width=1\linewidth]{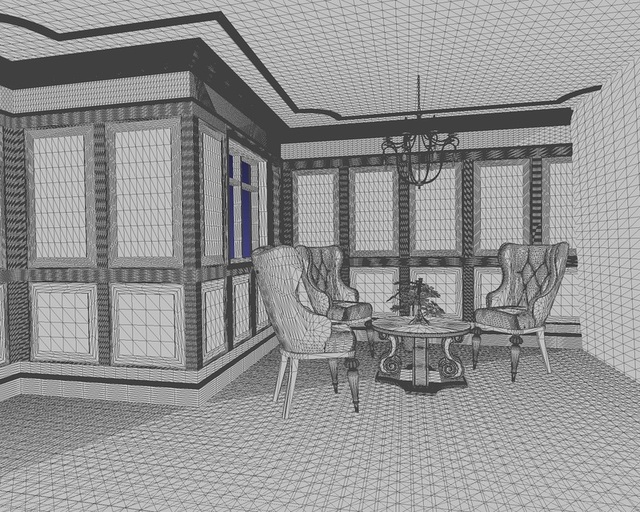}
         \caption{Triangulate mesh}
         \label{fig:y equals x}
     \end{subfigure}
     \hfill
     \begin{subfigure}[b]{0.32\linewidth}
         \centering
         \includegraphics[width=\linewidth]{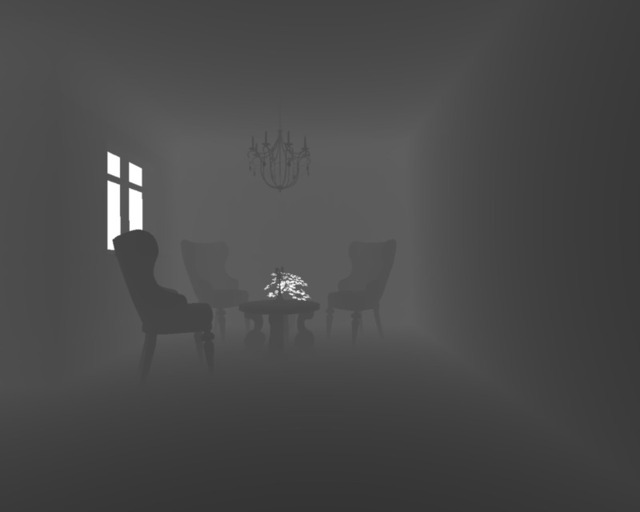}
         \caption{Depth}
         \label{fig:three sin x}
     \end{subfigure}
     \hfill
     \begin{subfigure}[b]{0.32\linewidth}
         \centering
         \includegraphics[width=\linewidth]{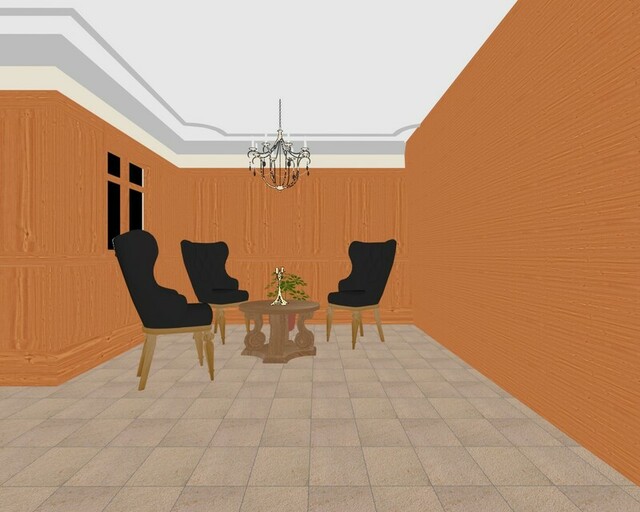}
         \caption{Albedo}
         \label{fig:five over x}
     \end{subfigure}
      \hfill
     \begin{subfigure}[b]{0.32\linewidth}
         \centering
         \includegraphics[width=\linewidth]{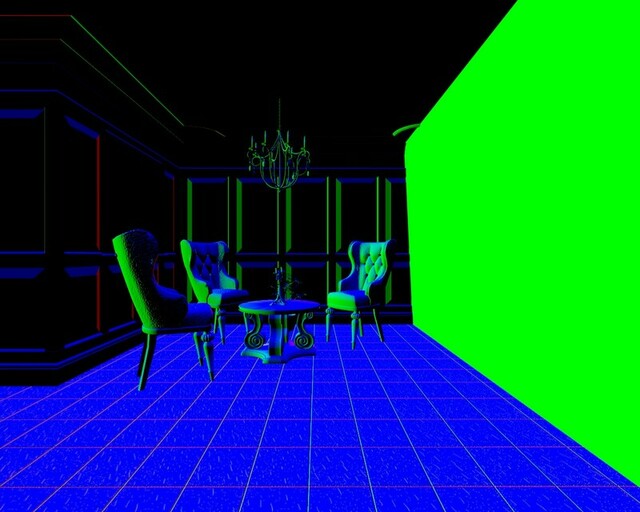}
         \caption{Normals}
         \label{fig:five over x}
     \end{subfigure}
     \begin{subfigure}[b]{0.32\linewidth}
         \centering
         \includegraphics[width=\linewidth]{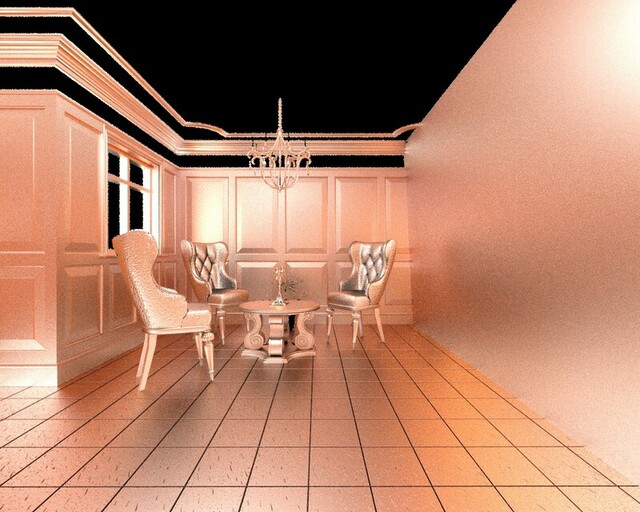}
         \caption{Specular Indirect}
         \label{fig:five over x}
     \end{subfigure}
     \begin{subfigure}[b]{0.32\linewidth}
         \centering
         \includegraphics[width=\linewidth]{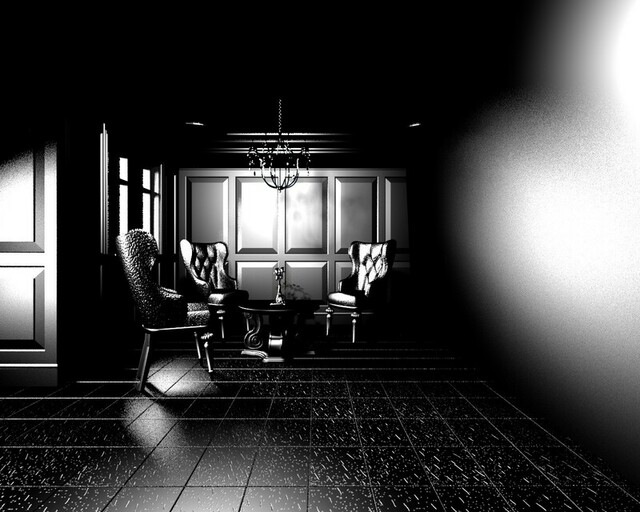}
         \caption{Specular Direct}
         \label{fig:five over x}
     \end{subfigure}
     \begin{subfigure}[b]{0.32\linewidth}
         \centering
         \includegraphics[width=\linewidth]{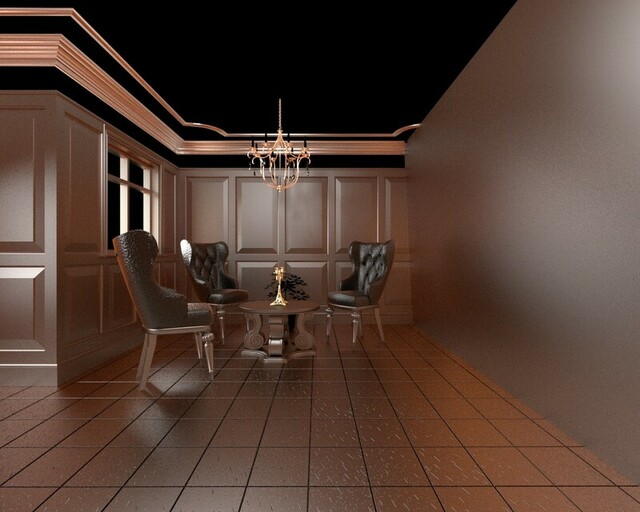}
         \caption{Specular}
         \label{fig:five over x}
     \end{subfigure}
     \begin{subfigure}[b]{0.32\linewidth}
         \centering
         \includegraphics[width=\linewidth]{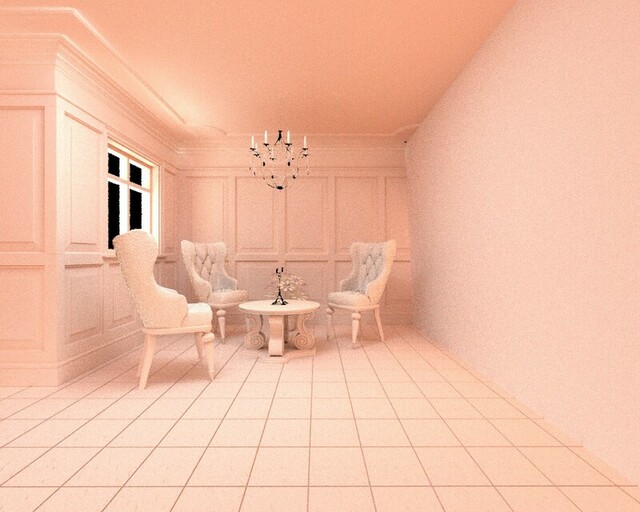}
         \caption{Diffuse Indirect}
         \label{fig:five over x}
     \end{subfigure}
     \begin{subfigure}[b]{0.32\linewidth}
         \centering
         \includegraphics[width=\linewidth]{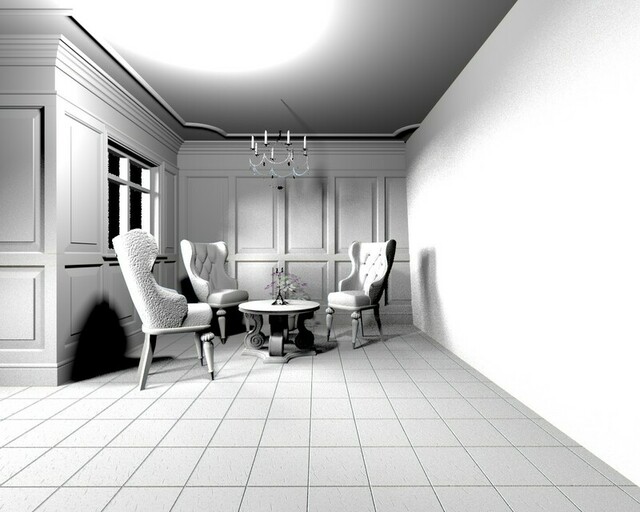}
         \caption{Diffuse Direct}
         \label{fig:five over x}
     \end{subfigure}
     \begin{subfigure}[b]{0.32\linewidth}
         \centering
         \includegraphics[width=\linewidth]{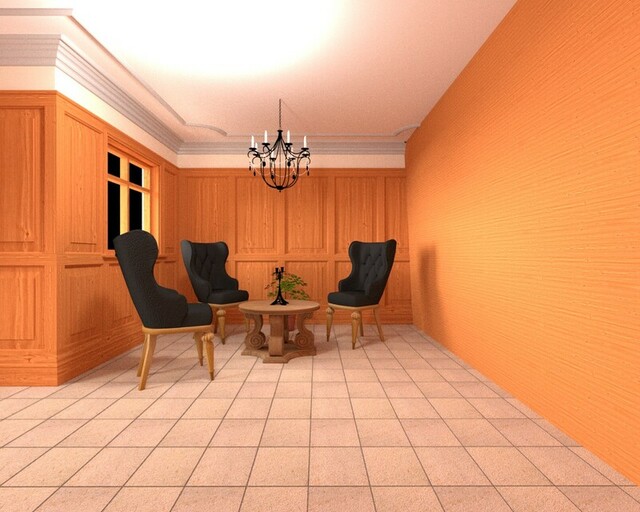}
         \caption{Diffuse}
         \label{fig:five over x}
     \end{subfigure}
     \begin{subfigure}[b]{0.32\linewidth}
         \centering
         \includegraphics[width=\linewidth]{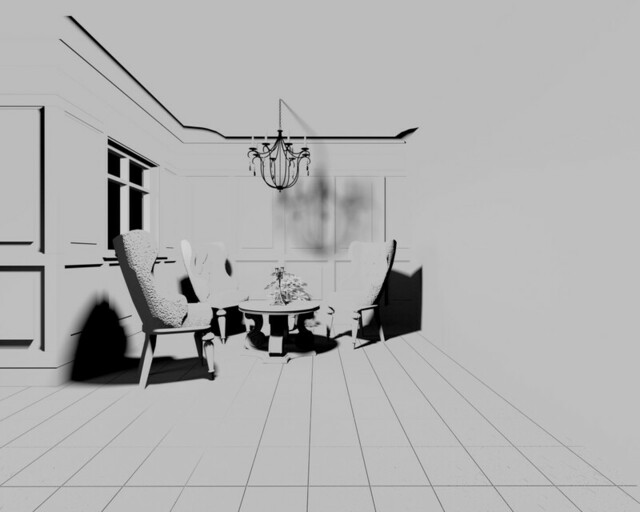}
         \caption{Shadow}
         \label{fig:five over x}
     \end{subfigure}
     \begin{subfigure}[b]{0.32\linewidth}
         \centering
         \includegraphics[width=\linewidth]{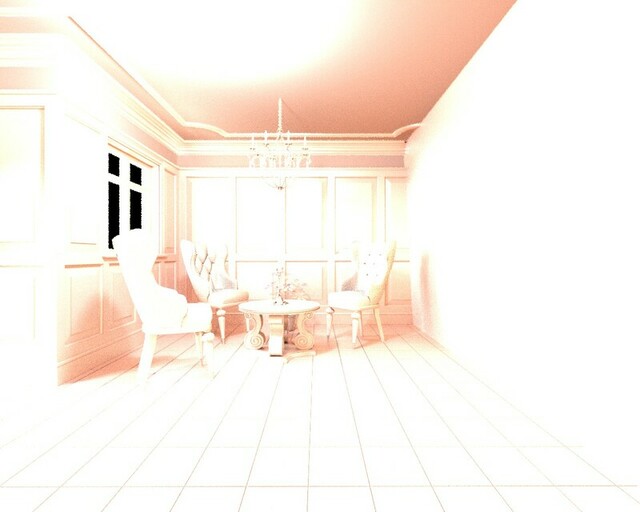}
         \caption{Lightmap}
         \label{fig:five over x}
     \end{subfigure}
     \begin{subfigure}[b]{0.32\linewidth}
         \centering
         \includegraphics[width=\linewidth]{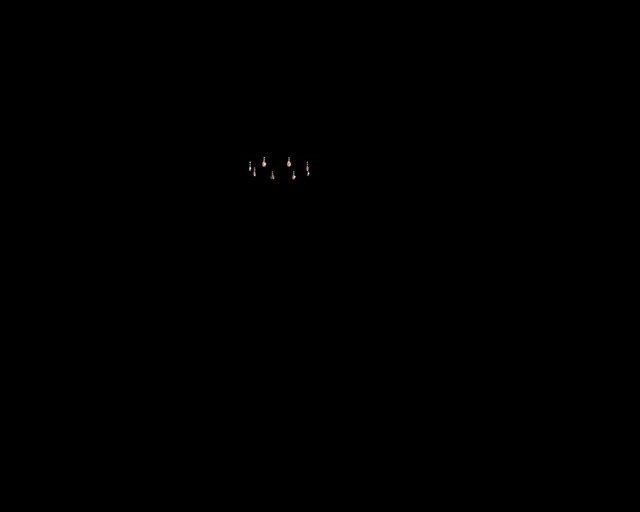}
         \caption{Transmission}
         \label{fig:five over x}
     \end{subfigure}
        \caption{Sample from scene 13}
        \label{fig:three graphs}
\end{figure}

\begin{figure}[!ht]
     \centering
     \begin{subfigure}[b]{0.49\linewidth}
         \centering
         \includegraphics[width=1\linewidth]{IMG/supp_mat/supp_mat_14/Image.jpg}
         \caption{Rendered Image}
         \label{fig:y equals x}
     \end{subfigure}
     \begin{subfigure}[b]{0.49\linewidth}
         \centering
         \includegraphics[width=1\linewidth]{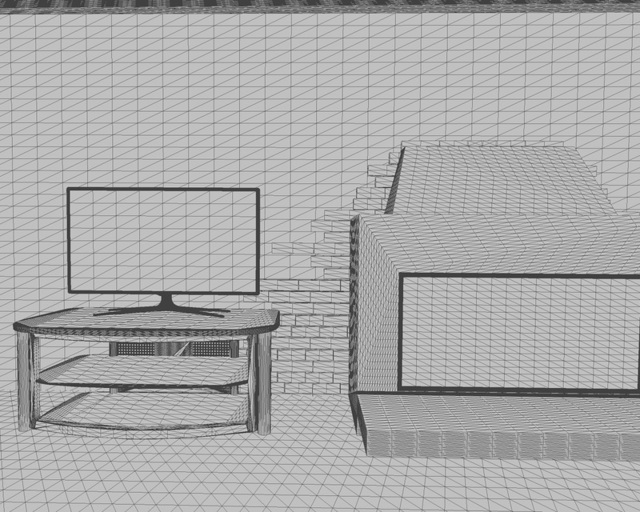}
         \caption{Triangulate mesh}
         \label{fig:y equals x}
     \end{subfigure}
     \hfill
     \begin{subfigure}[b]{0.32\linewidth}
         \centering
         \includegraphics[width=\linewidth]{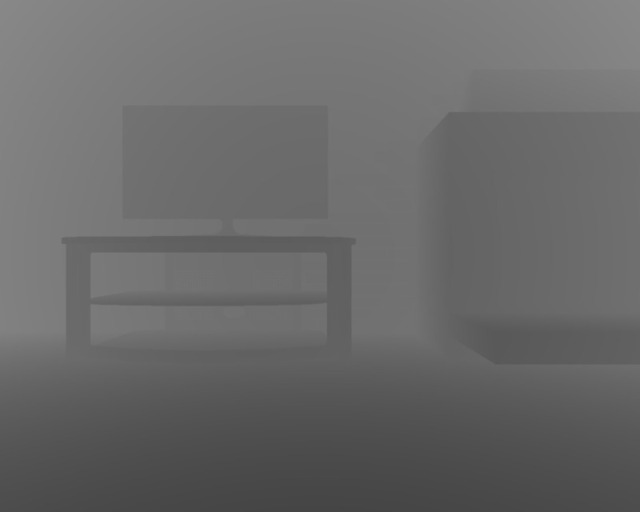}
         \caption{Depth}
         \label{fig:three sin x}
     \end{subfigure}
     \hfill
     \begin{subfigure}[b]{0.32\linewidth}
         \centering
         \includegraphics[width=\linewidth]{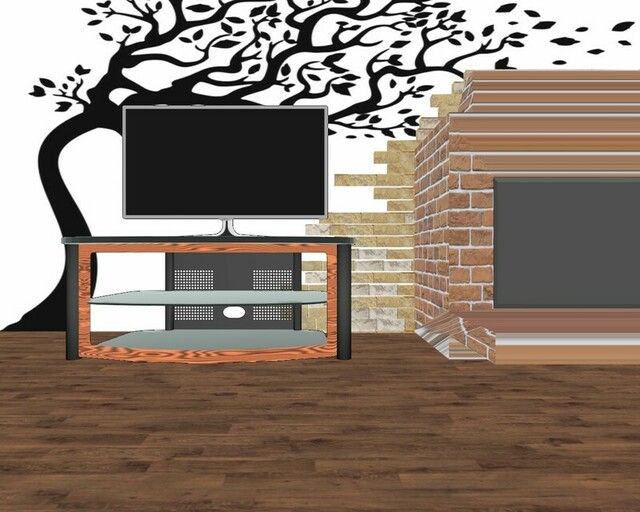}
         \caption{Albedo}
         \label{fig:five over x}
     \end{subfigure}
      \hfill
     \begin{subfigure}[b]{0.32\linewidth}
         \centering
         \includegraphics[width=\linewidth]{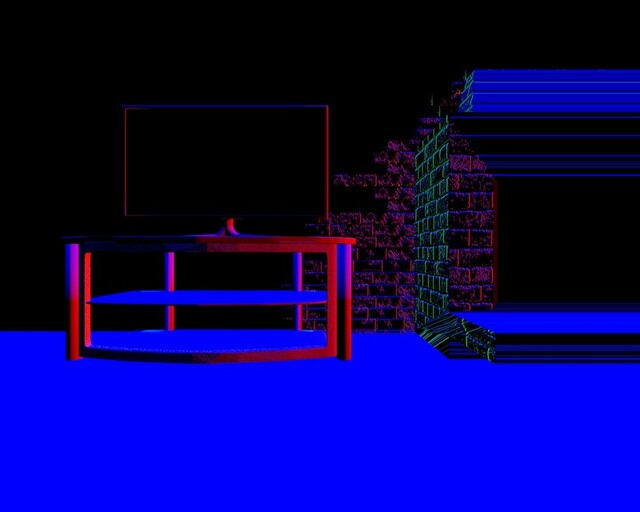}
         \caption{Normals}
         \label{fig:five over x}
     \end{subfigure}
     \begin{subfigure}[b]{0.32\linewidth}
         \centering
         \includegraphics[width=\linewidth]{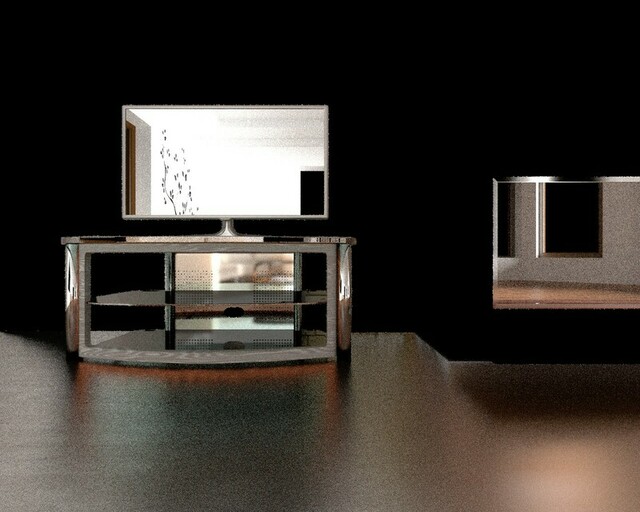}
         \caption{Specular Indirect}
         \label{fig:five over x}
     \end{subfigure}
     \begin{subfigure}[b]{0.32\linewidth}
         \centering
         \includegraphics[width=\linewidth]{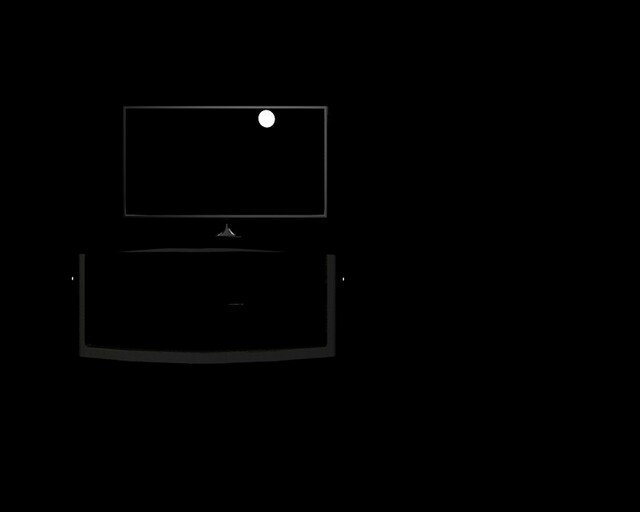}
         \caption{Specular Direct}
         \label{fig:five over x}
     \end{subfigure}
     \begin{subfigure}[b]{0.32\linewidth}
         \centering
         \includegraphics[width=\linewidth]{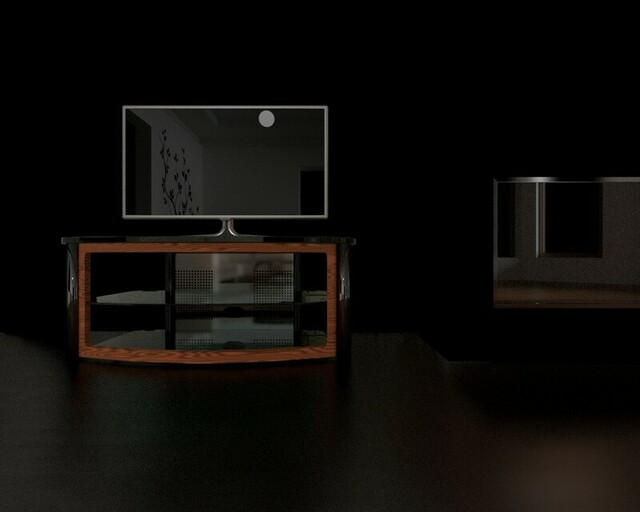}
         \caption{Specular}
         \label{fig:five over x}
     \end{subfigure}
     \begin{subfigure}[b]{0.32\linewidth}
         \centering
         \includegraphics[width=\linewidth]{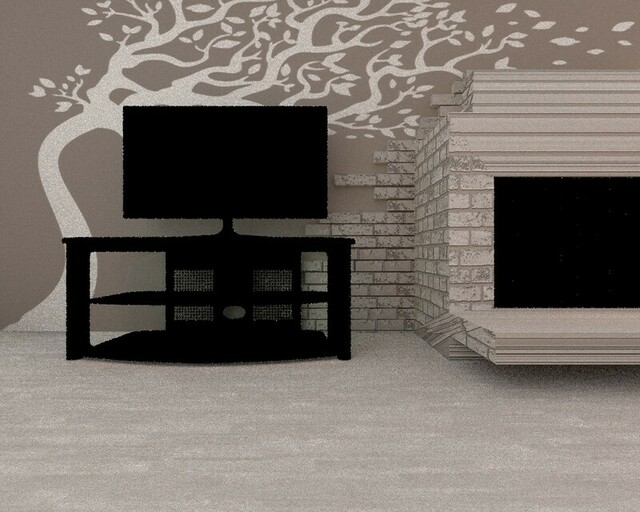}
         \caption{Diffuse Indirect}
         \label{fig:five over x}
     \end{subfigure}
     \begin{subfigure}[b]{0.32\linewidth}
         \centering
         \includegraphics[width=\linewidth]{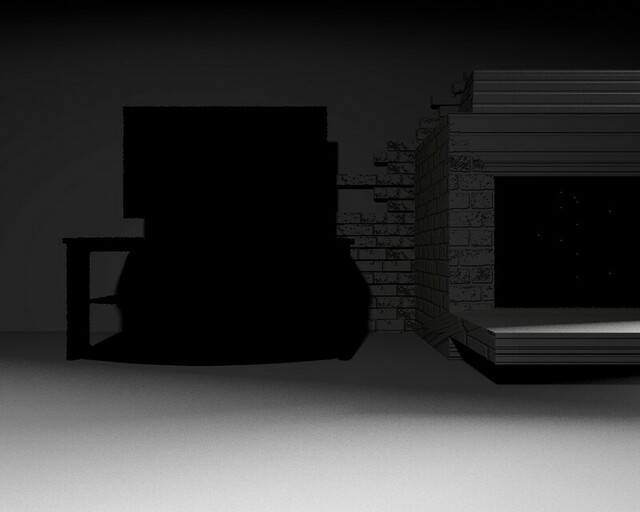}
         \caption{Diffuse Direct}
         \label{fig:five over x}
     \end{subfigure}
     \begin{subfigure}[b]{0.32\linewidth}
         \centering
         \includegraphics[width=\linewidth]{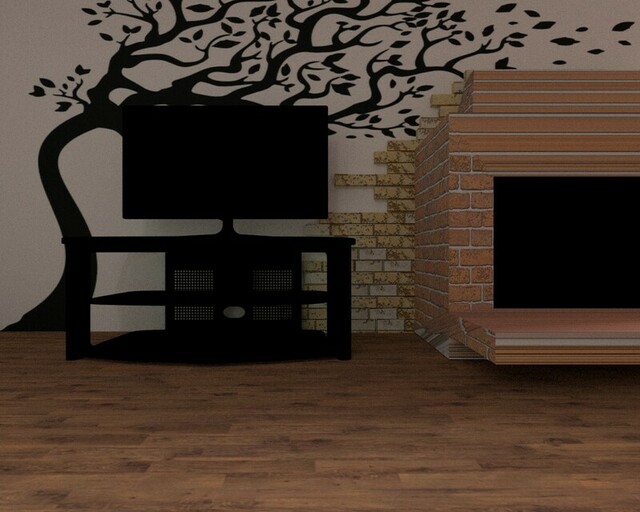}
         \caption{Diffuse}
         \label{fig:five over x}
     \end{subfigure}
     \begin{subfigure}[b]{0.32\linewidth}
         \centering
         \includegraphics[width=\linewidth]{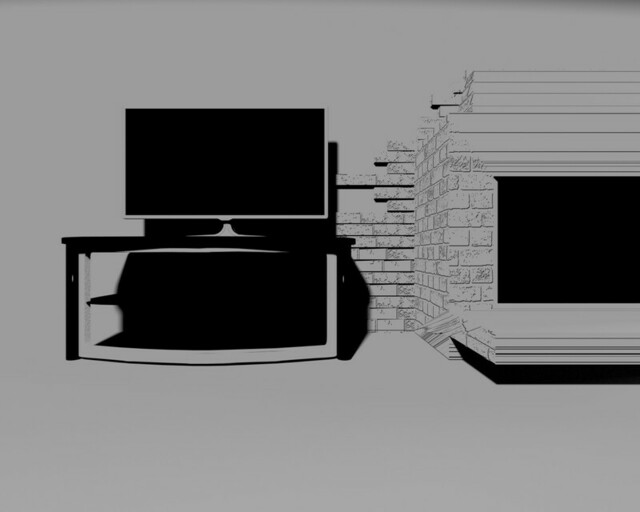}
         \caption{Shadow}
         \label{fig:five over x}
     \end{subfigure}
     \begin{subfigure}[b]{0.32\linewidth}
         \centering
         \includegraphics[width=\linewidth]{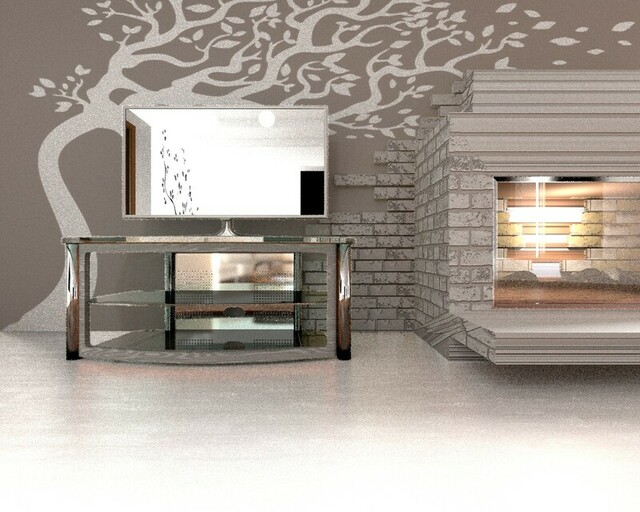}
         \caption{Lightmap}
         \label{fig:five over x}
     \end{subfigure}
     \begin{subfigure}[b]{0.32\linewidth}
         \centering
         \includegraphics[width=\linewidth]{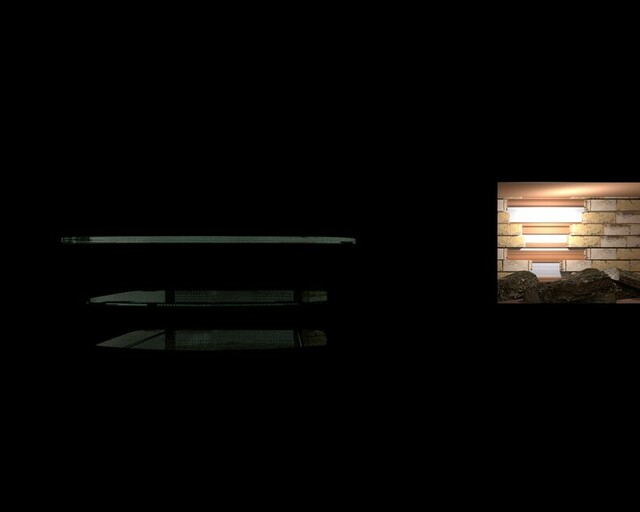}
         \caption{Transmission}
         \label{fig:five over x}
     \end{subfigure}
        \caption{Sample from scene 14}
        \label{fig:three graphs}
\end{figure}

\begin{figure}[!ht]
     \centering
     \begin{subfigure}[b]{0.49\linewidth}
         \centering
         \includegraphics[width=1\linewidth]{IMG/supp_mat/supp_mat_15/Image.jpg}
         \caption{Rendered Image}
         \label{fig:y equals x}
     \end{subfigure}
     \begin{subfigure}[b]{0.49\linewidth}
         \centering
         \includegraphics[width=1\linewidth]{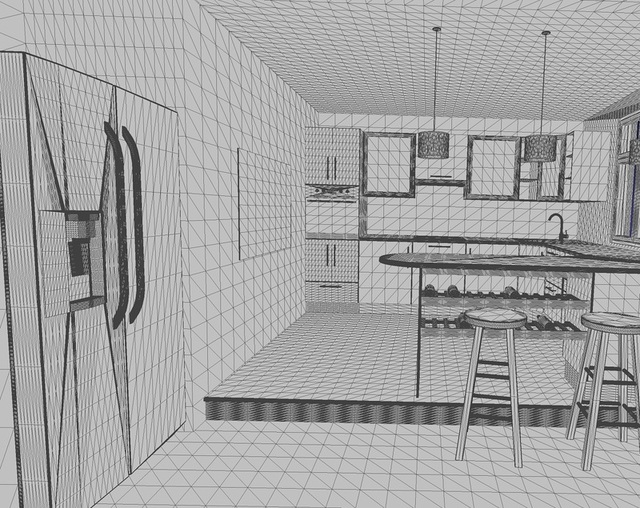}
         \caption{Triangulate mesh}
         \label{fig:y equals x}
     \end{subfigure}
     \hfill
     \begin{subfigure}[b]{0.32\linewidth}
         \centering
         \includegraphics[width=\linewidth]{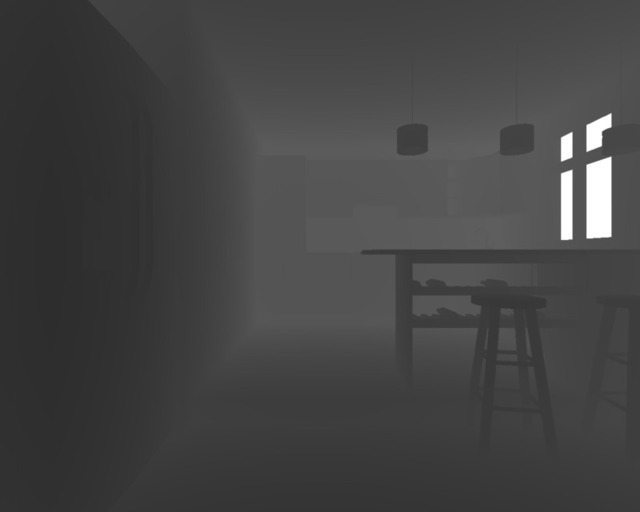}
         \caption{Depth}
         \label{fig:three sin x}
     \end{subfigure}
     \hfill
     \begin{subfigure}[b]{0.32\linewidth}
         \centering
         \includegraphics[width=\linewidth]{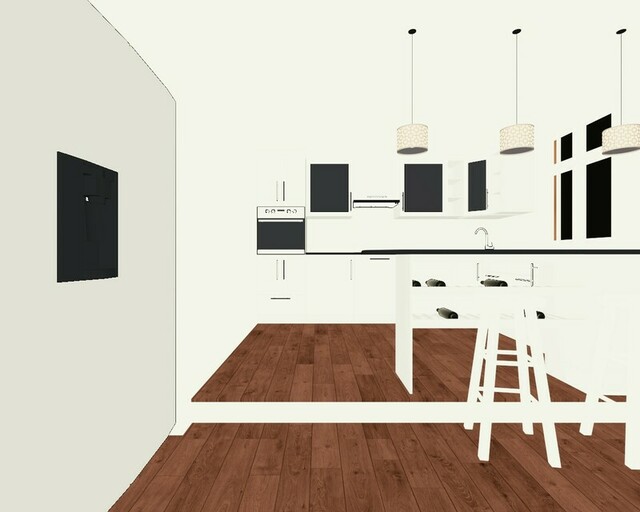}
         \caption{Albedo}
         \label{fig:five over x}
     \end{subfigure}
      \hfill
     \begin{subfigure}[b]{0.32\linewidth}
         \centering
         \includegraphics[width=\linewidth]{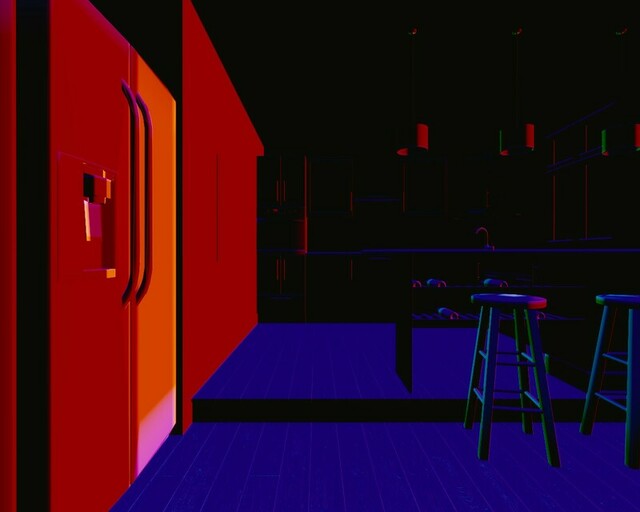}
         \caption{Normals}
         \label{fig:five over x}
     \end{subfigure}
     \begin{subfigure}[b]{0.32\linewidth}
         \centering
         \includegraphics[width=\linewidth]{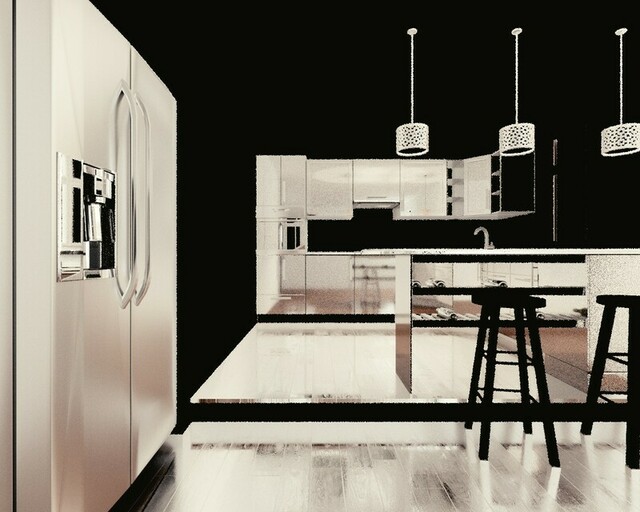}
         \caption{Specular Indirect}
         \label{fig:five over x}
     \end{subfigure}
     \begin{subfigure}[b]{0.32\linewidth}
         \centering
         \includegraphics[width=\linewidth]{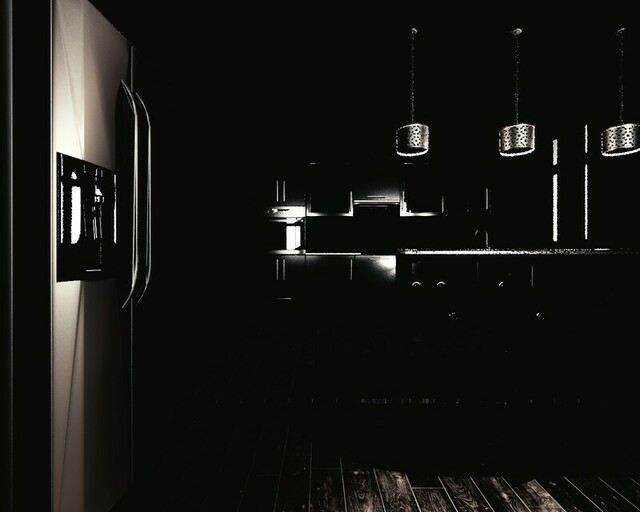}
         \caption{Specular Direct}
         \label{fig:five over x}
     \end{subfigure}
     \begin{subfigure}[b]{0.32\linewidth}
         \centering
         \includegraphics[width=\linewidth]{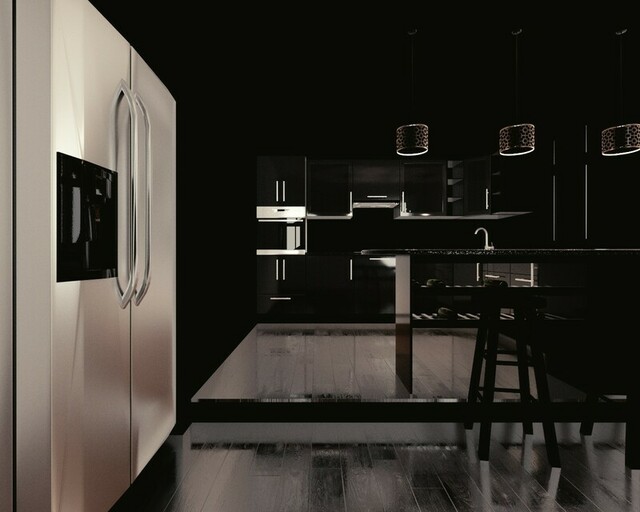}
         \caption{Specular}
         \label{fig:five over x}
     \end{subfigure}
     \begin{subfigure}[b]{0.32\linewidth}
         \centering
         \includegraphics[width=\linewidth]{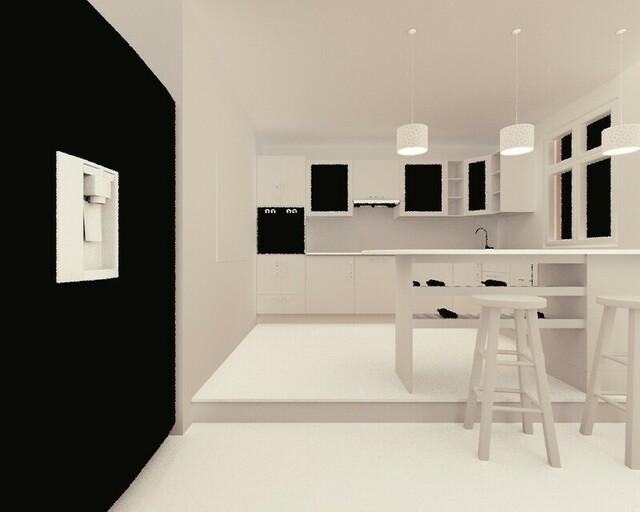}
         \caption{Diffuse Indirect}
         \label{fig:five over x}
     \end{subfigure}
     \begin{subfigure}[b]{0.32\linewidth}
         \centering
         \includegraphics[width=\linewidth]{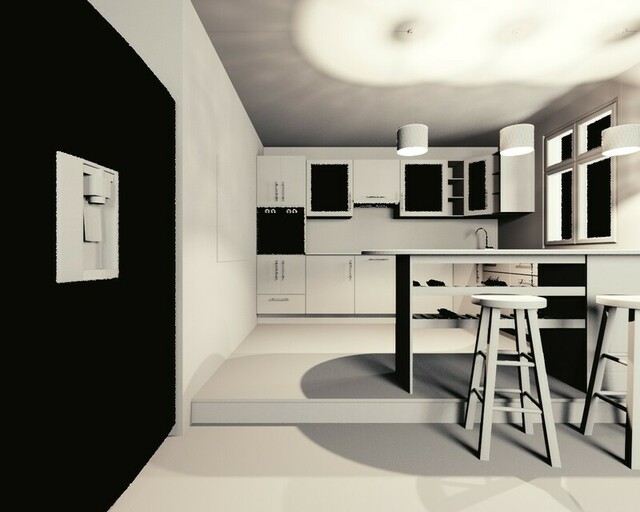}
         \caption{Diffuse Direct}
         \label{fig:five over x}
     \end{subfigure}
     \begin{subfigure}[b]{0.32\linewidth}
         \centering
         \includegraphics[width=\linewidth]{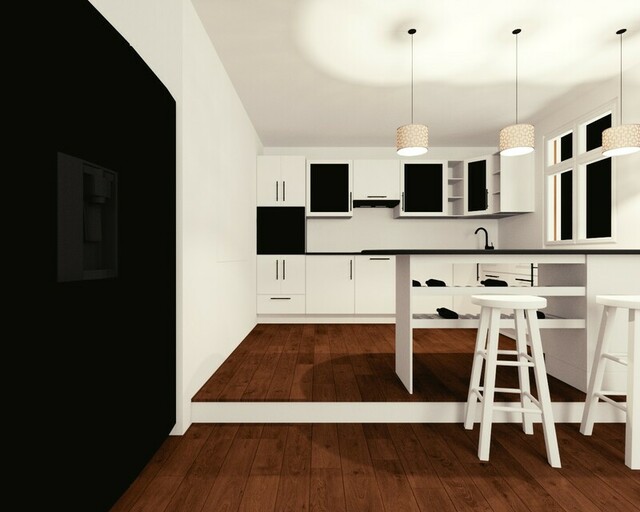}
         \caption{Diffuse}
         \label{fig:five over x}
     \end{subfigure}
     \begin{subfigure}[b]{0.32\linewidth}
         \centering
         \includegraphics[width=\linewidth]{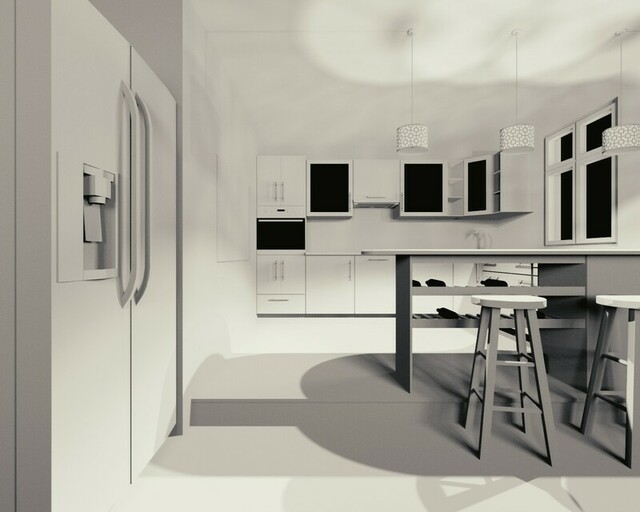}
         \caption{Shadow}
         \label{fig:five over x}
     \end{subfigure}
     \begin{subfigure}[b]{0.32\linewidth}
         \centering
         \includegraphics[width=\linewidth]{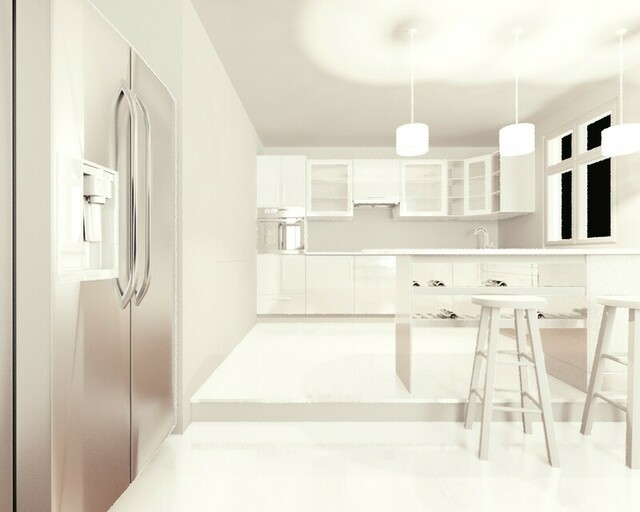}
         \caption{Lightmap}
         \label{fig:five over x}
     \end{subfigure}
     \begin{subfigure}[b]{0.32\linewidth}
         \centering
         \includegraphics[width=\linewidth]{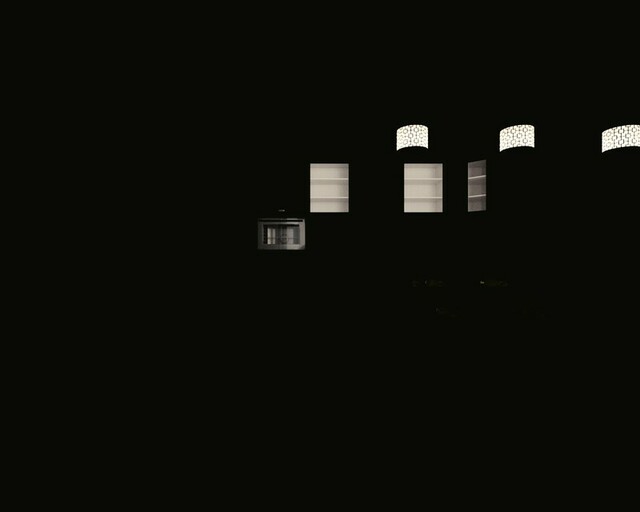}
         \caption{Transmission}
         \label{fig:five over x}
     \end{subfigure}
        \caption{Sample from scene 15}
        \label{fig:three graphs}
\end{figure}

\begin{figure}[!ht]
     \centering
     \begin{subfigure}[b]{0.49\linewidth}
         \centering
         \includegraphics[width=1\linewidth]{IMG/supp_mat/supp_mat_16/Image.jpg}
         \caption{Rendered Image}
         \label{fig:y equals x}
     \end{subfigure}
    \begin{subfigure}[b]{0.49\linewidth}
         \centering
         \includegraphics[width=1\linewidth]{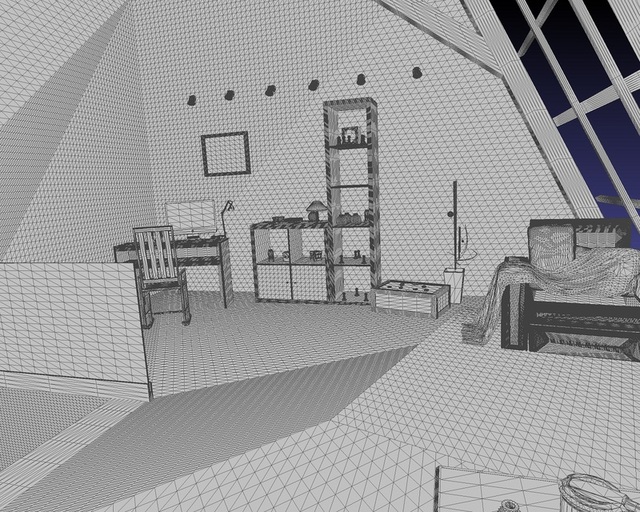}
         \caption{Triangulate mesh}
         \label{fig:y equals x}
     \end{subfigure}
     \hfill
     \begin{subfigure}[b]{0.32\linewidth}
         \centering
         \includegraphics[width=\linewidth]{IMG/supp_mat/supp_mat_16/Depth.jpg}
         \caption{Depth}
         \label{fig:three sin x}
     \end{subfigure}
     \hfill
     \begin{subfigure}[b]{0.32\linewidth}
         \centering
         \includegraphics[width=\linewidth]{IMG/supp_mat/supp_mat_16/Albedo.jpg}
         \caption{Albedo}
         \label{fig:five over x}
     \end{subfigure}
      \hfill
     \begin{subfigure}[b]{0.32\linewidth}
         \centering
         \includegraphics[width=\linewidth]{IMG/supp_mat/supp_mat_16/Normal.jpg}
         \caption{Normals}
         \label{fig:five over x}
     \end{subfigure}
     \begin{subfigure}[b]{0.32\linewidth}
         \centering
         \includegraphics[width=\linewidth]{IMG/supp_mat/supp_mat_16/GlossInd.jpg}
         \caption{Specular Indirect}
         \label{fig:five over x}
     \end{subfigure}
     \begin{subfigure}[b]{0.32\linewidth}
         \centering
         \includegraphics[width=\linewidth]{IMG/supp_mat/supp_mat_16/GlossDir.jpg}
         \caption{Specular Direct}
         \label{fig:five over x}
     \end{subfigure}
     \begin{subfigure}[b]{0.32\linewidth}
         \centering
         \includegraphics[width=\linewidth]{IMG/supp_mat/supp_mat_16/Specular.jpg}
         \caption{Specular}
         \label{fig:five over x}
     \end{subfigure}
     \begin{subfigure}[b]{0.32\linewidth}
         \centering
         \includegraphics[width=\linewidth]{IMG/supp_mat/supp_mat_16/DiffInd.jpg}
         \caption{Diffuse Indirect}
         \label{fig:five over x}
     \end{subfigure}
     \begin{subfigure}[b]{0.32\linewidth}
         \centering
         \includegraphics[width=\linewidth]{IMG/supp_mat/supp_mat_16/DiffDir.jpg}
         \caption{Diffuse Direct}
         \label{fig:five over x}
     \end{subfigure}
     \begin{subfigure}[b]{0.32\linewidth}
         \centering
         \includegraphics[width=\linewidth]{IMG/supp_mat/supp_mat_16/Diffuse.jpg}
         \caption{Diffuse}
         \label{fig:five over x}
     \end{subfigure}
     \begin{subfigure}[b]{0.32\linewidth}
         \centering
         \includegraphics[width=\linewidth]{IMG/supp_mat/supp_mat_16/Shadow.jpg}
         \caption{Shadow}
         \label{fig:five over x}
     \end{subfigure}
     \begin{subfigure}[b]{0.32\linewidth}
         \centering
         \includegraphics[width=\linewidth]{IMG/supp_mat/supp_mat_16/Lightmap.jpg}
         \caption{Lightmap}
         \label{fig:five over x}
     \end{subfigure}
     \begin{subfigure}[b]{0.32\linewidth}
         \centering
         \includegraphics[width=\linewidth]{IMG/supp_mat/supp_mat_16/Transmission.jpg}
         \caption{Transmission}
         \label{fig:five over x}
     \end{subfigure}
        \caption{Sample from scene 16}
        \label{fig:three graphs}
\end{figure}

\begin{figure}[!ht]
     \centering
     \begin{subfigure}[b]{0.49\linewidth}
         \centering
         \includegraphics[width=1\linewidth]{IMG/supp_mat/supp_mat_17/Image.jpg}
         \caption{Rendered Image}
         \label{fig:y equals x}
     \end{subfigure}
     \begin{subfigure}[b]{0.49\linewidth}
         \centering
         \includegraphics[width=1\linewidth]{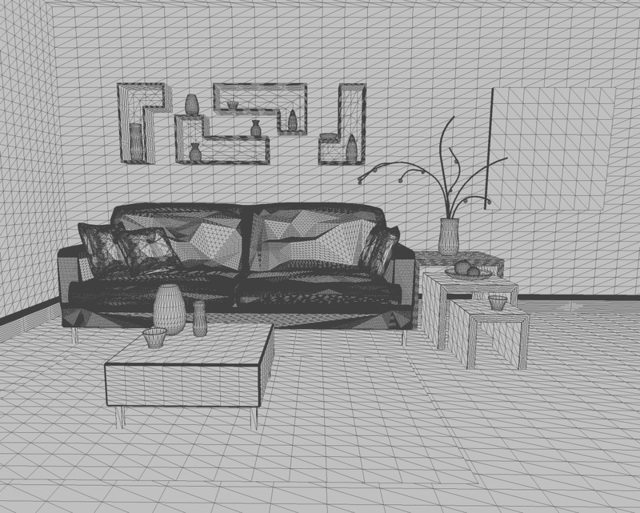}
         \caption{Triangulate mesh}
         \label{fig:y equals x}
     \end{subfigure}
     \hfill
     \begin{subfigure}[b]{0.32\linewidth}
         \centering
         \includegraphics[width=\linewidth]{IMG/supp_mat/supp_mat_17/Depth.jpg}
         \caption{Depth}
         \label{fig:three sin x}
     \end{subfigure}
     \hfill
     \begin{subfigure}[b]{0.32\linewidth}
         \centering
         \includegraphics[width=\linewidth]{IMG/supp_mat/supp_mat_17/Albedo.jpg}
         \caption{Albedo}
         \label{fig:five over x}
     \end{subfigure}
      \hfill
     \begin{subfigure}[b]{0.32\linewidth}
         \centering
         \includegraphics[width=\linewidth]{IMG/supp_mat/supp_mat_17/Normal.jpg}
         \caption{Normals}
         \label{fig:five over x}
     \end{subfigure}
     \begin{subfigure}[b]{0.32\linewidth}
         \centering
         \includegraphics[width=\linewidth]{IMG/supp_mat/supp_mat_17/GlossInd.jpg}
         \caption{Specular Indirect}
         \label{fig:five over x}
     \end{subfigure}
     \begin{subfigure}[b]{0.32\linewidth}
         \centering
         \includegraphics[width=\linewidth]{IMG/supp_mat/supp_mat_17/GlossDir.jpg}
         \caption{Specular Direct}
         \label{fig:five over x}
     \end{subfigure}
     \begin{subfigure}[b]{0.32\linewidth}
         \centering
         \includegraphics[width=\linewidth]{IMG/supp_mat/supp_mat_17/Specular.jpg}
         \caption{Specular}
         \label{fig:five over x}
     \end{subfigure}
     \begin{subfigure}[b]{0.32\linewidth}
         \centering
         \includegraphics[width=\linewidth]{IMG/supp_mat/supp_mat_17/DiffInd.jpg}
         \caption{Diffuse Indirect}
         \label{fig:five over x}
     \end{subfigure}
     \begin{subfigure}[b]{0.32\linewidth}
         \centering
         \includegraphics[width=\linewidth]{IMG/supp_mat/supp_mat_17/DiffDir.jpg}
         \caption{Diffuse Direct}
         \label{fig:five over x}
     \end{subfigure}
     \begin{subfigure}[b]{0.32\linewidth}
         \centering
         \includegraphics[width=\linewidth]{IMG/supp_mat/supp_mat_17/Diffuse.jpg}
         \caption{Diffuse}
         \label{fig:five over x}
     \end{subfigure}
     \begin{subfigure}[b]{0.32\linewidth}
         \centering
         \includegraphics[width=\linewidth]{IMG/supp_mat/supp_mat_17/Shadow.jpg}
         \caption{Shadow}
         \label{fig:five over x}
     \end{subfigure}
     \begin{subfigure}[b]{0.32\linewidth}
         \centering
         \includegraphics[width=\linewidth]{IMG/supp_mat/supp_mat_17/Lightmap.jpg}
         \caption{Lightmap}
         \label{fig:five over x}
     \end{subfigure}
     \begin{subfigure}[b]{0.32\linewidth}
         \centering
         \includegraphics[width=\linewidth]{IMG/supp_mat/supp_mat_17/Transmission.jpg}
         \caption{Transmission}
         \label{fig:five over x}
     \end{subfigure}
        \caption{Sample from scene 17}
        \label{fig:three graphs}
\end{figure}

\begin{figure}[!ht]
     \centering
     \begin{subfigure}[b]{0.49\linewidth}
         \centering
         \includegraphics[width=1\linewidth]{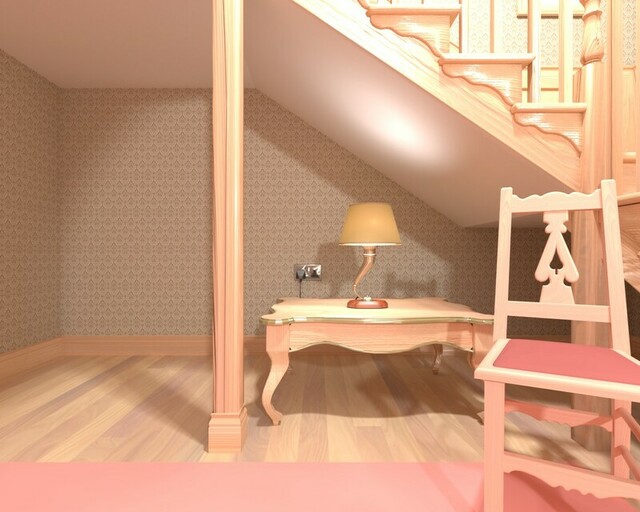}
         \caption{Rendered Image}
         \label{fig:y equals x}
     \end{subfigure}
     \begin{subfigure}[b]{0.49\linewidth}
         \centering
         \includegraphics[width=1\linewidth]{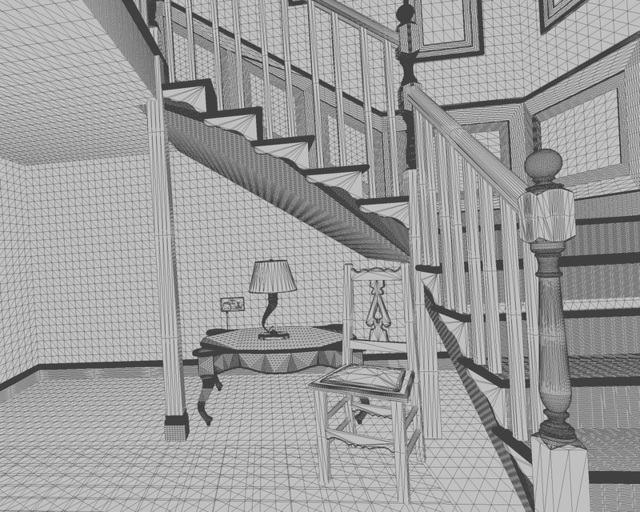}
         \caption{Triangulate mesh}
         \label{fig:y equals x}
     \end{subfigure}
     \hfill
     \begin{subfigure}[b]{0.32\linewidth}
         \centering
         \includegraphics[width=\linewidth]{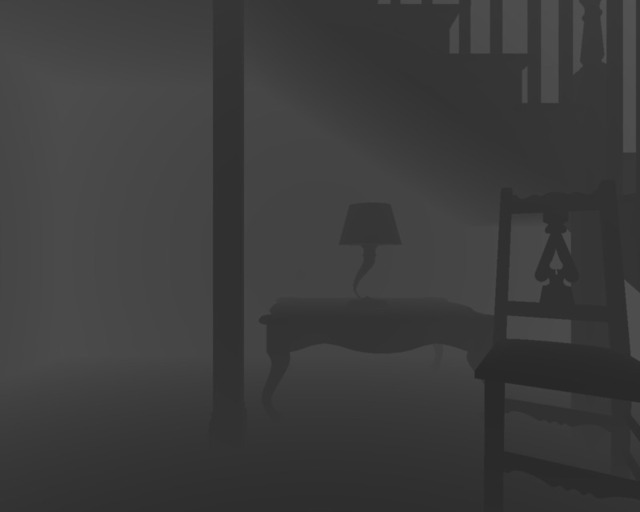}
         \caption{Depth}
         \label{fig:three sin x}
     \end{subfigure}
     \hfill
     \begin{subfigure}[b]{0.32\linewidth}
         \centering
         \includegraphics[width=\linewidth]{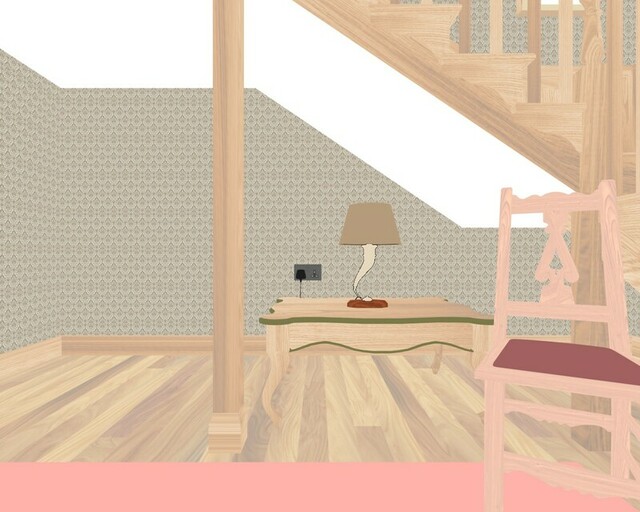}
         \caption{Albedo}
         \label{fig:five over x}
     \end{subfigure}
      \hfill
     \begin{subfigure}[b]{0.32\linewidth}
         \centering
         \includegraphics[width=\linewidth]{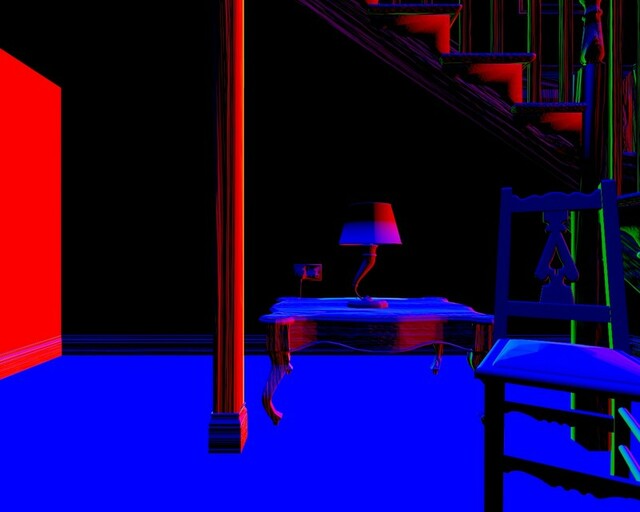}
         \caption{Normals}
         \label{fig:five over x}
     \end{subfigure}
     \begin{subfigure}[b]{0.32\linewidth}
         \centering
         \includegraphics[width=\linewidth]{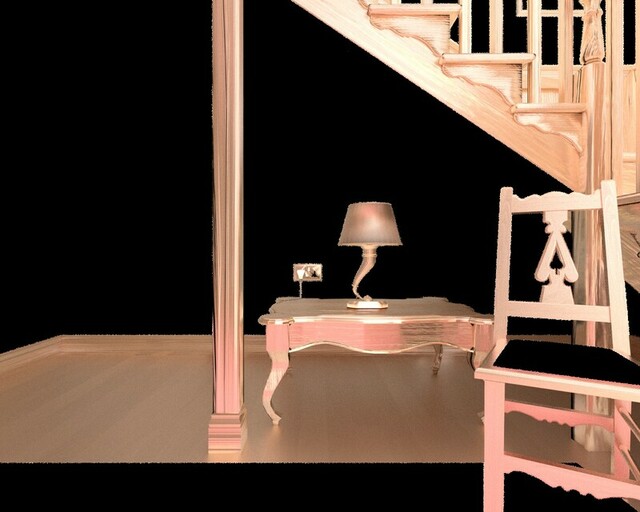}
         \caption{Specular Indirect}
         \label{fig:five over x}
     \end{subfigure}
     \begin{subfigure}[b]{0.32\linewidth}
         \centering
         \includegraphics[width=\linewidth]{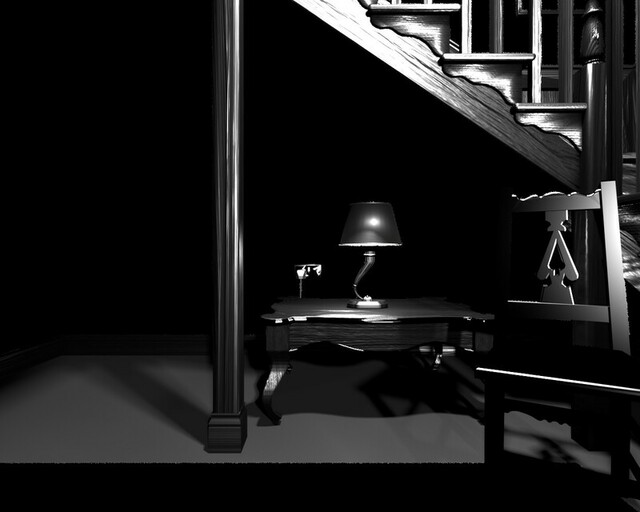}
         \caption{Specular Direct}
         \label{fig:five over x}
     \end{subfigure}
     \begin{subfigure}[b]{0.32\linewidth}
         \centering
         \includegraphics[width=\linewidth]{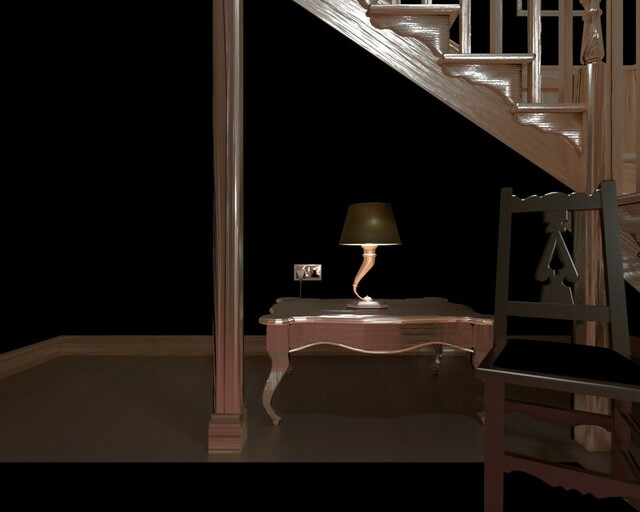}
         \caption{Specular}
         \label{fig:five over x}
     \end{subfigure}
     \begin{subfigure}[b]{0.32\linewidth}
         \centering
         \includegraphics[width=\linewidth]{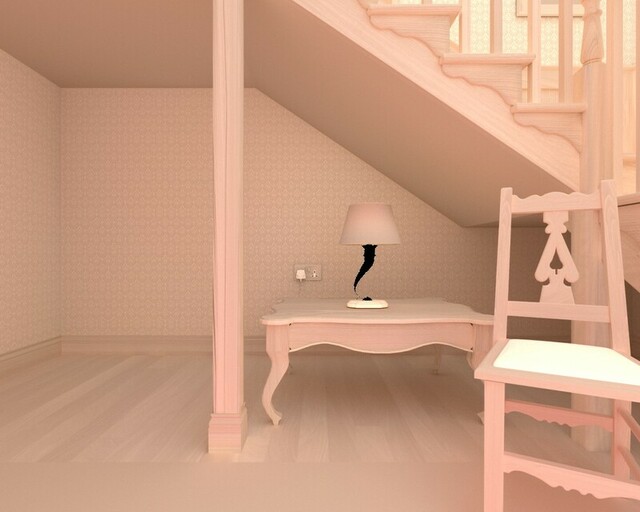}
         \caption{Diffuse Indirect}
         \label{fig:five over x}
     \end{subfigure}
     \begin{subfigure}[b]{0.32\linewidth}
         \centering
         \includegraphics[width=\linewidth]{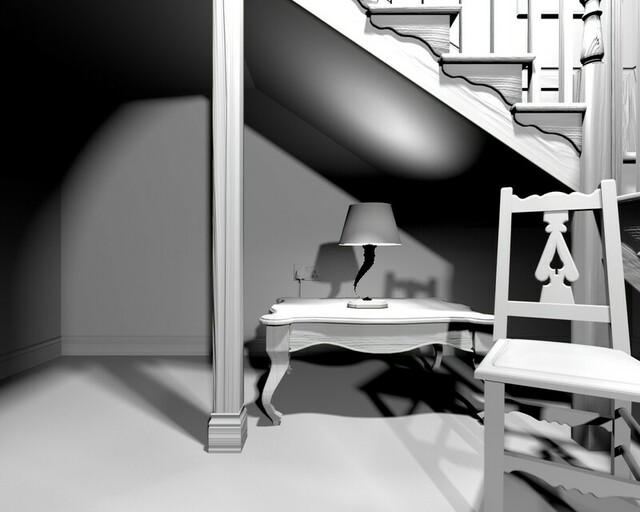}
         \caption{Diffuse Direct}
         \label{fig:five over x}
     \end{subfigure}
     \begin{subfigure}[b]{0.32\linewidth}
         \centering
         \includegraphics[width=\linewidth]{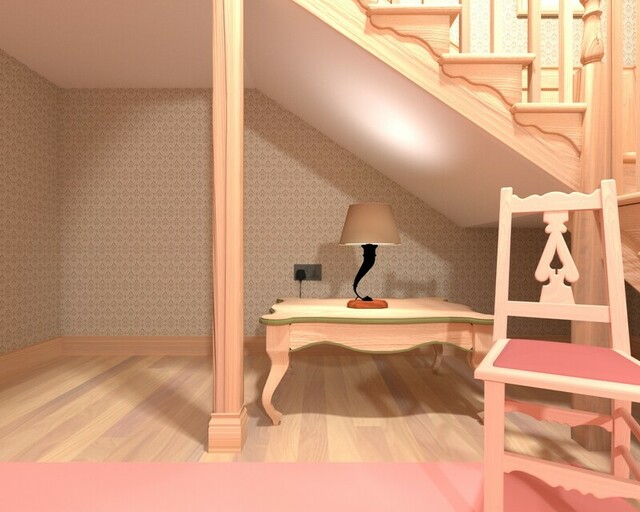}
         \caption{Diffuse}
         \label{fig:five over x}
     \end{subfigure}
     \begin{subfigure}[b]{0.32\linewidth}
         \centering
         \includegraphics[width=\linewidth]{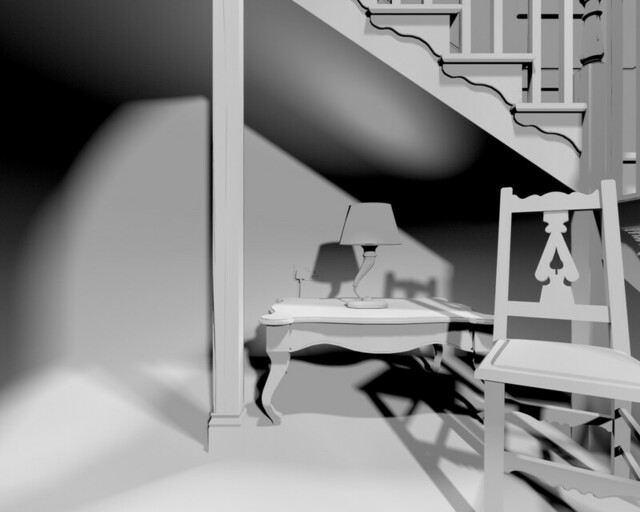}
         \caption{Shadow}
         \label{fig:five over x}
     \end{subfigure}
     \begin{subfigure}[b]{0.32\linewidth}
         \centering
         \includegraphics[width=\linewidth]{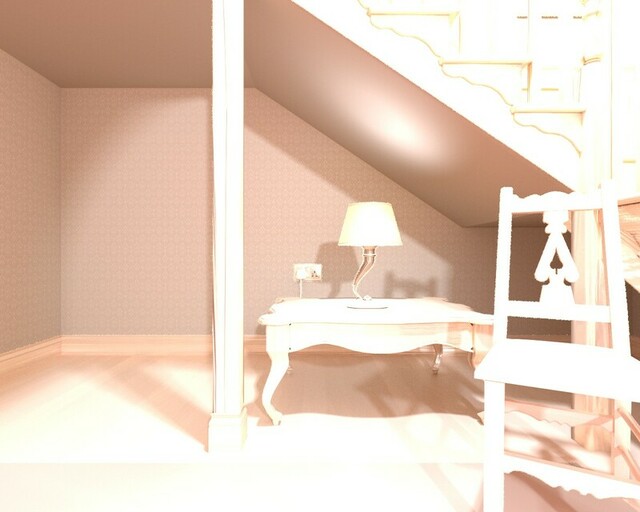}
         \caption{Lightmap}
         \label{fig:five over x}
     \end{subfigure}
     \begin{subfigure}[b]{0.32\linewidth}
         \centering
         \includegraphics[width=\linewidth]{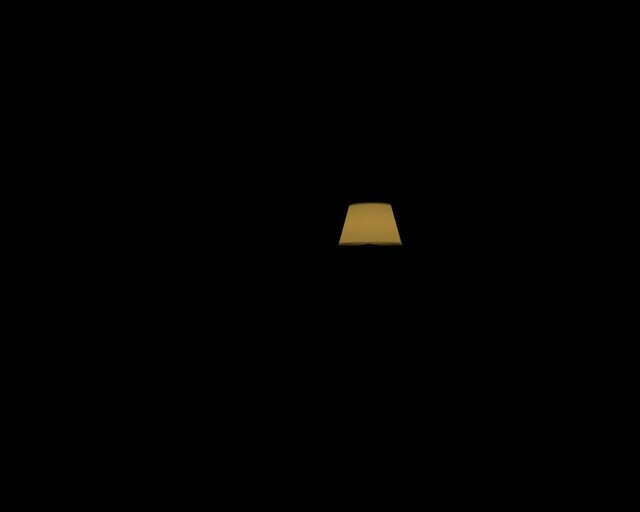}
         \caption{Transmission}
         \label{fig:five over x}
     \end{subfigure}
        \caption{Sample from scene 18}
        \label{fig:three graphs}
\end{figure}

\end{document}